\documentclass[11pt]{article}

\usepackage[]{ACL2023}

\usepackage{times}
\usepackage{latexsym}

\usepackage[T1]{fontenc}

\usepackage[utf8]{inputenc}
\usepackage{lipsum,caption,graphicx}
\usepackage{multirow}
\usepackage{booktabs}
\usepackage{afterpage}
\usepackage{adjustbox}
\usepackage{tabularx}
\usepackage{subfig}
\usepackage{xr}
\usepackage{xspace}
\usepackage{tablefootnote}
\usepackage{amsmath}
\usepackage{hyperref}
\usepackage{float}
\usepackage{adjustbox}

\usepackage{microtype}
\usepackage{xspace}

\usepackage{inconsolata}
\newcommand{\LaGoNN}{\textsc{LaGoNN}\xspace}
\newcommand{\LaGoNNexp}{\textsc{LaGoNN$_{exp}$}\xspace}
\newcommand{\LaGoNNlite}{\textsc{LaGoNN$_{lite}$}\xspace}
\newcommand{\LaGoNNcheap}{\textsc{LaGoNN$_{cheap}$}\xspace}
\newcommand{\labell}{\textsc{LABEL}\xspace}
\newcommand{\textt}{\textsc{TEXT}\xspace}
\newcommand{\all}{\textsc{ALL}\xspace}
\newcommand{\labdist}{\textsc{LabDist}\xspace}
\newcommand{\dist}{\textsc{DISTANCE}\xspace}
\newcommand{\labround}{\textsc{LabRound}\xspace}
\title{Like a Good Nearest Neighbor:\\ Practical Content Moderation and Text Classification}

\author{Luke Bates and Iryna Gurevych \\
  Ubiquitous Knowledge Processing Lab (UKP Lab)\protect\\ Department of Computer Science and Hessian Center for AI (hessian.AI)\protect\\ Technical University of Darmstadt \protect\\ \url{www.ukp.tu-darmstadt.de}\protect\\
  \texttt{firstname.lastname@tu-darmstadt.de}}

\begin{document}

\maketitle
\begin{abstract}
Few-shot text classification systems have impressive capabilities but are infeasible to deploy and use reliably due to their dependence on prompting and billion-parameter language models. SetFit \citep{https://doi.org/10.48550/arxiv.2209.11055} is a recent, practical approach that fine-tunes a Sentence Transformer under a contrastive learning paradigm and achieves similar results to more unwieldy systems. Inexpensive text classification is important for addressing the problem of domain drift in all classification tasks, and especially in detecting harmful content, which plagues social media platforms. Here, we propose Like a Good Nearest Neighbor (\LaGoNN), a modification to SetFit that introduces no learnable parameters but alters input text with information from its nearest neighbor, for example, the label and text, in the training data, making novel data appear similar to an instance on which the model was optimized. \LaGoNN is effective at flagging undesirable content and text classification, and improves SetFit's performance. To demonstrate \LaGoNN's value, we conduct a thorough study of text classification systems in the context of content moderation under four label distributions, and in general and multilingual classification settings.\footnote{Our code and data are available at \url{https://github.com/UKPLab/lagonn}.} 

\end{abstract}
\section{Introduction}
Text classification is the most important tool for NLP practitioners, and there has been substantial progress in advancing the state-of-the-art, especially with the advent of large, pretrained language models (PLM) \citep{devlin-etal-2019-bert}. Modern research focuses on in-context learning \citep{GPT3_2020}, pattern exploiting training \citep{schick-schutze-2021-exploiting, schick-schutze-2021-just, schick-schutze-2022-true}, adapter-based fine-tuning with learned label embeddings \citep{karimi-mahabadi-etal-2022-prompt}, and parameter efficient fine-tuning \citep{liu2020tfew}. These methods have achieved impressive results on the SuperGLUE \citep{NEURIPS2019_4496bf24} and RAFT \citep{alex2021raft} few-shot benchmarks, but most are difficult to use because of their reliance on billion-parameter PLMs, pay-to-use APIs, and/or prompting. Constructing prompts is not trivial and may require domain expertise.

One exception to these cumbersome systems is SetFit. SetFit does not rely on prompting or billion-parameter PLMs, and instead fine-tunes a pretrained Sentence Transformer (ST) \citep{S-BERT-reimers-gurevych-2019} under a contrastive learning paradigm. SetFit has comparable performance to more unwieldy systems while being one to two orders of magnitude faster to train and run inference.

An important application of text classification is aiding or automating content moderation, which is the task of determining the appropriateness of user-generated content on the Internet \citep{Roberts2017}. From fake news to toxic comments to hate speech, it is difficult to browse social media without being exposed to potentially dangerous posts that may have an effect on our ability to reason \citep{Ecker2022}. Misinformation spreads at alarming rates \citep{doi:10.1126/science.aap9559}, and an ML system should be able to quickly aid human moderators. While there is work in NLP with this goal \citep{openai2022moderation, shido-etal-2022-textual, multilingreddit}, a general, practical, and open-sourced method that is effective across multiple domains remains an open challenge. Novel fake news topics or racial slurs emerge and change constantly. Retraining of ML-based systems is required to adapt this concept drift, but this is expensive, not only in terms of computation, but also in terms of the human effort needed to collect and label data.

SetFit's performance, speed, and low cost would make it ideal for effective content moderation, however, this type of text classification proves difficult for even state-of-the-art approaches. For example, detecting hate speech on Twitter \citep{basile-etal-2019-semeval}, a subtask on the RAFT few-shot benchmark, appears to be the most difficult dataset; at time of writing, it is the only task where the human baseline has not been surpassed, yet SetFit is among the top ten most performant systems.\footnote{\url{https://huggingface.co/spaces/ought/raft-leaderboard} (see "Tweet Eval Hate").} 

\begin{figure*}[h!]%
    \centering
    \includegraphics[width=160mm]{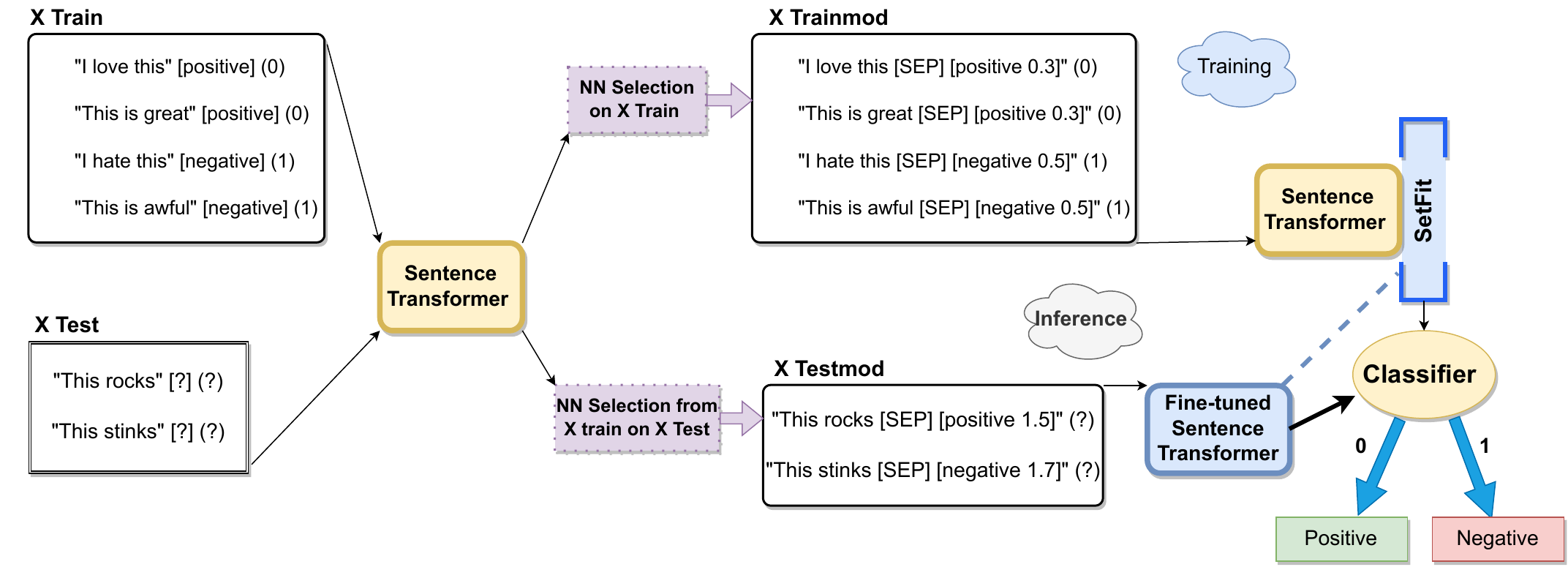}
    \caption{\LaGoNN \labdist uses an ST to encode training data, performs NN lookup, appends the NN's gold label and distance, and optionally SetFit to fine-tune the embedding model. We then embed this new instance and train a classifier. During inference, we use the embedding model to modify the test data with its NN's gold label and distance from the training data, compute the final representation, and call the classifier. Input text is in quotation marks, the NN's gold label and distance are in brackets, and the integer label is in parenthesis.}
    \label{LaGoNN}
\end{figure*}

Here, we propose a modification to SetFit, called Like a Good Nearest Neighbor (\LaGoNN). \LaGoNN  introduces no learnable parameters and instead modifies input text by retrieving information from its nearest neighbors (NN) seen during optimization. Specifically, we append the label, distance, and text of the NNs in the training data to a new instance and encode this modified version with an ST (see Figures \ref{lagonsimple} and \ref{LaGoNN} and Table \ref{table:LaGoNNtypes}). By making input data appear more similar to instances seen during training, we inexpensively exploit the ST's pretrained or fine-tuned knowledge when considering a novel example. Our method can also be applied to the linear probing of an ST, requiring no expensive fine-tuning of the large embedding model. Finally, we propose a simple alteration to the SetFit training procedure, where we fine-tune the ST on a subset of the training data. This results in a more efficient and performant text classifier that can be used with \LaGoNN. We summarize our contributions as follows:
 
\begin{enumerate}
    \item We propose \LaGoNN, an inexpensive modification to Sentence Transformer- or SetFit-based text classification.
    \item We suggest an alternative training procedure to the standard fine-tuning of SetFit, that can be used with or without \LaGoNN, and results in a cheaper system with similar or improved performance to the more expensive SetFit.
    \item We perform an extensive study of \LaGoNN, SetFit, and standard transformer fine-tuning in the context of content moderation under different label distributions, and in general and multilingual text classification settings.
\end{enumerate}

\section{Related Work}
There is little work on using sentence embeddings as features for classification despite the pioneering work being five years old \cite{Sent_emb_2018}. STs are pretrained with the objective of maximizing the distance between semantically distinct text and minimizing the distance between text that is semantically similar in feature space. They are composed of a Siamese and triplet architecture that encodes text into dense vectors which can be used as features for ML. STs were first used to embed text for classification by \citet{Sent_tranformers_2021}, however, only pretrained representations were examined. 

SetFit uses a contrastive learning paradigm \citep{koch2015siamese, dong-etal-2022-cml} to optimize the ST embedding model. The ST is fine-tuned with a distance-based loss function, like cosine similarity, such that examples with different labels are separated in feature space. Input text is then encoded with the fine-tuned ST and a classifier, such as logistic regression, is trained. This approach creates a strong, few-shot
text classification system, transforming the ST from a sentence
encoder to a topic encoder.

Work done by \citet{DBLP:journals/corr/abs-2112-03254} showed that retrieving and concatenating text from training data and external sources, such as ConceptNet \citep{Speer_Chin_Havasi_2017} and the Wiktionary\footnote{\url{https://www.wiktionary.org/}} definition, can be viewed as a type of external attention that does not alter the architecture of the Transformer in question answering.  \citet{liu-etal-2022-makes} used PLMs and $k$-NN lookup to prepend examples that are similar to a GPT-3 query, aiding in prompt engineering for in-context learning.  \citet{wang-etal-2022-training} demonstrated that prepending and appending training data helps PLMs in summarization, language modelling, machine translation, and question answering, using BM25 as their retrieval model \cite{bm252, bm25}.

\begin{table*}[t]
    \centering
    \begin{tabular}{c c}    
    \textbf{Training Data} & \textbf{Test Data} \\
    "I love this." [positive 0.0] (0) &  "So good!" [?] (?) \\
    "This is great!" [positive 0.5] (0) & "Just terrible!" [?] (?) \\
    "I hate this." [negative 0.7] (1) & "Never again." [?] (?) \\
    "This is awful!" [negative 1.2] (1) & "This rocks!" [?] (?) \\
     & 
    \end{tabular}
    \begin{adjustbox}{width=1\textwidth}
    \begin{tabular}{c c}
    \textbf{\LaGoNN Configuration} &  \textbf{Train Modified} \\
    \midrule
    \labell &  "I love this. [SEP] [positive]" (0)  \\
    \dist &  "I love this. [SEP] [0.5]" (0)  \\
    \labdist & "I love this. [SEP] [positive 0.5]" (0) \\
    \textt & "I love this. [SEP] [positive 0.5] This is great!" (0) \\
    \all & "I love this. [SEP] [positive 0.5] This is great! [SEP] [negative 0.7]  I hate this." (0) \\
    \\

     \hfill & \textbf{Test Modified} \\
    \labell &  "So good! [SEP] [positive]" (?) \\
    \dist &  "So good! [SEP] [1.5]" (?) \\
    \labdist &  "So good! [SEP] [positive 1.5] \\
    \textt &  "So good! [SEP] [positive 1.5] I love this." (?) \\
    \all &  "So good! [SEP] [positive 1.5] I love this. [SEP] [negative 2.7]  This is awful!" (?)\\
    \end{tabular}
    \end{adjustbox}
    \caption{Toy training and test data and different \LaGoNN configurations considering the first training example. Text is in quotation marks and the integer label is in parenthesis. In brackets are the gold label or distance from the NN or both. Train and Test Modified are altered instances that are input into the final embedding model for training and inference, respectively. The input format is "\textit{original text} [SEP] [(NN gold) (label distance)] NN \textit{training instance text}". See Appendix \ref{sec:appendixmod} for examples of \LaGoNN \all modified text.}
    \label{table:LaGoNNtypes}
\end{table*}

We alter the SetFit training procedure by using fewer examples to adapt the embedding model for many-shot learning. \LaGoNN decorates input text with its NN's gold label, Euclidean distance, and text from the training data to exploit both the ST's distance-based pretraining and SetFit's distance-based fine-tuning objective. Compared to retrieval-based methods, \LaGoNN uses the same model for both retrieval and encoding, retrieving only information from the training data for classification.

\section{Like a Good Nearest Neighbor}
\citet{DBLP:journals/corr/abs-2112-03254} formulate a type of external attention, where textual information is retrieved from multiple sources and added to text input to give the model stronger reasoning ability without altering the internal architecture. Inspired by this approach, \LaGoNN exploits pretrained and fine-tuned knowledge through external attention, but the information we retrieve comes only from data used during optimization. We consider an embedding function, $f$, that encodes both training and test data, $f(X_{train})$ and $f(X_{test})$. Considering its success on realistic, few-shot data and our goal of practical content moderation, we choose an ST that can be fine-tuned with SetFit as our embedding function.

\subparagraph{Encoding and nearest neighbors}
\LaGoNN first uses a pretrained Sentence Transformer to embed training text in feature space, $f(X_{train})$, and NN lookup with scikit-learn \citep{sklearn_api} on the resulting embeddings.

\subparagraph{Nearest neighbor information} We extract text from the nearest neighbors and use it to decorate the original example. We experimented with different text that \LaGoNN could use. The first configuration we consider is the gold label of the NN, which we call \labell. We then consider the Euclidean distance of the NN, which we call \dist, giving the model access to a continuous measure of similarity. We then combine these two configurations, appending both the NN's gold label and Euclidean distance, referring to this as \labdist. Next, we consider the gold label, distance, and the text of the NN, which we refer to as \textt. Finally, we tried the same format as \textt but for all possible labels, which we call \all (see Table \ref{table:LaGoNNtypes} and Figure \ref{LaGoNN}). Information from the NN is appended to the text following a separator token to indicate this instance is composed of multiple sequences. If we consider multiple neighbors, we append the information we consider sequentially based on the Euclidean distance from the input text separated by a separator token. That is, the first NN's information is followed by "[SEP]" and the second NN's information which is then followed by "[SEP]" and the third NN's information, etc. See Appendix \ref{sec:appendix_configs} for a detailed study of all \LaGoNN configurations.

\subparagraph{Training}
\LaGoNN encodes the modified training data, optionally fine-tunes the embedding model via SetFit, and trains a classifier, $CLF(f(X_{trainmod}))$.

\subparagraph{Inference}
\LaGoNN uses information from the nearest neighbor in the training data to modify input text. We compute the embeddings of the test data, $f(X_{test})$, and select and extract information from the NN's training text, decorating the input instance with this information. Finally, we encode the modified data with the embedding model and call the classifier, $CLF(f(X_{testmod}))$.

\subparagraph{Intuition}
The ST's pretraining and SetFit's fine-tuning objective both rely on distance, creating a feature space appropriate for distance-based algorithms, such as our NN-lookup. We hypothesize that \LaGoNN's modifications make novel data appear semantically similar to their NNs in the training data, that is, more akin to an instance on which the encoder and classifier were optimized. \LaGoNN's utilization of distance and clear distinctions between classes inspired our use case of content moderation, where it is realistic to have few labels, harmful or neutral, for example. However, this work demonstrates that \LaGoNN is useful for general and multilingual text classification as well.

\interfootnotelinepenalty=10000 
\section{Experiments}
We first study \LaGoNN's performance on four binary and one ternary classification dataset related to the task of content moderation. Each dataset is composed of a training, validation, and test split (see Appendix \ref{data_balance} for details).

 We study our system by simulating growing training data over ten discrete steps sampled under four different label distributions: extreme, imbalanced, moderate, and balanced (see Table \ref{table:alldistributions}). On each step we add $100$ examples (100 on the first, 200 on the second, etc.) from the training split sampled under one of the four ratios. On each step, we train our method with the sampled data and evaluate on the test split. Considering growing training data has two benefits: 1) We can simulate a streaming data scenario, where new data are labeled and added for training and 2) We can investigate each method's sensitivity to the number of training examples. 
 
 This experimental setup is reflective of a practical setting, where we might construct a content flagging or text classification system with a relatively small number (100) of labeled examples for training. As time goes on, however, more samples are added and we must then determine whether or not it is worth the resources to retrain our system. We sampled over five seeds, reporting the mean and standard deviation.
\begin{figure*}[h]
    \centering
    \includegraphics[scale=0.27]{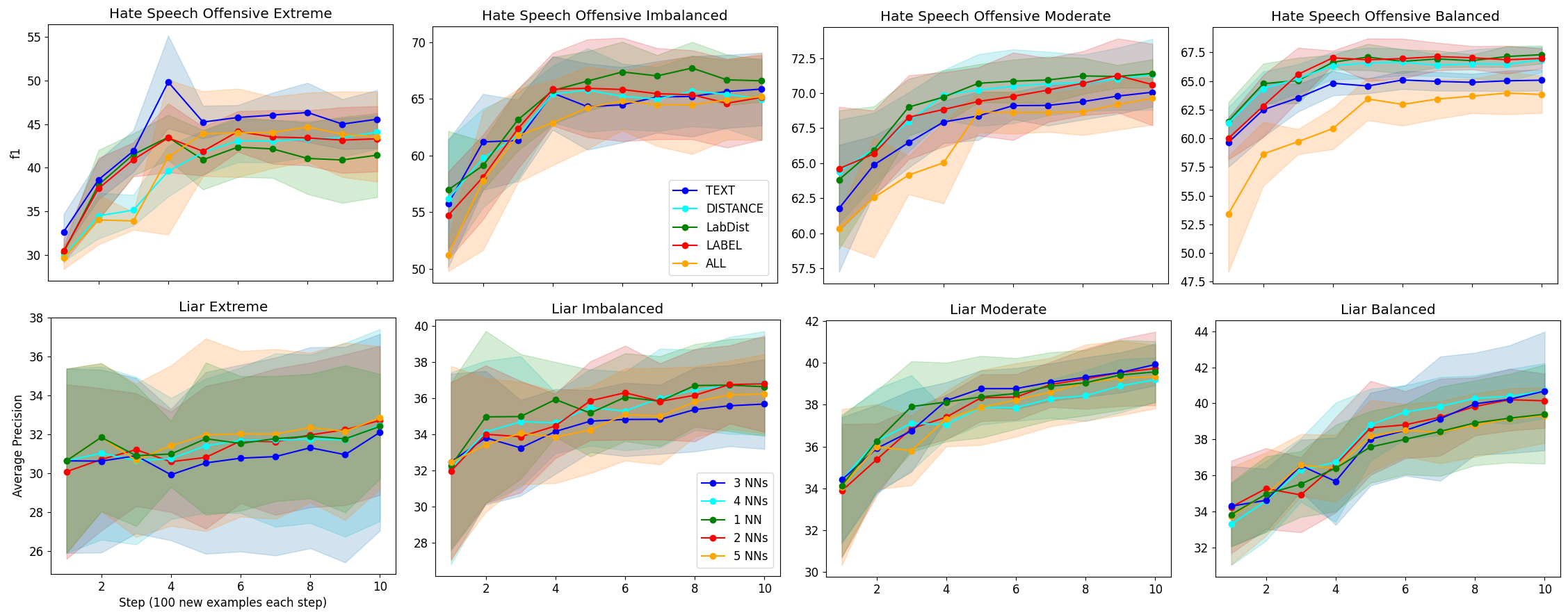}
    \captionof{figure}{First row: performance for all \LaGoNN configurations and balance regimes for the Hate Speech Offensive dataset. Second row: \LaGoNN performance for one to five neighbors for all balance regimes on a collapsed version of the LIAR dataset. We use the \LaGoNNlite fine-tuning strategy (see Section \ref{setfit_many}).}
    \label{config_pic}
\end{figure*}

\subsection{Baselines}
We compare \LaGoNN against a number of strong baselines, detailed below. We used default hyperparameters in all cases unless stated otherwise.

\subparagraph{RoBERTa}
RoBERTa-base is a pretrained language model \citep{liu2019roberta} that we fine-tuned with the transformers library \citep{wolf-etal-2020-transformers}. We select two versions of RoBERTa-base: an expensive version, where we perform standard fine-tuning on each step (RoBERTa$_{full}$) and a cheaper version, where we freeze the model body after step one and update the classification head on subsequent steps (RoBERTa$_{freeze}$). We set the learning rate to $1e^{-5}$, train for a maximum of 70 epochs, and use early stopping, selecting the best model after training. We consider RoBERTa$_{full}$ an upper bound as it has the most trainable parameters and requires the most time to train of all our methods.

\subparagraph{Linear probe}
We perform linear probing of a pretrained Sentence Transformer by fitting logistic regression with default hyperparameters on the training embeddings on each step. We choose this baseline because \LaGoNN can be applied as a modification in this scenario. We select MPNET \citep{NEURIPS2020_c3a690be} as the ST, for SetFit, and for \LaGoNN.\footnote{\url{https://huggingface.co/sentence-transformers/paraphrase-mpnet-base-v2}} We refer to this method as Probe.

\subparagraph{SetFit}
Here, we perform standard fine-tuning with SetFit on the first step, and then on subsequent steps, freeze the embedding model and retrain only the classification head. We choose this baseline as \LaGoNN relies on ST/SetFit for its modifications.

\subparagraph{$k$-nearest neighbors}
Similar to the above baseline, we fine-tune the embedding model via SetFit, but swap out the classification head for a $k$NN classifier, where $k = 3$. We select this baseline as \LaGoNN also relies on an NN lookup. $k = 3$ was chosen during our development stage as it yielded the strongest performance. We refer to this method as $k$NN.

\subparagraph{SetFit expensive}
For this baseline we perform standard fine-tuning with SetFit on each step. On the first step, this method is equivalent to SetFit. We refer to this as SetFit$_{exp}$.

\subparagraph{\LaGoNN cheap}
This method modifies data via \LaGoNN before fitting logistic regression. Even without adapting the embedding model, as the training data grow, modifications made to the test data may change. Only the classification head is fit on each step. We refer to this method as \LaGoNNcheap and it is comparable to Probe.

\subparagraph{\LaGoNN}
On the first step, we use \LaGoNN to modify our
data and perform standard fine-tuning with SetFit. On subsequent steps, we freeze the embedding model but continue to use it to modify our data. We only fit logistic regression on later steps, 
 referring to this method as \LaGoNN. It is comparable to SetFit.

\subparagraph{\LaGoNN expensive}
Here we modify our data and fine-tune the embedding model on each step. We refer to this method as \LaGoNNexp and it is comparable to SetFit$_{exp}$. On the first step, this method is equivalent to \LaGoNN.

\subparagraph{Model choices} We again choose these systems to reflect different practical settings, where we might not have the resources to fine-tune our model (Probe/\LaGoNNcheap), we might be able to perform limited fine-tuning (RoBERTa$_{freeze}$, SetFit, $k$NN, \LaGoNN), or we may be able to fine-tune as much as we like (RoBERTa$_{full}$, SetFit$_{exp}$, \LaGoNNexp).

\begin{table*}[th]
\centering
\begin{adjustbox}{width=1\textwidth}
\begin{tabular}{l c c c c | c c c c}
\textbf{Method}& & \textbf{InsincereQs} &  & &  & \textbf{AmazonCF} &   & \\
\textit{Extreme} & $1^{st}$ & $5^{th}$ & $10^{th}$ & Average & $1^{st}$ & $5^{th}$ & $10^{th}$  & Average \\
\midrule
RoBERTa$_{full}$\ & $19.9_{8.4}$ & $30.9_{7.9}$ & $42.0_{7.4}$ & $33.5_{6.7}$ & $21.8_{6.6}$ & $63.9_{10.2}$ & $72.3_{3.0}$ & $59.6_{16.8}$  \\
SetFit$_{exp}$\  & $24.1_{6.3}$ & $29.2_{6.7}$ & $36.7_{7.3}$  & $31.7_{3.4}$  & $22.3_{8.8}$ & $64.2_{3.3}$ &  $68.6_{4.6}$ & $56.8_{14.9}$  \\
\LaGoNNexp \ & $\textbf{30.7}_{8.9}$  & $37.6_{6.1}$  & $39.0_{6.1}$  & $36.1_{2.3}$ & $\textbf{26.1}_{17.5}$  & $\textbf{68.4}_{4.4}$ & $\textbf{74.9}_{2.9}$ & $\textbf{63.2}_{16.7}$ \\
\midrule
RoBERTa$_{freeze}$\ & $19.9_{8.4}$  & $34.1_{5.4}$  & $37.9_{5.9}$ & $32.5_{5.5}$  & $21.8_{6.6}$ & $41.0_{12.7}$ &  $51.3_{10.7}$ & $40.6_{8.9}$ \\
$k$NN\     & $6.8_{0.42}$  & $15.9_{3.4}$ & $16.9_{4.3}$  & $14.4_{3.0}$  & $10.3_{0.2}$ & $15.3_{4.2}$ & $18.4_{3.7}$ & $15.6_{2.4}$ \\

SetFit\ & $24.1_{6.3}$  & $31.7_{4.9}$ & $36.1_{5.4}$  & $31.8_{3.6}$  & $22.3_{8.8}$ & $32.4_{11.5}$ & $42.3_{8.8}$ & $34.5_{5.9}$  \\
\LaGoNN \ & $\textbf{30.7}_{8.9}$  & $39.3_{4.9}$ & $41.2_{4.7}$  & $38.4_{3.0}$ & $\textbf{26.1}_{17.5}$ & $31.1_{19.4}$ & $33.0_{19.1}$ & $30.9_{2.3}$ \\
\midrule
Probe\   & $24.3_{8.4}$  & $39.8_{5.6}$ & $44.8_{4.2}$ & $38.3_{6.2}$  & $24.2_{9.0}$ & $46.3_{4.4}$ & $54.6_{2.0}$ & $45.1_{10.3}$ \\
\LaGoNNcheap\  & $23.6_{7.8}$  & $\textbf{40.7}_{5.9}$ & $\textbf{45.3}_{4.4}$ & $\textbf{38.6}_{6.6}$ & $20.1_{6.9}$  & $38.3_{4.9}$ & $47.8_{3.4}$  & $38.2_{9.5}$ \\
\midrule
\textit{Balanced}\\
RoBERTa$_{full}$\  & $47.1_{4.2}$ & $52.1_{3.6}$ & $55.7_{2.6}$ & $52.5_{2.9}$ & $73.6_{2.1}$ & $78.6_{3.9}$ & $\textbf{82.4}_{1.1}$ & $78.9_{2.2}$ \\
SetFit$_{exp}$\ & $43.5_{4.2}$ & $47.1_{4.6}$ & $48.5_{3.9}$ & $48.0_{1.7}$  & $73.8_{4.4}$ & $69.8_{4.0}$ & $64.1_{4.6}$  & $69.6_{3.6}$  \\
\LaGoNNexp \ & $ 42.8_{5.3}$  & $47.6_{2.9}$ &  $47.0_{1.7}$ & $46.2_{2.0}$ & $\textbf{76.0}_{3.0}$& $73.4_{2.6}$ & $72.3_{2.9}$ & $72.5_{3.4}$ \\
\midrule
RoBERTa$_{freeze}$\ & $47.1_{4.2}$  & $52.1_{0.4}$ & $53.3_{1.7}$ & $51.5_{2.1}$ & $73.6_{2.1}$ & $76.8_{1.6}$ & $77.9_{1.0}$ & $76.5_{1.3}$ \\
$k$NN\             & $22.3_{2.3}$  & $30.2_{2.3}$ & $30.9_{1.8}$ & $29.5_{2.5}$ & $41.7_{3.4}$ & $57.9_{3.3}$ & $58.3_{3.3}$ & $56.8_{5.1}$ \\
SetFit\         & $43.5_{4.2}$  & $53.8_{2.2}$ & $55.5_{1.6}$ & $52.8_{3.5}$ & $73.8_{4.4}$ & $79.2_{1.9}$ & $80.1_{1.0}$ & $78.6_{1.8}$ \\
\LaGoNN \         & $42.8_{5.3}$  & $54.1_{2.9}$ & $56.3_{1.3}$ & $53.4_{3.7}$ & $\textbf{76.0}_{3.0}$ & $\textbf{80.1}_{2.0}$ & $81.4_{1.1}$ & $\textbf{79.8}_{1.4}$ \\
\midrule
Probe\           & $47.5_{1.6}$  & $52.4_{1.7}$ & $55.3_{1.1}$ & $52.2_{2.5}$ & $52.4_{3.4}$ & $64.7_{2.5}$ & $67.5_{0.4}$ & $63.4_{4.4}$ \\
\LaGoNNcheap \ &$\textbf{49.3}_{2.6}$ & $\textbf{54.4}_{1.4}$ & $\textbf{57.6}_{0.7}$ & $\textbf{54.2}_{2.7}$ & $48.1_{3.4}$ & $62.0_{2.0}$ & $ 65.3_{0.8}$ & $60.5_{5.0}$ \\

\bottomrule
\end{tabular}
\end{adjustbox}
\caption{\label{table:extreme}
Average performance (average precision $\times$ 100) on Insincere Questions and Amazon Counterfactual. The first, fifth, and tenth step are followed by the average over all ten steps. The average gives insight into the overall strongest performer by aggregating all steps. We group methods with a comparable number of trainable parameters together. The extreme label distribution results are followed by balanced (see Appendix \ref{sec:appendixadditional} for additional results).
}
\end{table*}
\subsection{\LaGoNN configurations}
We perform extensive experiments over the different \LaGoNN configurations. We note that while \dist and \labell show similar performance, \labdist in general is the most performant and consistent classifier. We base this assertion on the fact that across all of our experiments, \labdist  is generally in the top three most-performant configurations and is easily the stablest, based on the standard deviation over five seeds, where \dist and \labell are less reliable and show greater oscillation. These observations are supported by Figure \ref{config_pic} and in Appendix \ref{sec:appendix_configs}. \textt and \all are arguably the most interesting \LaGoNN configurations, but are often unstable, low-performing classifiers.  In Figure \ref{config_pic}, we provide a comparison between the different configurations on the Hate Speech Offensive dataset. As \labdist is the most performant configuration, it is the version of our method about which we report results hereafter, and we consider it the default configuration of \LaGoNN. However, this is a hyperparameter that can be easily experimented with and tuned. Detailed ablations can be found in Appendix \ref{sec:appendix_configs}.

\subsection{\LaGoNN $k$ nearest neighbors}
To determine how many neighbors we should consider for \LaGoNN, we perform thorough experiments for one to five neighbors over all datasets, \LaGoNN configurations, and balance regimes under the \LaGoNNlite fine-tuning strategy (see Section \ref{setfit_many}). We find that one to three neighbors tends to result in the strongest classifier, but this varies and is a hyperparameter that can be searched over. In Figure \ref{config_pic}, we provide a representative example of our NN results for the \labdist configuration for the LIAR dataset, however, detailed ablations can be found in Appendix \ref{nns_lagonn}.

\section{Content Moderation Results}
Table \ref{table:extreme} and Figure \ref{LaGoNN_results} show our results. In the cases of the extreme and imbalanced regimes, the performance of SetFit$_{exp}$ steadily increases with the number of training examples. As the label distribution shifts to the balanced regime, however, the performance quickly saturates or even degrades as the number of training examples grows. \LaGoNN, RoBERTa$_{full}$, and SetFit, other fine-tuned PLM classifiers, do not exhibit this behavior. \LaGoNNexp, being based on SetFit$_{exp}$, exhibits a similar trend, but the performance degradation is mitigated; on the $10^{th}$ step of Amazon Counterfactual in Table \ref{table:extreme} SetFit$_{exp}$'s performance decreased by 9.7, while \LaGoNNexp only fell by 3.7. Note that we only consider the first NN here. 

\begin{figure*}[t]
    \centering
    \includegraphics[scale=0.35]{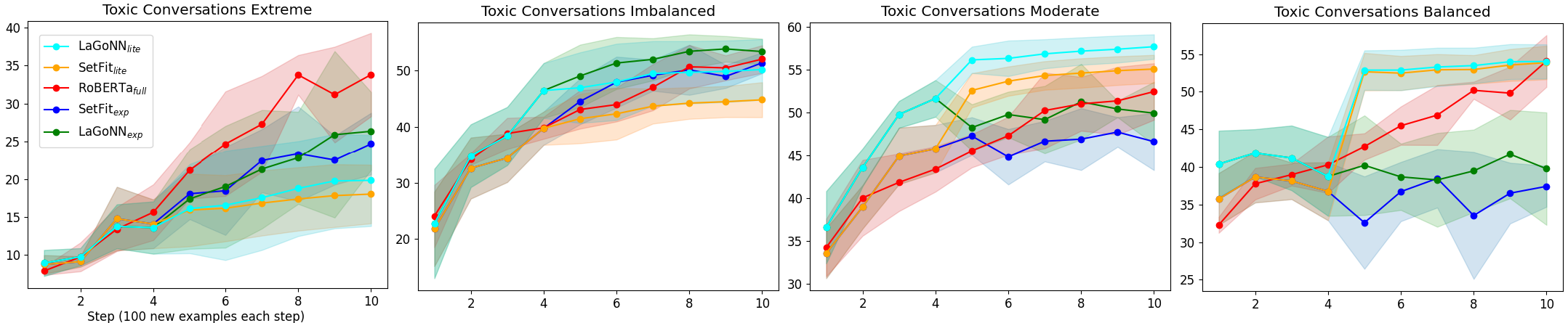}
    \captionof{figure}{Average performance for all sampling regimes on Toxic Conversations. More expensive models, such as \LaGoNNexp, SetFit$_{exp}$, and RoBERTa$_{full}$ perform best when the label distribution is imbalanced. As the distribution becomes more balanced, inexpensive models, such as \LaGoNNlite, show similar or improved performance. The measure is average precision and we only consider one neighbor for the \LaGoNN-based methods (see Appendix \ref{sec:appendixadditional2} for additional results).}
    \label{LaGoNN_resultsmany}
\end{figure*}

\LaGoNN and \LaGoNNexp generally outperform SetFit and SetFit$_{exp}$, respectively, often resulting in a more stable model, as reflected in the standard deviation. We find that \LaGoNN and \LaGoNNexp exhibit stronger predictive power with fewer examples than RoBERTa$_{full}$ despite having fewer trainable parameters. On the first step of Insincere Questions under the extreme setting, \LaGoNN's performance is more than 10 points higher.

\LaGoNNcheap outperforms all other methods on the Insincere Questions dataset for all balance regimes, despite being the third fastest (see Table \ref{table:time}) and having the second fewest trainable parameters. We attribute this result to the fact that this dataset is composed of questions from Quora\footnote{\url{https://www.quora.com/}} and our ST backbone was pretrained on similar data. This intuition is supported by Probe, the cheapest method, which despite having the fewest trainable parameters, shows comparable performance. 

\subsection{SetFit for efficient many-shot learning}
\label{setfit_many}
Respectively comparing SetFit to SetFit$_{exp}$ and \LaGoNN to \LaGoNNexp suggests that fine-tuning the ST embedding model on moderate or balanced data hurts model performance as the number of training samples grows. We therefore hypothesize that randomly sampling a subset of training data to fine-tune the encoder, freezing, embedding the remaining data, and training the classifier will result in a stronger model.  

To test our hypothesis, we add two models to our experimental setup: SetFit$_{lite}$ and \LaGoNNlite. SetFit$_{lite}$ and \LaGoNNlite are respectively equivalent to SetFit$_{exp}$ and \LaGoNNexp, except after the fourth step (400 samples), we freeze the encoder and only retrain the classifier on subsequent steps, similar to SetFit and \LaGoNN.

Figures \ref{LaGoNN_resultsmany} and \ref{LaGoNN_resultsmany2} show our results with these two new models. As expected, in the cases of extreme and imbalanced distributions, \LaGoNNexp, SetFit$_{exp}$, and RoBERTa$_{full}$, are the strongest performers. We note very different results for both \LaGoNNlite and SetFit$_{lite}$ compared to \LaGoNNexp and SetFit$_{exp}$ on Toxic Conversations under the moderate and balanced label distributions. As their expensive counterparts start to plateau or degrade on the fourth step, these two new models dramatically increase, showing improved or comparable performance to RoBERTa$_{full}$, despite being optimized on less data; for example, \LaGoNNlite reaches an average precision of approximately 55 after being optimized on only 500 examples. RoBERTa$_{full}$ does not exhibit similar performance until the tenth step. Finally, we point out that \LaGoNN-based methods generally provide a performance boost for SetFit-based methods.

\section{\LaGoNN as a General Classifier}
\label{general_main}
\LaGoNN is effective for general text classification. Thus far, we have focused on the important topic of content moderation, but here we turn our attention to general text classification, conducting experiments on 12 additional datasets (see Appendix \ref{general_data} for details and Appendix \ref{sec:multilingualappendix} for multilingual experiments). Our experimental setup remains largely the same, but here we restrict ourselves to the balanced sampling regime as it is nontrivial to design sampling strategies for datasets with more than three labels. We respectively compare \LaGoNNlite against SetFit$_{lite}$ and \LaGoNNexp against SetFit$_{exp}$, showing results for one to five neighbors with \LaGoNN.

\begin{figure*}[t]
    \centering
    \includegraphics[scale=0.267]{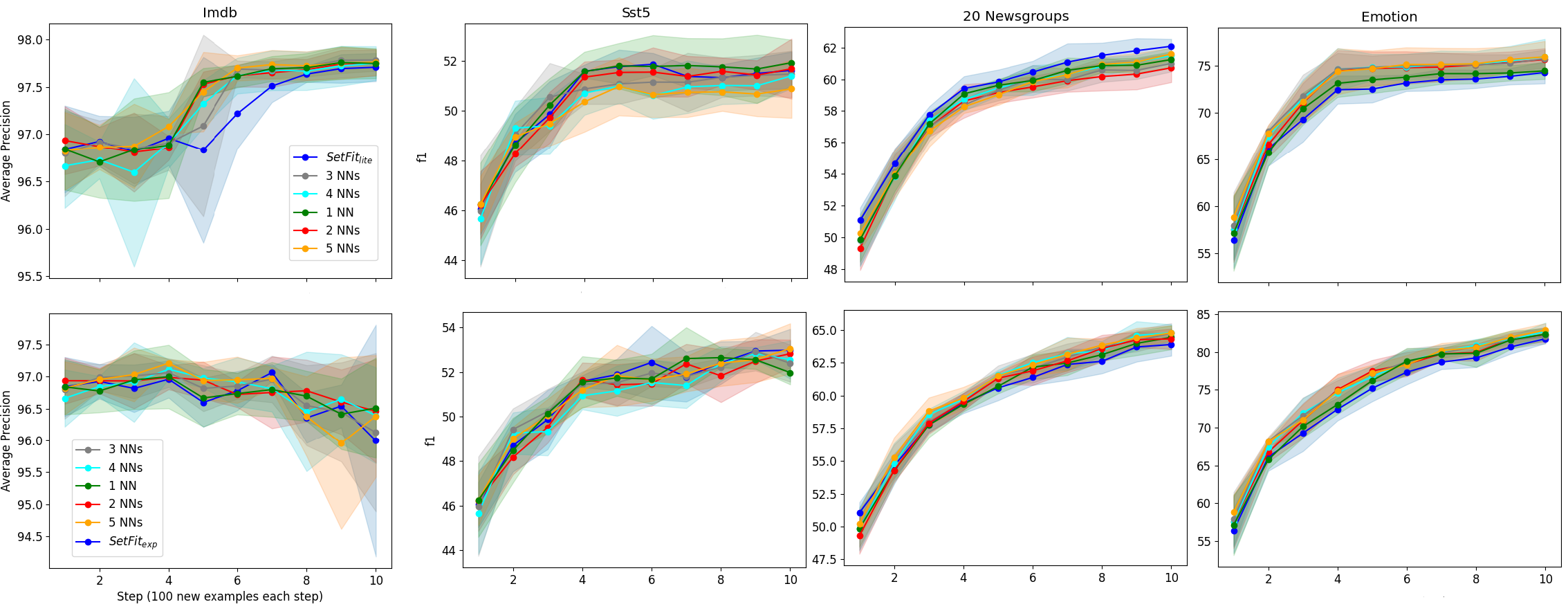}
    \captionof{figure}{Average performance on four datasets in the balanced sampling regime; the measure is average precision for IMDB, macro-f1 elsewhere. First row: SetFit$_{lite}$ compared to \LaGoNNexp \labdist with modifications for one to five neighbors. Second row: SetFit$_{exp}$ compared to \LaGoNNexp.  See Appendix \ref{sec:generalappendix} for additional results.}
    \label{LaGoNN_last}
\end{figure*}

In Figure \ref{LaGoNN_last}, we demonstrate that \LaGoNN continues to stabilize and improve SetFit, regardless of the number of neighbors we consider. This is especially clear for IMDB, where in the case of \LaGoNNlite vs SetFit$_{lite}$, all versions of our method saturate to an average precision of 98 with 300 fewer training samples. If we consider SetFit$_{exp}$ vs \LaGoNNexp, consistent with our analysis of other binary datasets, classifier performance begins to degrade if we continue to fine-tune the ST, but \LaGoNN mitigates this performance drop.

Continuing to fine-tune the embedding model is beneficial when we have many labels. For 20 Newsgroups and Emotion, which have 20 and 28 labels respectively, \LaGoNNexp is the strongest model and shows no indication of plateauing or degrading, even with 1,000 samples. We attribute this to the relatively high number of labels present in both of these datasets. Our findings related to SST-5 and our multilingual experiments (see Appendix \ref{sec:multilingualappendix}) support this; in intermediate cases when we have five labels, all models saturate quickly and there are minimal gains with continued fine-tuning.

\section{Discussion}
Flagging potentially dangerous text presents a challenge even for state-of-the-art approaches. The content moderation datasets we consider proved more difficult than our general text classification datasets for all models, despite typically having fewer labels. It is imperative that we develop reliable and practical text classifiers for content moderation, such that we can inexpensively re-tune them for novel forms of hate speech, toxicity, and fake news.  

Our results suggest that \LaGoNNexp, a relatively expensive technique, can detect harmful content when dealing with imbalanced label distributions, as is common with realistic datasets. This is intuitive from the perspective that less common instances are more difficult to learn and require more effort. An exception would be our examination of Insincere Questions, where \LaGoNNcheap excelled in the extreme and balanced settings. This demonstrates that if we choose our PLM with care for related downstream tasks, \LaGoNN can inexpensively extract pretrained knowledge and improve performance without the need for costly fine-tuning. Indeed, considering the performance of SetFit suggests that, in this case, fine-tuning hurts performance and we actually overfit. However, even here, our proposed modifications with \LaGoNN increase model robustness and lessen the effects of overfitting.

Fine-tuning with SetFit hurts performance on more balanced datasets that are not few-shot. We have observed that SetFit should not be applied "out of the box" to balanced, non-few-shot data. This can be detrimental to performance, directly affecting our own approach. However, \LaGoNN can stabilize SetFit's predictions and reduce its performance drop in many cases. Figures \ref{LaGoNN_results}, \ref{LaGoNN_resultsmany}, and \ref{LaGoNN_last} show that when the label distribution is moderate or balanced (see Table \ref{table:alldistributions}), SetFit$_{exp}$ plateaus, yet cheaper systems, such as \LaGoNN, continue to learn. This is likely due to SetFit's fine-tuning objective, which optimizes an ST using cosine similarity loss to separate examples belonging to different labels in feature space, assuming independence between labels. This may be too strong an assumption as we fine-tune with more data, which is counter-intuitive for data-hungry transformers; RoBERTa$_{full}$, optimized with cross-entropy loss, showed improved performance as we added training data data.

For balanced data, it is sufficient to fine-tune the Sentence Transformer via SetFit with 50 to 100 examples per label, while 150 to 200 instances appear to be sufficient when the training data are moderately balanced. The encoder can then be frozen and all available data embedded to train a classifier. This is more performant and efficient than full-model fine-tuning. \LaGoNN is applicable to this case, inexpensively boosting and stabilizing SetFit's performance. All models fine-tuned on Hate Speech Offensive exhibited similar, upward-trending learning curves, but we note the speed of \LaGoNN relative to RoBERTa$_{full}$ or SetFit$_{exp}$ (see Figure \ref{LaGoNN_resultsmany} and Table \ref{table:time}).

\section{Conclusion}
We have proposed \LaGoNN, an inexpensive modification to SetFit. \LaGoNN improves SetFit's performance by modifying text with the nearest neighbors in the training data. To demonstrate the merit of \LaGoNN, we examined text classification systems for content moderation with different label distributions and for general and multilingual classification. We studied 17 datasets with growing training data. When the training labels are imbalanced, expensive systems, such as \LaGoNNexp are performant. \LaGoNNexp also excels on balanced datasets with many labels. However, when the labels are binary or ternary, typical for content moderation, and the distribution is balanced, fine-tuning with SetFit yields minimal gains. We therefore proposed an alternative but strong training procedure. \LaGoNN is a practical method for detecting harmful content and text classification.

\section{Acknowledgments}
We would like to thank Derek Hommel and Nils Reimers for sharing inspiring discussions with us. We would also like to extend our gratitude to Tom Aarsen, Max Glockner, Yongxin Huang, Timour Igamberdiev, Sukannya Purkayastha, and Kexin Wang for their invaluable feedback on an early draft of our manuscript. This work was funded by the German Federal Ministry of Education and Research
and the Hessian Ministry of Science and the Arts
(HMWK) within the projects "The Third Wave
of Artificial Intelligence - 3AI", hessian.AI, and
within their joint support of the National Research
Center for Applied Cybersecurity ATHENE.

\section{Limitations}
In the current work, we have only considered text data, but social media content can of course consist of text, images, and videos. As \LaGoNN depends only on an embedding model, an obvious extension to our approach would be examining the modifications we suggest, but on multimodal data. This is an interesting direction that we leave for future research. We did not study our method when there are fewer than 100 training examples, and investigating \LaGoNN in a few-shot learning setting is fascinating topic for future study. Finally, we note that our system could be misused to detect undesirable content that is not necessarily harmful. For example, a social media website could detect and silence users who complain about the platform. This is not our intended use case, but could result from any classifier, and potential misuse is an unfortunate drawback of all technology.

\section{Ethics Statement}
It is our sincere goal that our work contributes to the social good in multiple ways. We first hope to have furthered research on text classification that can be feasibly applied to combat undesirable content, such as misinformation, on the Internet, which could potentially cause someone harm. To this end, we have tried to describe our approach as accurately as possible and released our code and data, such that our work is transparent and can be easily reproduced and expanded upon. We hope that we have also created a useful but efficient system which reduces the need to expend energy in the form expensive computation. For example, \LaGoNN does not rely on billion-parameter language models that demand thousand-dollar GPUs to use. \LaGoNN makes use of GPUs no more than SetFit, despite being more computationally expensive. We have additionally proposed a simple method to make SetFit, an already relatively inexpensive method, even more efficient.

\bibliography{main}
\bibliographystyle{acl_natbib}

\appendix
\label{sec:appendix}
\section{Appendix}
\subsection{Content moderation data and balance regimes}
\label{data_balance}
In this Appendix section, we provide a background on the datasets we studied in our experiments and summarize the label distribution (see Table \ref{dataset}) of our content moderation datasets and the different sampling regimes (see Table \ref{table:alldistributions}) we studied in our content moderation experiments. 
LIAR was created from Politifact\footnote{\url{https://www.politifact.com/}} for fake news detection and is composed of the data fields \textit{context}, \textit{speaker}, and \textit{statement}, which are labeled with varying levels of truthfulness \citep{wang-2017-LIAR}. We used a collapsed version of this dataset where a statement can only be true or false. We did not use \textit{speaker}, but did use \textit{context} and \textit{statement}, separated by a separator token. Quora Insincere Questions\footnote{\url{https://www.kaggle.com/c/quora-insincere-questions-classification}} is composed of neutral and toxic questions, where the author is not asking in good faith. Hate Speech Offensive\footnote{\url{https://huggingface.co/datasets/hate_speech_offensive}} has three labels and is composed of tweets that can contain either neutral text, offensive language, or hate speech \citep{hateoffensive}.\footnote{For Hate Speech Offensive, 0 and 2 denote undesirable text and 1 denotes neither.} Amazon Counterfactual\footnote{\url{https://huggingface.co/datasets/SetFit/amazon_counterfactual_en}} contains sentences from product reviews, and the labels can be "factual" or "counterfactual" \citep{oneill-etal-2021-wish}. "Counterfactual" indicates that the customer said something that cannot be true. Finally, Toxic Conversations\footnote{\url{https://huggingface.co/datasets/SetFit/toxic_conversations}} is a dataset of comments where the author wrote with unintended bias\footnote{\url{https://www.kaggle.com/c/jigsaw-unintended-bias-in-toxicity-classification}} (see Table \ref{dataset}).

\begin{table} [h!]
    \begin{adjustbox}{width=0.5\textwidth}
    \begin{tabular}{c|c}
    \textbf{Dataset (and Detection Task)} & \textbf{Number of Labels} \\
    \midrule
     LIAR (Fake News) & 2 \\
     Insincere Questions (Toxicity) & 2 \\
     Hate Speech Offensive & 3 \\
     Amazon Counterfactual (English) & 2\\
     Toxic Conversations & 2 \\
    \end{tabular}
    \end{adjustbox}
    \caption{Summary of content moderation datasets and number of labels. We provide the type of task in parenthesis in unclear cases.}
    \label{dataset}
\end{table}

\begin{table} [H]
    \centering
    \begin{adjustbox}{width=0.45\textwidth}
    \begin{tabular}{c|c|c}
    \textbf{Regime} & \textbf{Binary} & \textbf{Ternary} \\
    \midrule
     Extreme & 0: 98\% 1: 2\% & 0: 95\%, 1: 2\%, 2: 3\% \\
     Imbalanced & 0: 90\% 1: 10\% & 0: 80\%, 1: 5\%, 2: 15\% \\
     Moderate & 0: 75\% 1: 25\% & 0: 65\%, 1: 10\%, 2: 25\% \\
     Balanced & 0: 50\% 1: 50\% & 0: 33\%, 1: 33\%, 2: 33\% \\
    \end{tabular}
    \end{adjustbox}
    \caption{Label distributions for sampling training data. 0 represents neutral while 1 and 2 represent different types of undesirable text.}
    \label{table:alldistributions}
\end{table}

\subsection{General text classification data}
\label{general_data}
In this Appendix section, we provide additional information on the datasets we examined in our general text classification experiments. The Internet Movie Database (IMDB) dataset \cite{maas-EtAl:2011:ACL-HLT2011} is composed of movie reviews that are classified as either positive or negative.\footnote{\url{https://huggingface.co/datasets/SetFit/imdb}} Student Question Categories contains questions from qualifying examinations in India,\footnote{\url{https://www.kaggle.com/datasets/mrutyunjaybiswal/iitjee-neet-aims-students-questions-data}} where the label is the subject the question appeared in and can be from Physics, Chemistry, Biology, or Mathematics.\footnote{\url{https://huggingface.co/datasets/SetFit/student-question-categories}} SST5 is an alternative version of the Stanford Sentiment Treebank \cite{socher-etal-2013-recursive} that has five labels, ranging from very positive to very negative.\footnote{\url{https://huggingface.co/datasets/SetFit/sst5}} We also include the original version of LIAR, which has six labels of varying levels of truthfulness.\footnote{\url{https://huggingface.co/datasets/LIAR}} We also used 20 Newsgroups\footnote{\url{https://scikit-learn.org/0.19/datasets/twenty_newsgroups.html\#the-20-newsgroups-text-dataset}} \cite{misc_twenty_newsgroups_113} which contains newspaper articles labeled with the topic they cover.\footnote{\url{https://huggingface.co/datasets/SetFit/20_newsgroups}} And finally, we ran experiments on GoEmotions \cite{demszky-etal-2020-goemotions}, a dataset of Reddit comments labeled with 28 classes based on the emotional charge of the post.\footnote{\url{https://huggingface.co/datasets/SetFit/go_emotions}}

The evaluation measure was average precision in the case of IMDB, macro F1 elsewhere. In cases where the a validation split was not available, we created one by sampling 30\% of the test split. Please see Table \ref{general_tabel} for a summary regarding the datasets and label information.

\begin{table} [h!]
    \begin{adjustbox}{width=0.5\textwidth}
    \begin{tabular}{c|c}
    \textbf{Dataset (and Detection Task)} & \textbf{Number of Labels} \\
    \midrule
     IMDB (Sentiment Analysis) & 2 \\
     Student Questions (Question Type) & 4 \\
     SST5 (Sentiment Analysis) & 5 \\
     LIAR (Fake News) & 6\\
     20 Newsgroups (Topic) & 20 \\
     GoEmotions (Emotion) & 28 \\
    \end{tabular}
    \end{adjustbox}
    \caption{Summary of datasets and number of labels used in the general text classification experiments. We provide the type of task in parenthesis in unclear cases.}
    \label{general_tabel}
\end{table}

\subsection{Observations about \LaGoNN}
\begin{figure}[t]%
    \centering
    \includegraphics[width=58mm]{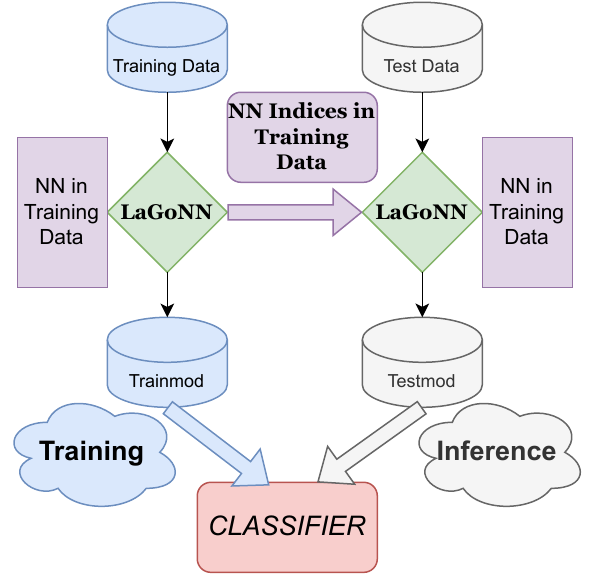}
    \caption{We embed training data, retrieve the text, gold label, and distance for each instance from its nearest neighbor and modify the original text with this information. Then we embed the modified training data and train a classifier. During inference, the NN from the training data is selected, the original text is modified with the text, gold label, and distance from this NN, and the classifier is called.} 
    \label{lagonsimple}
\end{figure}
Here, at the suggestion of an anonymous reviewer, we include a little background on \LaGoNN. We originally attempted to use Sentence Transformers/SetFit as a retrieval model that would modify input text and then pass this input to a Transformer-based classifier, such as RoBERTa, instead of back into the ST as in LaGoNN. We experimented with different ST retrieval models and Transformer classifiers, but this system was often beaten by baselines, and performant versions were too expensive to justify their use. The failure of this system is what ultimately inspired \LaGoNN. We had hoped to construct a system that did not need to be updated after step one and could simply perform inference on subsequent steps, an active learning setup. While the performance of this version of \LaGoNN did not degrade, it also did not appear to learn anything and we found it necessary to update parameters on each step. 
We additionally tried fine-tuning the embedding model via SetFit first before modifying data, however, this hurt performance in all cases. We include this information for transparency and because we find it interesting.

\subsection{\LaGoNN's computational expense}
In this Appendix section we discuss and provide results for \LaGoNN's computation time.
\LaGoNN is more computationally expensive than Sentence Transformer- or SetFit-based text classification. \LaGoNN introduces additional inference with the encoder, NN-lookup, and string modification. As the computational complexity of transformers increases with sequence length \citep{NIPS2017_3f5ee243}, additional expense is created when \LaGoNN appends textual information before inference with the ST. In Table \ref{table:time}, we provide a speed comparison of comparable methods computed on the same hardware.\footnote{We used a 40 GB NVIDIA A100 Tensor Core GPU.} On average, \LaGoNN introduced 24.2 additional seconds of computation compared to its relative counterpart.

\begin{table}[h!]
    \centering
    \begin{tabular}{c|c}
    \textbf{Method}& \textbf{Time in seconds} \\
    \midrule
     Probe & 22.9 \\
    \LaGoNNcheap & 44.2 \\
    SetFit & 42.9 \\
    \LaGoNN & 63.4 \\
    SetFit$_{exp}$ & 207.3 \\
    \LaGoNNexp & 238.0 \\
    \midrule
    RoBERTa$_{full}$ & 446.9\\
    \end{tabular}
    \caption{Speed comparison between \LaGoNN \labdist with one neighbor and comparable methods. Time includes training on $1,000$ examples and inference on $51,000$ examples.}
    \label{table:time}
\end{table}

\newpage

\begin{figure*}[t!]
    \centering
    \includegraphics[scale=0.28]{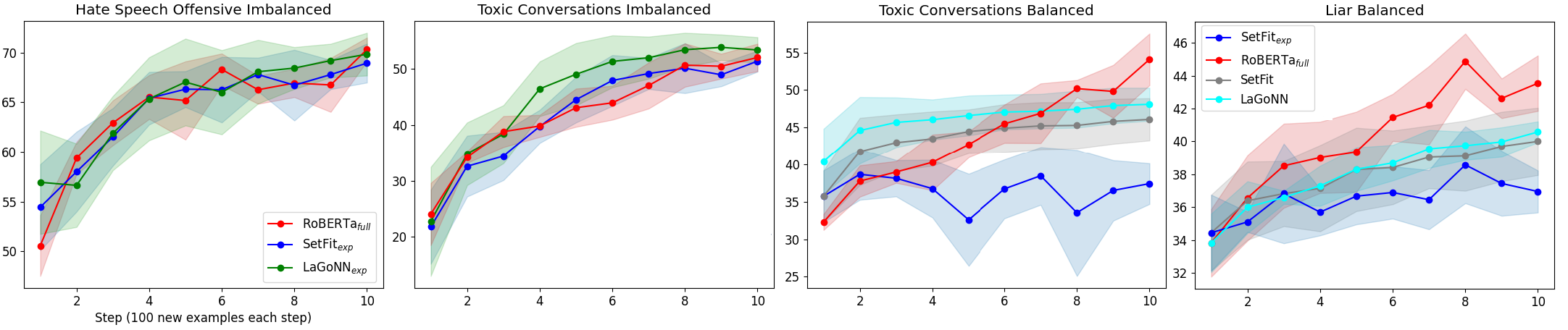}
    \captionof{figure}{Average performance in the imbalanced and balanced regimes relative to comparable methods. We include RoBERTa$_{full}$ results for reference. The measure is macro-F1 for Hate Speech Offensive, average precision elsewhere.}
    \label{LaGoNN_results}
\end{figure*}
\subsection{Additional results: initial experiments}
\label{sec:appendixadditional}
Here we provide additional results from our initial experimental setup that, due to space limitations, could not be included in the main text. We note that a version of \LaGoNN outperforms or has the same performance of all methods, including our upper bound RoBERTa$_{full}$, on 54\% of all displayed results, and is the best performer relative to Sentence Transformer-based methods on  72\%. This excludes \LaGoNNcheap. This method showed strong performance on the Insincere Questions dataset, but hurts performance in other cases. 
\indent In cases when SetFit-based methods do outperform our system, the performances are comparable, usually within a point, yet they can be quite dramatic when \LaGoNN-based methods are the strongest. Below, we report the mean average precision $\times 100$ for all methods over five seeds with the standard deviation, except in the case of Hate Speech Offensive, where the evaluation measure is the macro-F1. Each table shows the results for a given dataset and a given label-balance distribution on the first, fifth, and tenth step followed by the average for all ten steps. In the table caption we provide a summary/interpretation of the results for a given setting.
The LIAR dataset seems to be the most difficult for all methods. This is expected because it likely does not include enough context to determine the truth of a statement.\

\begin{table}[H]
\centering
\begin{adjustbox}{width=0.5\textwidth}
\begin{tabular}{l c c c c}
\textbf{Method}& & \textbf{Insincere-Questions} &  & \\
\textit{Imbalanced} & $1^{st}$ & $5^{th}$ & $10^{th}$ & Average \\
\midrule
RoBERTa$_{full}$\ & $39.8_{5.5}$ & $\textbf{53.1}_{4.6}$ & $\textbf{55.7}_{1.2}$ & $50.6_{4.4}$  \\
SetFit$_{exp}$\  & $43.7_{2.7}$ & $52.2_{1.9}$ & $53.8_{0.9}$  & $51.4_{2.9}$  \\
\LaGoNNexp \ & $\textbf{44.5}_{4.5}$  & $52.7_{2.4}$  & $55.4_{2.0}$  & $\textbf{51.8}_{3.0}$  \\
\midrule
RoBERTa$_{freeze}$\ & $39.8_{5.5}$  & $44.1_{3.6}$  & $46.3_{2.4}$ & $44.0_{2.0}$  \\
$k$NN\     & $23.9_{2.2}$  & $30.3_{3.0}$ & $31.6_{2.4}$  & $30.0_{2.1}$   \\

SetFit\ & $43.7_{2.7}$  & $47.6_{1.6}$ & $50.1_{2.1}$  & $47.6_{1.8}$   \\
\LaGoNN \ & $\textbf{44.5}_{4.5}$  & $48.1_{2.2}$ & $50.3_{1.7}$  & $48.1_{1.9}$  \\
\midrule
Probe\   & $40.4_{4.2}$  & $49.4_{2.3}$ & $52.3_{1.7}$ & $49.0_{3.3}$  \\
\LaGoNNcheap\  & $40.8_{4.3}$  & $51.1_{2.4}$ & $54.5_{1.4}$ & $50.4_{4.0}$ \\
\bottomrule
\end{tabular}
\end{adjustbox}
\caption{\LaGoNN and \LaGoNNexp are the strongest performers on the first step, but are overtaken by RoBERTa$_{full}$ on later steps. The average of all steps shows that \LaGoNNexp is the overall strongest performer, but we note that \LaGoNNcheap shows comparable performance to RoBERTa$_{full}$ despite being much less expensive.}
\end{table}

\begin{table}[H]
\centering
\begin{adjustbox}{width=0.5\textwidth}
\begin{tabular}{l c c c c}
\textbf{Method}& & \textbf{Insincere Questions} &  & \\
\textit{Moderate} & $1^{st}$ & $5^{th}$ & $10^{th}$ & Average \\
\midrule
RoBERTa$_{full}$\ & $48.1_{2.3}$ & $\textbf{54.7}_{1.9}$ & $\textbf{57.5}_{1.5}$ & $53.9_{2.9}$  \\
SetFit$_{exp}$\  & $48.9_{1.7}$ & $53.9_{0.7}$ & $54.2_{1.5}$  & $52.3_{1.6}$  \\
\LaGoNNexp \ & $\textbf{49.8}_{1.6}$  & $52.2_{1.9}$  & $53.2_{3.3}$  & $52.0_{1.4}$  \\
\midrule
RoBERTa$_{freeze}$\ & $48.1_{2.3}$  & $50.2_{2.2}$  & $52.0_{1.4}$ & $50.2_{1.4}$  \\
$k$NN\     & $28.0_{2.4}$  & $33.9_{2.8}$ & $33.6_{2.0}$  & $33.5_{1.9}$   \\

SetFit\ & $48.9_{1.7}$  & $53.6_{1.9}$ & $55.8_{1.7}$  & $53.3_{2.2}$   \\
\LaGoNN \ & $\textbf{49.8}_{1.6}$  & $54.4_{1.3}$ & $56.9_{0.5}$  & $\textbf{54.2}_{2.2}$  \\
\midrule
Probe\   & $45.7_{2.1}$  & $52.3_{1.8}$ & $54.4_{1.1}$ & $51.4_{2.5}$  \\
\LaGoNNcheap\  & $45.7_{2.2}$  & $54.4_{1.6}$ & $56.4_{0.6}$ & $53.2_{3.2}$ \\
\bottomrule
\end{tabular}
\end{adjustbox}
\caption{\LaGoNN and \LaGoNNexp are the strongest performers on the first step, but are overtaken by RoBERTa$_{full}$ on later steps. The average of all steps shows that \LaGoNN is the overall strongest performer, but we note that \LaGoNNcheap shows comparable performance to RoBERTa$_{full}$ despite being much less expensive.}
\end{table}

\begin{table}[H]
\centering
\begin{adjustbox}{width=0.5\textwidth}
\begin{tabular}{l c c c c}
\textbf{Method}& & \textbf{Amazon Counterfactual} &  & \\
\textit{Imbalanced} & $1^{st}$ & $5^{th}$ & $10^{th}$ & Average \\
\midrule
RoBERTa$_{full}$\ & $68.2_{4.5}$ & $\textbf{81.0}_{1.7}$ & $\textbf{82.2}_{1.0}$ & $79.2_{3.9}$  \\
SetFit$_{exp}$\  & $72.0_{2.1}$ & $78.4_{2.8}$ & $78.8_{1.2}$  & $78.0_{2.1}$  \\
\LaGoNNexp \ & $\textbf{74.3}_{3.8}$  & $80.1_{1.4}$  & $79.0_{1.6}$  & $\textbf{79.5}_{1.9}$  \\
\midrule
RoBERTa$_{freeze}$\ & $68.2_{4.5}$  & $75.0_{2.2}$  & $77.0_{2.4}$ & $74.2_{2.6}$  \\
$k$NN\     & $51.0_{4.1}$  & $60.0_{3.1}$ & $61.3_{2.1}$  & $59.7_{3.0}$   \\

SetFit\ & $72.0_{2.1}$  & $74.4_{2.3}$ & $76.7_{1.8}$  & $74.8_{1.4}$   \\
\LaGoNN \ & $\textbf{74.3}_{3.8}$  & $76.1_{3.6}$ & $77.3_{3.2}$  & $76.1_{1.0}$  \\
\midrule
Probe\   & $46.6_{2.8}$  & $60.3_{1.4}$ & $64.2_{1.2}$ & $59.2_{5.2}$  \\
\LaGoNNcheap\  & $38.2_{3.2}$  & $55.3_{1.8}$ & $61.0_{1.2}$ & $54.4_{6.7}$ \\
\bottomrule
\end{tabular}
\end{adjustbox}
\caption{\LaGoNN and \LaGoNNexp are the strongest performers on the first step, but are overtaken by RoBERTa$_{full}$ on later steps. However, the average of all steps shows that \LaGoNNexp is the overall strongest performer.}
\end{table}

\begin{table}[H]
\centering
\begin{adjustbox}{width=0.5\textwidth}
\begin{tabular}{l c c c c}
\textbf{Method}& & \textbf{Amazon Counterfactual} &  & \\
\textit{Moderate} & $1^{st}$ & $5^{th}$ & $10^{th}$ & Average \\
\midrule
RoBERTa$_{full}$\ & $73.9_{2.5}$ & $80.0_{1.0}$ & $80.1_{2.3}$ & $79.1_{2.1}$  \\
SetFit$_{exp}$\  & $76.5_{1.6}$ & $77.0_{2.4}$ & $74.7_{0.5}$  & $76.5_{1.0}$  \\
\LaGoNNexp \ & $\textbf{78.6}_{2.2}$  & $78.0_{2.1}$  & $76.3_{4.9}$  & $78.2_{1.0}$  \\
\midrule
RoBERTa$_{freeze}$\ & $73.9_{2.5}$  & $76.6_{1.4}$  & $78.5_{0.7}$ & $76.4_{1.7}$  \\
$k$NN\     & $54.5_{3.1}$  & $64.2_{1.9}$ & $66.6_{1.3}$  & $64.7_{3.5}$   \\

SetFit\ & $76.5_{1.6}$  & $80.6_{0.5}$ & $81.2_{0.3}$  & $80.0_{1.4}$   \\
\LaGoNN \ & $\textbf{78.6}_{2.2}$  & $\textbf{81.2}_{1.4}$ & $\textbf{81.6}_{1.1}$  & $\textbf{80.8}_{0.9}$  \\
\midrule
Probe\   & $52.3_{2.0}$  & $64.1_{1.8}$ & $67.2_{1.4}$ & $63.1_{4.3}$  \\
\LaGoNNcheap\  & $47.3_{3.4}$  & $60.7_{1.5}$ & $65.2_{1.4}$ & $59.5_{5.2}$ \\
\bottomrule
\end{tabular}
\end{adjustbox}
\caption{\LaGoNNexp and \LaGoNN are the strongest performers on the first step, but \LaGoNN is strongest classifier on subsequent steps and is also the overall strongest performer based on the average over all steps.}
\end{table}
\clearpage

\begin{table}[H]
\centering
\begin{adjustbox}{width=0.5\textwidth}
\begin{tabular}{l c c c c}
\textbf{Method}& & \textbf{Toxic Conversations} &  & \\
\textit{Extreme} & $1^{st}$ & $5^{th}$ & $10^{th}$ & Average \\
\midrule
RoBERTa$_{full}$\ & $7.9_{0.5}$ & $21.2_{3.7}$ & $\textbf{33.8}_{5.5}$ & $21.9_{9.3}$  \\
SetFit$_{exp}$\  & $8.8_{1.2}$ & $18.1_{3.4}$ & $24.7_{4.1}$  & $17.6_{5.5}$  \\
\LaGoNNexp \ & $8.9_{1.7}$  & $17.4_{6.6}$  & $26.4_{5.2}$  & $17.9_{6.0}$  \\
\midrule
RoBERTa$_{freeze}$\ & $7.9_{0.5}$  & $12.8_{2.4}$  & $19.1_{3.2}$ & $13.5_{3.5}$  \\
$k$NN\     & $7.9_{0.0}$  & $8.7_{0.4}$ & $8.7_{0.2}$  & $8.5_{0.3}$   \\

SetFit\ & $8.8_{1.2}$  & $13.1_{2.5}$ & $16.3_{3.0}$  & $13.0_{2.6}$   \\
\LaGoNN \ & $8.9_{1.7}$  & $13.8_{3.9}$ & $17.1_{4.8}$  & $13.4_{2.6}$  \\
\midrule
Probe\   & $\textbf{13.1}_{2.8}$  & $\textbf{24.6}_{2.6}$ & $30.1_{2.1}$ & $\textbf{23.9}_{5.6}$  \\
\LaGoNNcheap\  & $11.3_{2.2}$  & $21.7_{2.7}$ & $27.4_{2.3}$ & $21.3_{5.3}$ \\
\bottomrule
\end{tabular}
\end{adjustbox}
\caption{Probe is strongest performer on every step, except the 10$^{th}$ where it is overtaken by RoBERTa$_{full}$. If we average over all steps, we see that Probe is the strongest performer. We note, however, that \LaGoNN and \LaGoNNexp outperform SetFit and SetFit$_{exp}$ on all steps.}
\end{table}

\begin{table}[H]
\centering
\begin{adjustbox}{width=0.5\textwidth}
\begin{tabular}{l c c c c}
\textbf{Method}& & \textbf{Toxic Conversations} &  & \\
\textit{Imbalanced} & $1^{st}$ & $5^{th}$ & $10^{th}$ & Average \\
\midrule
RoBERTa$_{full}$\ & $\textbf{24.1}_{5.6}$ & $43.1_{3.4}$ & $52.1_{2.5}$ & $42.4_{8.2}$  \\
SetFit$_{exp}$\  & $21.8_{6.6}$ & $44.5_{4.1}$ & $51.4_{1.9}$  & $42.1_{9.3}$  \\
\LaGoNNexp \ & $22.7_{9.8}$  & $\textbf{49.1}_{5.6}$  & $\textbf{53.4}_{2.3}$  & $\textbf{45.6}_{9.8}$  \\
\midrule
RoBERTa$_{freeze}$\ & $\textbf{24.1}_{5.6}$  & $31.2_{4.4}$  & $34.0_{4.0}$ & $30.5_{3.1}$  \\
$k$NN\     & $11.5_{2.5}$  & $14.7_{4.0}$ & $15.3_{3.2}$  & $14.6_{1.1}$   \\

SetFit\ & $21.8_{6.6}$  & $26.7_{5.3}$ & $30.2_{4.0}$  & $26.6_{2.7}$   \\
\LaGoNN \ & $22.7_{9.8}$  & $27.6_{8.9}$ & $30.3_{8.7}$  & $27.4_{2.4}$  \\
\midrule
Probe\   & $23.3_{2.7}$  & $33.0_{2.8}$ & $37.1_{1.8}$ & $32.5_{4.2}$  \\
\LaGoNNcheap\  & $20.5_{3.2}$  & $31.1_{3.2}$ & $35.6_{1.8}$ & $30.5_{4.6}$ \\
\bottomrule
\end{tabular}
\end{adjustbox}
\caption{RoBERTa$_{full}$ and RoBERTa$_{freeze}$ are the strongest performers on the first step, but are overtaken by \LaGoNNexp for the subsequent steps. The overall strongest performer based on the average over all steps is \LaGoNNexp.}
\end{table}

\begin{table}[H]
\centering
\begin{adjustbox}{width=0.5\textwidth}
\begin{tabular}{l c c c c}
\textbf{Method}& & \textbf{Toxic Conversations} &  & \\
\textit{Moderate} & $1^{st}$ & $5^{th}$ & $10^{th}$ & Average \\
\midrule
RoBERTa$_{full}$\ & $34.2_{3.4}$ & $45.5_{1.9}$ & $52.4_{3.3}$ & $45.7_{5.6}$  \\
SetFit$_{exp}$\  & $33.6_{2.9}$ & $47.2_{2.2}$ & $46.6_{3.3}$  & $44.3_{4.3}$  \\
\LaGoNNexp \ & $\textbf{36.6}_{4.2}$  & $\textbf{48.2}_{2.7}$  & $\textbf{49.9}_{3.7}$  & $\textbf{48.0}_{4.4}$  \\
\midrule
RoBERTa$_{freeze}$\ & $34.2_{3.4}$  & $38.4_{2.1}$  & $39.5_{1.8}$ & $38.0_{1.5}$  \\
$k$NN\     & $19.4_{1.9}$  & $21.5_{3.4}$ & $22.4_{2.9}$  & $21.6_{0.8}$   \\

SetFit\ & $33.6_{2.9}$  & $39.2_{2.9}$ & $41.6_{2.7}$  & $38.6_{2.4}$   \\
\LaGoNN \ & $\textbf{36.6}_{4.2}$  & $42.7_{3.7}$ & $45.0_{3.5}$  & $42.0_{2.5}$  \\
\midrule
Probe\   & $29.0_{2.7}$  & $36.1_{1.2}$ & $39.1_{1.5}$ & $35.5_{3.3}$  \\
\LaGoNNcheap\  & $26.1_{2.7}$  & $34.3_{1.3}$ & $37.5_{1.8}$ & $33.6_{3.6}$ \\
\bottomrule
\end{tabular}
\end{adjustbox}
\caption{\LaGoNN and \LaGoNNexp are the strongest performers on the first step and \LaGoNNexp remains the strongest for subsequent steps, also being the strongest classifier overall based on the average.}
\end{table}

\begin{table}[H]
\centering
\begin{adjustbox}{width=0.5\textwidth}
\begin{tabular}{l c c c c}
\textbf{Method}& & \textbf{Toxic Conversations} &  & \\
\textit{Balanced} & $1^{st}$ & $5^{th}$ & $10^{th}$ & Average \\
\midrule
RoBERTa$_{full}$\ & $32.3_{1.1}$ & $42.7_{1.8}$ & $\textbf{54.1}_{3.4}$ & $43.8_{6.3}$  \\
SetFit$_{exp}$\  & $35.7_{3.4}$ & $32.6_{6.2}$ & $37.4_{2.7}$  & $36.5_{1.9}$  \\
\LaGoNNexp \ & $\textbf{40.4}_{4.4}$  & $40.2_{6.6}$  & $39.8_{7.5}$  & $40.0_{1.2}$  \\
\midrule
RoBERTa$_{freeze}$\ & $32.3_{1.1}$  & $39.2_{1.5}$  & $41.0_{0.6}$ & $38.5_{2.4}$  \\
$k$NN\     & $17.4_{0.8}$  & $23.7_{2.6}$ & $24.3_{2.7}$  & $23.1_{2.0}$   \\

SetFit\ & $35.7_{3.4}$  & $44.5_{2.9}$ & $46.1_{2.8}$  & $43.6_{2.9}$   \\
\LaGoNN \ & $\textbf{40.4}_{4.4}$  & $\textbf{46.6}_{2.7}$ & $48.1_{2.2}$  & $\textbf{46.1}_{2.2}$  \\
\midrule
Probe\   & $29.5_{2.4}$  & $35.9_{0.9}$ & $40.2_{0.9}$ & $36.1_{3.5}$  \\
\LaGoNNcheap\  & $26.8_{2.7}$  & $34.5_{1.3}$ & $38.5_{0.8}$ & $34.4_{3.7}$ \\
\bottomrule
\end{tabular}
\end{adjustbox}
\caption{\LaGoNN and \LaGoNNexp are the strongest performers on the first step. \LaGoNN remains the strongest until the 10$^{th}$, where it is overtaken by RoBERTa$_{full}$. Overall, \LaGoNN is the strongest classifier based on the average. Note the performance of SetFit$_{exp}$ and \LaGoNNexp. While both degrade after the first step, \LaGoNNexp's performance drop is dramatically mitigated.}
\end{table}

\clearpage

\begin{table}[H]
\centering
\begin{adjustbox}{width=0.5\textwidth}
\begin{tabular}{l c c c c}
\textbf{Method}& & \textbf{Hate Speech Offensive} &  & \\
\textit{Extreme} & $1^{st}$ & $5^{th}$ & $10^{th}$ & Average \\
\midrule
RoBERTa$_{full}$\ & $30.2_{1.4}$ & $43.5_{2.5}$ & $\textbf{51.2}_{2.2}$ & $\textbf{44.3}_{7.4}$  \\
SetFit$_{exp}$ \  & $30.3_{0.8}$ & $\textbf{44.0}_{1.3}$ & $51.1_{2.0}$  & $43.8_{6.5}$  \\
\LaGoNNexp \ & $30.3_{0.7}$  & $40.7_{2.9}$  & $49.1_{4.4}$  & $42.2_{6.2}$  \\
\midrule
RoBERTa$_{freeze}$\ & $30.2_{1.4}$  & $33.5_{3.1}$  & $34.4_{3.4}$ & $33.1_{1.4}$  \\
$k$NN\     & $\textbf{31.5}_{1.2}$  & $35.9_{2.7}$ & $37.4_{2.0}$  & $35.8_{1.7}$   \\

SetFit \ & $30.3_{0.8}$  & $38.4_{2.5}$ & $41.1_{1.5}$  & $37.8_{3.3}$   \\
\LaGoNN \ & $30.3_{0.7}$  & $35.7_{2.6}$ & $39.1_{2.4}$  & $35.6_{2.7}$  \\
\midrule
Probe\   & $29.0_{0.2}$  & $34.7_{1.5}$ & $40.1_{2.1}$ & $35.1_{3.8}$  \\
\LaGoNNcheap\  & $29.0_{0.1}$  & $36.9_{1.8}$ & $40.5_{2.1}$ & $36.2_{3.7}$ \\
\bottomrule
\end{tabular}
\end{adjustbox}
\caption{$k$NN is the strongest performer on the first step, while SetFit$_{exp}$ is on the 5$^{th}$, and RoBERTa$_{full}$ is the strongest on the 10$^{th}$ while also being strongest overall performer for all steps. \LaGoNN-based methods are generally beaten by ST/SetFit-based baselines, with the exception of \LaGoNNcheap which consistently outperforms Probe. }
\end{table}

\begin{table}[H]
\centering
\begin{adjustbox}{width=0.5\textwidth}
\begin{tabular}{l c c c c}
\textbf{Method}& & \textbf{Hate Speech Offensive} &  & \\
\textit{Imbalanced} & $1^{st}$ & $5^{th}$ & $10^{th}$ & Average \\
\midrule
RoBERTa$_{full}$\ & $50.6_{3.0}$ & $65.2_{3.9}$ & $\textbf{70.3}_{1.2}$ & $64.2_{5.3}$  \\
SetFit$_{exp}$\  & $54.4_{4.3}$ & $66.3_{1.8}$ & $68.9_{2.0}$  & $64.3_{4.5}$  \\
\LaGoNNexp \ & $\textbf{57.0}_{5.2}$  & $\textbf{67.0}_{4.4}$  & $69.8_{2.1}$  & $\textbf{64.9}_{4.6}$  \\
\midrule
RoBERTa$_{freeze}$\ & $50.6_{3.0}$  & $54.1_{1.6}$  & $55.3_{2.3}$ & $54.1_{1.3}$  \\
$k$NN\     & $55.6_{4.8}$  & $57.3_{2.3}$ & $58.8_{3.6}$  & $57.4_{1.1}$   \\

SetFit\ & $54.4_{4.3}$  & $57.0_{3.9}$ & $58.2_{3.8}$  & $57.2_{1.1}$   \\
\LaGoNN \ & $\textbf{57.0}_{5.2}$  & $58.2_{4.1}$ & $58.3_{3.4}$  & $58.3_{0.6}$  \\
\midrule
Probe\   & $46.5_{2.2}$  & $57.8_{1.7}$ & $60.3_{1.2}$ & $56.5_{4.5}$  \\
\LaGoNNcheap\  & $47.1_{1.3}$  & $56.5_{2.2}$ & $59.5_{2.5}$ & $55.6_{3.8}$ \\
\bottomrule
\end{tabular}
\end{adjustbox}
\caption{\LaGoNN and \LaGoNNexp are the strongest performers on the first step, with \LaGoNNexp being the strongest on the 5$^{th}$ and RoBERTa$_{full}$ taking over on the 10$^{th}$. \LaGoNNexp is the strongest performer overall based on the average over all steps.}
\end{table}

\begin{table}[H]
\centering
\begin{adjustbox}{width=0.5\textwidth}
\begin{tabular}{l c c c c}
\textbf{Method}& & \textbf{Hate Speech Offensive} &  & \\
\textit{Moderate} & $1^{st}$ & $5^{th}$ & $10^{th}$ & Average \\
\midrule
RoBERTa$_{full}$\ & $61.9_{3.4}$ & $70.8_{1.0}$ & $\textbf{72.5}_{1.4}$ & $69.9_{3.2}$  \\
SetFit$_{exp}$\  & $\textbf{64.3}_{4.2}$ & $70.6_{2.4}$ & $72.4_{0.5}$  & $69.8_{2.8}$  \\
\LaGoNNexp \ & $63.8_{4.9}$  & $\textbf{71.0}_{2.1}$  & $72.3_{1.0}$  & $\textbf{70.0}_{3.0}$  \\
\midrule
RoBERTa$_{freeze}$\ & $61.9_{3.4}$  & $63.2_{4.1}$  & $64.1_{4.5}$ & $63.2_{0.6}$  \\
$k$NN\     & $\textbf{64.3}_{4.0}$  & $63.3_{2.9}$ & $63.9_{2.5}$  & $63.7_{0.4}$   \\

SetFit \ & $\textbf{64.3}_{4.2}$  & $67.3_{3.2}$ & $67.6_{2.3}$  & $66.9_{1.1}$   \\
\LaGoNN \ & $63.8_{4.9}$  & $65.0_{5.3}$ & $66.7_{5.9}$  & $65.3_{0.9}$  \\
\midrule
Probe\   & $55.6_{1.7}$  & $63.8_{0.8}$ & $66.1_{0.3}$ & $63.2_{3.0}$  \\
\LaGoNNcheap\  & $56.0_{3.6}$  & $62.2_{1.4}$ & $66.0_{0.9}$ & $62.3_{2.9}$ \\
\bottomrule
\end{tabular}
\end{adjustbox}
\caption{$k$NN, SetFit, and SetFit$_{exp}$ start the strongest, but are overtaken by \LaGoNNexp on the 5$^{th}$ step, which is in turn overtaken by RoBERTa$_{full}$ on the 10$^{th}$ step. Overall \LaGoNNexp is the strongest performer based on the average.}
\end{table}

\begin{table}[H]
\centering
\begin{adjustbox}{width=0.5\textwidth}
\begin{tabular}{l c c c c}
\textbf{Method}& & \textbf{Hate Speech Offensive} &  & \\
\textit{Balanced} & $1^{st}$ & $5^{th}$ & $10^{th}$ & Average \\
\midrule
RoBERTa$_{full}$\ & $59.7_{3.5}$ & $\textbf{66.9}_{1.2}$ & $\textbf{69.2}_{1.8}$ & $\textbf{66.4}_{2.7}$  \\
SetFit$_{exp}$\  & $60.7_{1.3}$ & $66.3_{1.6}$ & $67.5_{0.9}$  & $65.9_{2.2}$  \\
\LaGoNNexp \ & $\textbf{61.5}_{1.7}$  & $66.4_{1.4}$  & $67.7_{0.9}$  & $66.1_{1.8}$  \\
\midrule
RoBERTa$_{freeze}$\ & $59.7_{3.5}$  & $60.4_{2.7}$  & $63.1_{2.3}$ & $61.0_{1.3}$  \\
$k$NN\     & $60.7_{1.3}$  & $59.6_{2.8}$ & $59.5_{2.5}$  & $59.5_{0.5}$   \\

SetFit \ & $60.7_{1.3}$  & $62.5_{0.7}$ & $63.4_{1.0}$  & $62.3_{1.0}$   \\
\LaGoNN \ & $\textbf{61.5}_{1.7}$  & $62.8_{1.5}$ & $64.2_{1.0}$  & $63.0_{0.9}$  \\
\midrule
Probe\   & $54.9_{1.4}$  & $58.5_{0.9}$ & $60.9_{0.4}$ & $58.7_{1.7}$  \\
\LaGoNNcheap\  & $54.2_{2.3}$  & $58.6_{0.6}$ & $60.6_{0.5}$ & $58.5_{1.8}$ \\
\bottomrule
\end{tabular}
\end{adjustbox}
\caption{\LaGoNN and \LaGoNNexp are the strongest performers on the first step, but are overtaken by RoBERTa$_{full}$ on later steps, which also is the strongest overall classifier. We note that \LaGoNN and \LaGoNNexp consistently outperform SetFit and SetFit$_{exp}$, respectively.}
\end{table}

\begin{table}[H]
\centering
\begin{adjustbox}{width=0.5\textwidth}
\begin{tabular}{l c c c c}
\textbf{Method}& & \textbf{LIAR} &  & \\
\textit{Extreme} & $1^{st}$ & $5^{th}$ & $10^{th}$ & Average \\
\midrule
RoBERTa$_{full}$\ & $\textbf{32.0}_{2.7}$ & $\textbf{34.7}_{2.9}$ & $35.1_{4.3}$ & $33.7_{1.0}$  \\
SetFit$_{exp}$\  & $31.2_{3.8}$ & $30.4_{3.1}$ & $31.8_{2.9}$  & $31.5_{0.7}$  \\
\LaGoNNexp \ & $30.6_{4.7}$  & $30.3_{2.0}$  & $31.3_{2.0}$  & $31.1_{0.6}$  \\
\midrule
RoBERTa$_{freeze}$\ & $\textbf{32.0}_{2.7}$  & $32.8_{4.5}$  & $34.2_{5.0}$ & $33.2_{0.7}$  \\
$k$NN\     & $27.0_{0.5}$  & $27.3_{0.8}$ & $27.9_{0.8}$  & $27.4_{0.3}$   \\

SetFit\ & $31.2_{3.8}$  & $33.7_{5.1}$ & $\textbf{35.7}_{5.1}$  & $\textbf{34.3}_{1.6}$   \\
\LaGoNN \ & $30.6_{4.7}$  & $32.0_{4.6}$ & $33.7_{5.4}$  & $32.6_{0.9}$  \\
\midrule
Probe\   & $30.7_{2.0}$  & $30.6_{3.9}$ & $31.7_{2.9}$ & $31.1_{0.4}$  \\
\LaGoNNcheap\  & $30.7_{2.0}$  & $30.5_{3.8}$ & $31.4_{2.6}$ & $31.0_{0.4}$ \\
\bottomrule
\end{tabular}
\end{adjustbox}
\caption{RoBERTa$_{freeze}$ and RoBERTa$_{full}$ start out as the strongest performers but are eventually overtaken by SetFit on the 10$^{th}$ step, and SetFit ends up being the strongest performer over all steps based on the average.}
\end{table}

\begin{table}[H]
\centering
\begin{adjustbox}{width=0.5\textwidth}
\begin{tabular}{l c c c c}
\textbf{Method}& & \textbf{LIAR} &  & \\
\textit{Imbalanced} & $1^{st}$ & $5^{th}$ & $10^{th}$ & Average \\
\midrule
RoBERTa$_{full}$\ & $31.4_{3.2}$ & $35.8_{2.6}$ & $\textbf{40.0}_{4.3}$ & $36.2_{2.4}$  \\
SetFit$_{exp}$\  & $\textbf{32.3}_{4.5}$ & $35.9_{3.1}$ & $36.4_{2.2}$  & $35.2_{1.1}$  \\
\LaGoNNexp \ & $\textbf{32.3}_{4.6}$  & $35.7_{3.4}$  & $36.5_{2.3}$  & $35.7_{1.4}$  \\
\midrule
RoBERTa$_{freeze}$\ & $31.4_{3.2}$  & $34.1_{2.6}$  & $35.6_{3.2}$ & $34.0_{1.4}$  \\
$k$NN\     & $27.0_{0.2}$  & $28.5_{1.0}$ & $29.0_{1.0}$  & $28.7_{0.7}$   \\

SetFit\ & $\textbf{32.3}_{4.5}$  & $\textbf{36.5}_{3.1}$ & $38.5_{3.4}$  & $\textbf{36.3}_{2.0}$   \\
\LaGoNN \ & $\textbf{32.3}_{4.6}$  & $34.9_{2.2}$ & $36.9_{2.5}$  & $35.3_{1.4}$  \\
\midrule
Probe\   & $30.7_{3.0}$  & $32.8_{1.8}$ & $35.0_{1.6}$ & $33.5_{1.5}$  \\
\LaGoNNcheap\  & $30.4_{3.0}$  & $32.9_{1.8}$ & $35.4_{1.7}$ & $33.5_{1.7}$ \\
\bottomrule
\end{tabular}
\end{adjustbox}
\caption{SetFit, SetFit$_{exp}$, \LaGoNN, and \LaGoNNexp start out as the strongest performers. On the 5$^{th}$ step, SetFit is overtaken the other systems, but is eventually overtaken by RoBERTa$_{full}$. Overall SetFit is the strongest system, but we note that \LaGoNNexp outperforms SetFit$_{exp}$.}
\end{table}

\begin{table}[H]
\centering
\begin{adjustbox}{width=0.5\textwidth}
\begin{tabular}{l c c c c}
\textbf{Method}& & \textbf{LIAR} &  & \\
\textit{Moderate} & $1^{st}$ & $5^{th}$ & $10^{th}$ & Average \\
\midrule
RoBERTa$_{full}$\ & $33.9_{3.1}$ & $38.4_{2.7}$ & $\textbf{43.9}_{2.2}$ & $\textbf{39.5}_{3.0}$  \\
SetFit$_{exp}$\  & $33.0_{2.6}$ & $37.2_{1.8}$ & $38.7_{1.5}$  & $37.4_{1.6}$  \\
\LaGoNNexp \ & $\textbf{34.1}_{3.4}$  & $\textbf{38.7}_{2.3}$  & $39.0_{1.8}$  & $37.8_{1.5}$  \\
\midrule
RoBERTa$_{freeze}$\ & $33.9_{3.1}$  & $35.3_{2.6}$  & $36.8_{2.2}$ & $35.4_{1.0}$  \\
$k$NN\     & $29.2_{0.8}$  & $29.7_{1.5}$ & $30.0_{0.6}$  & $29.8_{0.3}$   \\

SetFit\ & $33.0_{2.6}$  & $37.2_{3.9}$ & $39.4_{3.5}$  & $37.0_{1.8}$   \\
\LaGoNN \ & $\textbf{34.1}_{3.4}$  & $37.0_{3.1}$ & $38.6_{3.0}$  & $36.8_{1.3}$  \\
\midrule
Probe\   & $31.6_{1.1}$  & $34.7_{2.5}$ & $37.0_{2.5}$ & $34.9_{1.7}$  \\
\LaGoNNcheap\  & $31.4_{0.9}$  & $35.3_{2.3}$ & $37.6_{2.0}$ & $35.3_{1.9}$ \\
\bottomrule
\end{tabular}
\end{adjustbox}
\caption{\LaGoNN and \LaGoNNexp start out as the strongest performers and \LaGoNNexp continues to be strong, until the 10$^{th}$ step where it is overtaken by RoBERTa$_{full}$, which ends up as the most performant classifier over all steps based on the average.}
\end{table}

\begin{table}[H]
\centering
\begin{adjustbox}{width=0.5\textwidth}
\begin{tabular}{l c c c c}
\textbf{Method}& & \textbf{LIAR} &  & \\
\textit{Balanced} & $1^{st}$ & $5^{th}$ & $10^{th}$ & Average \\
\midrule
RoBERTa$_{full}$\ & $33.8_{2.1}$ & $\textbf{39.4}_{2.4}$ & $\textbf{43.5}_{1.7}$ & $\textbf{40.2}_{3.2}$  \\
SetFit$_{exp}$\  & $\textbf{34.4}_{2.3}$ & $36.7_{1.7}$ & $37.0_{1.3}$  & $36.5_{1.1}$  \\
\LaGoNNexp \ & $33.8_{1.8}$  & $34.2_{2.7}$  & $37.2_{1.9}$  & $36.2_{1.4}$  \\
\midrule
RoBERTa$_{freeze}$\ & $33.8_{2.1}$  & $36.6_{1.6}$  & $38.6_{1.5}$ & $36.7_{1.5}$  \\
$k$NN\     & $30.1_{0.4}$  & $31.3_{2.1}$ & $30.6_{1.1}$  & $30.9_{0.4}$   \\

SetFit \ & $\textbf{34.4}_{2.3}$  & $38.3_{2.5}$ & $40.0_{2.0}$  & $37.9_{1.6}$   \\
\LaGoNN \ & $33.8_{1.8}$  & $38.3_{1.3}$ & $40.6_{0.6}$  & $38.1_{2.0}$  \\
\midrule
Probe\   & $32.1_{1.9}$  & $35.2_{1.4}$ & $37.2_{2.5}$ & $35.2_{1.7}$  \\
\LaGoNNcheap\  & $31.9_{1.9}$  & $36.0_{1.0}$ & $37.5_{2.5}$ & $35.7_{1.8}$ \\
\bottomrule
\end{tabular}
\end{adjustbox}
\caption{SetFit and SetFit$_{exp}$ are the most performant systems on the first step, but are overtaken by RoBERTa$_{full}$, the strongest overall classifier. We note that \LaGoNN outperforms SetFit after the first step and in aggregate.}
\end{table}

\subsection{Additional results: secondary experiments}
\label{sec:appendixadditional2}
Here, we provide additional results from our second set of experiments that, due to space limitations, could not be included in the main text. We note that a version of \LaGoNN outperforms or has the same performance of all methods, including our upper bound RoBERTa$_{full}$, on 60\% of all displayed results, and is the best performer relative to Sentence Transformer-based methods on 65\%. This excludes \LaGoNNcheap. This method showed strong performance on the Insincere Questions dataset, but hurts performance in other cases.  
\indent In cases when SetFit-based methods do outperform our system, the performances are comparable, usually within one point, yet they can be quite different when \LaGoNN-based methods are the strongest. Below, we report the mean average precision $\times 100$ for all methods over five seeds with the standard deviation, except in the case of Hate Speech Offensive, where the evaluation measure is the macro-F1. Each table shows the results for a given dataset and a given label-balance distribution on the first, fifth, and tenth step followed by the average for all ten steps. In the table caption we provide a summary/interpretation of the results for a given setting. 
LIAR appears to be the most difficult dataset for all methods. This is expected because it likely does not include enough context to determine the truth of a statement.

\begin{figure*}[t]
    \centering
    \includegraphics[scale=0.35]{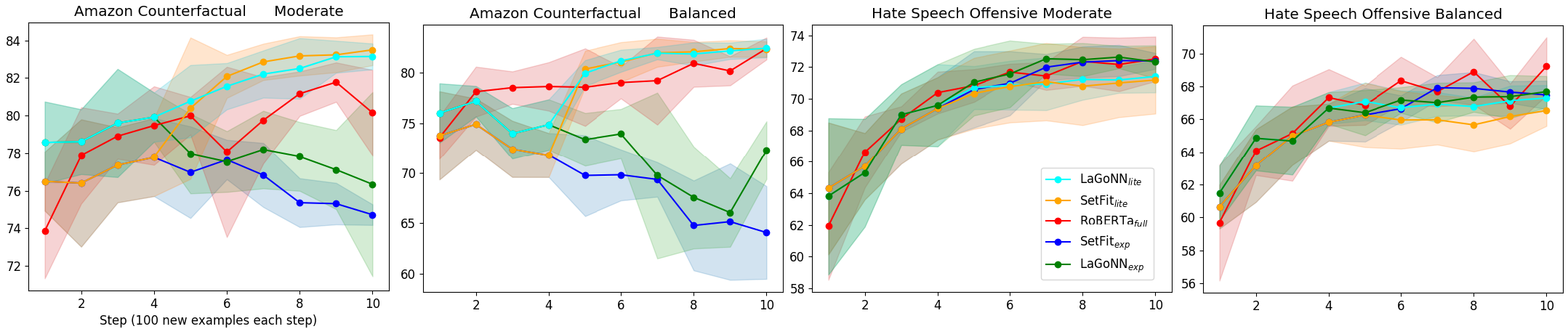}
    \captionof{figure}{Average performance for all the moderate and balanced sampling regimes on Amazon Counterfactual and Hate Speech Offensive. More expensive models, such as \LaGoNNexp, SetFit$_{exp}$, and RoBERTa$_{full}$ perform best when the label distribution is imbalanced. As the distribution becomes more balanced, inexpensive models, such as \LaGoNNlite, show similar or improved performance. The measure is average precision for Amazon Counterfactual and the macro F1 for Hate Speech Offensive. We only consider one neighbor for the \LaGoNN-based methods.}
    \label{LaGoNN_resultsmany2}
\end{figure*}

\begin{table}[H]
\centering
\begin{adjustbox}{width=0.5\textwidth}
\begin{tabular}{l c c c c}
\textbf{Method}& & \textbf{Insincere Questions} &  & \\
\textit{Extreme} & $1^{st}$ & $5^{th}$ & $10^{th}$ & Average \\
\midrule
RoBERTa$_{full}$\ & $19.9_{8.4}$ & $30.9_{7.9}$ & $42.0_{7.4}$ & $33.5_{6.7}$  \\
SetFit$_{exp}$\  & $24.1_{6.3}$ & $29.2_{6.7}$ & $36.7_{7.3}$  & $31.7_{3.4}$  \\
\LaGoNNexp \ & $\textbf{30.7}_{8.9}$  & $37.6_{6.1}$  & $39.0_{6.1}$  & $36.1_{2.3}$  \\
\midrule
SetFit$_{lite}$\ & $24.1_{6.3}$  & $38.1_{6.3}$ & $41.1_{6.5}$  & $35.6_{5.5}$   \\
\LaGoNNlite\  & $\textbf{30.7}_{8.9}$  & $\textbf{41.8}_{8.3}$ & $\textbf{43.4}_{8.5}$ & $\textbf{39.3}_{4.4}$ \\

\midrule
RoBERTa$_{freeze}$\ & $19.9_{8.4}$  & $34.1_{5.4}$  & $37.9_{5.2}$ & $32.5_{5.4}$  \\
$k$NN\     & $6.8_{0.4}$  & $15.9_{3.4}$ & $16.9_{4.3}$  & $14.4_{3.0}$   \\

SetFit\ & $24.1_{6.3}$  & $31.7_{4.9}$ & $36.1_{5.4}$  & $31.8_{3.6}$   \\
\LaGoNN \ & $\textbf{30.7}_{8.9}$  & $39.3_{4.9}$ & $41.2_{4.7}$  & $38.4_{3.0}$  \\
\midrule
Probe\   & $24.3_{8.4}$  & $39.8_{5.6}$ & $44.8_{4.2}$ & $38.3_{6.2}$  \\
\LaGoNNcheap\  & $23.6_{7.8}$  & $40.7_{5.9}$ & $45.3_{4.4}$ & $38.6_{6.6}$ \\
\bottomrule
\end{tabular}
\end{adjustbox}
\caption{\LaGoNN, \LaGoNNlite, and \LaGoNNexp start out as the strongest models, but \LaGoNNlite remains the most performant by the 10$^{th}$ step. It is also the overall strongest performer based on the average. We note the strength of \LaGoNNcheap relative to far more expensive methods.}
\end{table}

\begin{table}[H]
\centering
\begin{adjustbox}{width=0.5\textwidth}
\begin{tabular}{l c c c c}
\textbf{Method}& & \textbf{Insincere Questions} &  & \\
\textit{Imbalanced} & $1^{st}$ & $5^{th}$ & $10^{th}$ & Average \\
\midrule
RoBERTa$_{full}$\ & $39.8_{5.5}$ & $53.1_{4.6}$ & $55.7_{1.2}$ & $50.6_{4.4}$  \\
SetFit$_{exp}$\  & $43.7_{2.7}$ & $52.2_{1.9}$ & $53.8_{0.9}$  & $51.4_{2.9}$  \\
\LaGoNNexp \ & $\textbf{44.5}_{4.5}$  & $52.7_{2.4}$  & $55.4_{2.0}$  & $51.8_{3.0}$  \\
\midrule
SetFit$_{lite}$\ & $43.7_{2.7}$  & $52.9_{2.6}$ & $55.8_{1.8}$  & $52.2_{3.4}$   \\
\LaGoNNlite\  & $\textbf{44.5}_{4.5}$  & $\textbf{53.5}_{2.7}$ & $\textbf{55.9}_{2.4}$ & $\textbf{52.6}_{3.5}$ \\

\midrule
RoBERTa$_{freeze}$\ & $39.8_{5.5}$  & $44.1_{3.6}$  & $46.3_{2.4}$ & $44.0_{2.0}$  \\
$k$NN\     & $23.9_{2.2}$  & $30.3_{3.0}$ & $31.6_{2.4}$  & $30.0_{2.1}$   \\

SetFit \ & $43.7_{2.7}$  & $47.6_{1.6}$ & $50.1_{2.1}$  & $47.6_{1.8}$   \\
\LaGoNN \ & $\textbf{44.5}_{4.5}$  & $48.1_{2.2}$ & $50.3_{1.7}$  & $48.1_{1.9}$  \\
\midrule
Probe\   & $40.4_{4.2}$  & $49.4_{2.3}$ & $52.3_{1.7}$ & $49.0_{3.3}$  \\
\LaGoNNcheap\  & $40.8_{4.3}$  & $51.1_{2.4}$ & $54.5_{1.4}$ & $50.4_{4.0}$ \\
\bottomrule
\end{tabular}
\end{adjustbox}
\caption{\LaGoNN, \LaGoNNlite, and \LaGoNNexp start out as the strongest models, but \LaGoNNlite remains the most performant by the 10$^{th}$ step. It is also the overall strongest performer based on the average. We note the strength of \LaGoNNcheap relative to far more expensive methods.}
\end{table}

\begin{table}[H]
\centering
\begin{adjustbox}{width=0.5\textwidth}
\begin{tabular}{l c c c c}
\textbf{Method}& & \textbf{Insincere Questions} &  & \\
\textit{Moderate} & $1^{st}$ & $5^{th}$ & $10^{th}$ & Average \\
\midrule
RoBERTa$_{full}$\ & $48.1_{2.3}$ & $54.7_{1.9}$ & $57.5_{1.5}$ & $53.9_{2.9}$  \\
SetFit$_{exp}$\  & $48.9_{1.7}$ & $53.9_{0.7}$ & $54.2_{1.5}$  & $52.3_{1.6}$  \\
\LaGoNNexp \ & $\textbf{49.8}_{1.6}$  & $52.2_{1.9}$  & $53.2_{3.3}$  & $52.0_{1.4}$  \\
\midrule
SetFit$_{lite}$\ & $48.9_{1.7}$  & $\textbf{56.5}_{1.4}$ & $\textbf{58.7}_{0.6}$  & $\textbf{55.0}_{3.5}$   \\
\LaGoNNlite\  & $\textbf{49.8}_{1.6}$  & $56.1_{2.8}$ & $58.3_{1.5}$ & $54.6_{3.5}$ \\

\midrule
RoBERTa$_{freeze}$\ & $48.1_{2.3}$  & $50.2_{2.2}$  & $52.0_{1.4}$ & $50.2_{1.4}$  \\
$k$NN\     & $28.0_{2.4}$  & $33.9_{2.8}$ & $33.6_{2.0}$  & $33.5_{1.9}$   \\

SetFit\ & $48.9_{1.7}$  & $53.6_{1.9}$ & $55.8_{1.7}$  & $53.3_{2.2}$   \\
\LaGoNN \ & $\textbf{49.8}_{1.6}$  & $54.4_{1.3}$ & $56.9_{0.5}$  & $54.2_{2.2}$  \\
\midrule
Probe\   & $45.7_{2.1}$  & $52.3_{1.8}$ & $54.4_{1.1}$ & $51.4_{2.5}$  \\
\LaGoNNcheap\  & $45.7_{2.2}$  & $54.4_{1.6}$ & $56.4_{0.6}$ & $53.2_{3.2}$ \\
\bottomrule
\end{tabular}
\end{adjustbox}
\caption{\LaGoNN, \LaGoNNlite, and \LaGoNNexp start out as the strongest models, but SetFit$_{lite}$ overtakes the other methods by the 5$^{th}$ step and is the strongest performer based on the average. We note the strength of \LaGoNNcheap relative to far more expensive methods.}
\end{table}

\begin{table}[H]
\centering
\begin{adjustbox}{width=0.5\textwidth}
\begin{tabular}{l c c c c}
\textbf{Method}& & \textbf{Insincere Questions} &  & \\
\textit{Balanced} & $1^{st}$ & $5^{th}$ & $10^{th}$ & Average \\
\midrule
RoBERTa$_{full}$\ & $47.1_{4.2}$ & $52.1_{3.6}$ & $55.7_{2.6}$ & $52.5_{2.9}$  \\
SetFit$_{exp}$\  & $43.5_{4.2}$ & $47.1_{4.6}$ & $48.5_{3.9}$  & $48.0_{1.7}$  \\
\LaGoNNexp \ & $42.8_{5.3}$  & $47.6_{2.9}$  & $47.0_{1.7}$  & $46.2_{2.0}$  \\
\midrule
SetFit$_{lite}$\ & $43.5_{4.2}$  & $\textbf{54.6}_{2.4}$ & $\textbf{59.6}_{0.9}$  & $53.6_{5.8}$   \\
\LaGoNNlite\  & $42.8_{5.3}$  & $53.5_{3.7}$ & $58.6_{2.5}$ & $52.2_{6.4}$ \\

\midrule
RoBERTa$_{freeze}$\ & $47.1_{4.2}$  & $52.1_{0.4}$  & $53.3_{1.1}$ & $51.5_{2.1}$  \\
$k$NN\     & $22.3_{2.3}$  & $30.2_{2.3}$ & $30.9_{1.8}$  & $29.5_{2.5}$   \\

SetFit\ & $43.5_{4.2}$  & $53.8_{2.2}$ & $55.5_{1.6}$  & $52.8_{3.5}$   \\
\LaGoNN \ & $42.8_{5.3}$  & $54.1_{2.9}$ & $56.3_{1.3}$  & $53.4_{3.7}$  \\
\midrule
Probe\   & $47.5_{1.6}$  & $52.4_{1.7}$ & $55.3_{1.1}$ & $52.2_{2.5}$  \\
\LaGoNNcheap\  & $\textbf{49.3}_{2.6}$  & $54.4_{1.4}$ & $57.6_{0.7}$ & $\textbf{54.2}_{2.7}$ \\
\bottomrule
\end{tabular}
\end{adjustbox}
\caption{\LaGoNNcheap, starts out as the strongest model, but SetFit$_{lite}$ overtakes the other methods on the 5$^{th}$ and 10$^{th}$ step. Overall \LaGoNNcheap is the strongest model despite being one of the least expensive.}
\end{table}

\begin{table}[H]
\centering
\begin{adjustbox}{width=0.5\textwidth}
\begin{tabular}{l c c c c}
\textbf{Method}& & \textbf{Amazon Counterfactual} &  & \\
\textit{Extreme} & $1^{st}$ & $5^{th}$ & $10^{th}$ & Average \\
\midrule
RoBERTa$_{full}$\ & $21.8_{6.6}$ & $63.9_{10.2}$ & $72.3_{3.0}$ & $59.6_{16.8}$  \\
SetFit$_{exp}$\  & $22.3_{8.8}$ & $64.2_{3.3}$ & $68.6_{4.6}$  & $56.8_{14.9}$  \\
\LaGoNNexp \ & $\textbf{26.1}_{17.5}$  & $\textbf{68.4}_{4.4}$  & $\textbf{74.9}_{2.9}$  & $\textbf{63.2}_{16.7}$  \\
\midrule
SetFit$_{lite}$\ & $22.3_{8.8}$  & $62.4_{5.1}$ & $67.5_{5.2}$  & $56.5_{14.7}$   \\
\LaGoNNlite\  & $\textbf{26.1}_{17.5}$  & $68.3_{4.3}$ & $68.9_{4.3}$ & $60.6_{15.1}$ \\

\midrule
RoBERTa$_{freeze}$\ & $21.8_{6.6}$  & $41.0_{12.7}$  & $51.3_{10.7}$ & $40.6_{8.9}$  \\
$k$NN\     & $10.3_{0.2}$  & $15.3_{4.2}$ & $18.4_{3.7}$  & $15.6_{2.4}$   \\

SetFit\ & $22.3_{8.8}$  & $32.4_{11.5}$ & $42.3_{8.8}$  & $34.5_{5.9}$   \\
\LaGoNN \ & $\textbf{26.1}_{17.5}$  & $31.1_{19.4}$ & $33.0_{19.1}$  & $30.9_{2.3}$  \\
\midrule
Probe\   & $24.2_{9.0}$  & $46.3_{4.4}$ & $54.6_{2.0}$ & $45.1_{10.3}$  \\
\LaGoNNcheap\  & $20.1_{6.9}$  & $38.3_{4.9}$ & $47.8_{3.4}$ & $38.2_{9.5}$ \\
\bottomrule
\end{tabular}
\end{adjustbox}
\caption{\LaGoNN, \LaGoNNlite, and \LaGoNNexp  are the most performant models on the first step, but only \LaGoNNexp remains the most performant on subsequent steps, also being the strongest overall method based on the average over all steps.}
\end{table}

\begin{table}[H]
\centering
\begin{adjustbox}{width=0.5\textwidth}
\begin{tabular}{l c c c c}
\textbf{Method}& & \textbf{Amazon Counterfactual} &  & \\
\textit{Imbalanced} & $1^{st}$ & $5^{th}$ & $10^{th}$ & Average \\
\midrule
RoBERTa$_{full}$\ & $68.2_{4.5}$ & $\textbf{81.0}_{1.7}$ & $\textbf{82.2}_{1.0}$ & $79.2_{3.9}$  \\
SetFit$_{exp}$\  & $72.0_{2.1}$ & $78.4_{2.8}$ & $78.8_{1.2}$  & $78.0_{2.1}$  \\
\LaGoNNexp \ & $\textbf{74.3}_{3.8}$  & $80.1_{1.4}$  & $79.0_{1.6}$  & $79.5_{1.9}$  \\
\midrule
SetFit$_{lite}$\ & $72.0_{2.1}$  & $79.1_{1.4}$ & $81.6_{1.3}$  & $79.1_{2.7}$   \\
\LaGoNNlite\  & $\textbf{74.3}_{3.8}$  & $79.2_{1.7}$ & $81.9_{1.1}$ & $\textbf{80.2}_{2.2}$ \\

\midrule
RoBERTa$_{freeze}$\ & $68.2_{4.5}$  & $75.0_{2.2}$  & $77.0_{2.4}$ & $74.2_{2.6}$  \\
$k$NN\     & $51.0_{4.1}$  & $60.0_{3.1}$ & $61.3_{2.1}$  & $59.7_{3.0}$   \\

SetFit\ & $72.0_{2.1}$  & $74.4_{2.3}$ & $76.7_{1.8}$  & $74.8_{1.4}$   \\
\LaGoNN \ & $\textbf{74.3}_{3.8}$  & $76.1_{3.6}$ & $77.3_{3.2}$  & $76.1_{1.0}$  \\
\midrule
Probe\   & $46.6_{2.8}$  & $60.3_{1.4}$ & $64.2_{1.2}$ & $59.2_{5.2}$  \\
\LaGoNNcheap\  & $38.2_{3.2}$  & $55.3_{1.8}$ & $61.0_{1.2}$ & $54.4_{6.7}$ \\
\bottomrule
\end{tabular}
\end{adjustbox}
\caption{On the first step, \LaGoNN, \LaGoNNlite, and \LaGoNNexp start out the strongest but \LaGoNNlite performs slightly worse than RoBERTa$_{full}$ on the 5$^{th}$ and 10$^{th}$ step. However, \LaGoNNlite is the best overall method based on the average.}
\end{table}

\begin{table}[H]
\centering
\begin{adjustbox}{width=0.5\textwidth}
\begin{tabular}{l c c c c}
\textbf{Method}& & \textbf{Amazon Counterfactual} &  & \\
\textit{Moderate} & $1^{st}$ & $5^{th}$ & $10^{th}$ & Average \\
\midrule
RoBERTa$_{full}$\ & $73.9_{2.5}$ & $80.0_{1.0}$ & $80.1_{2.3}$ & $79.1_{2.1}$  \\
SetFit$_{exp}$\  & $76.5_{1.6}$ & $77.0_{2.4}$ & $74.7_{0.5}$  & $76.5_{1.0}$  \\
\LaGoNNexp \ & $\textbf{78.6}_{2.2}$  & $78.0_{2.1}$  & $76.3_{4.9}$  & $78.2_{1.0}$  \\
\midrule
SetFit$_{lite}$\ & $76.5_{1.6}$  & $80.4_{3.8}$ & $\textbf{83.5}_{0.8}$  & $80.3_{2.8}$   \\
\LaGoNNlite\  & $\textbf{78.6}_{2.2}$  & $80.8_{1.9}$ & $83.1_{0.7}$ & $\textbf{81.0}_{1.7}$ \\

\midrule
RoBERTa$_{freeze}$\ & $73.9_{2.5}$  & $76.6_{1.4}$  & $78.5_{0.7}$ & $76.4_{1.7}$  \\
$k$NN\     & $54.5_{3.1}$  & $64.2_{1.9}$ & $66.6_{1.3}$  & $64.7_{3.5}$   \\

SetFit \ & $76.5_{1.6}$  & $80.6_{0.5}$ & $81.2_{0.3}$  & $80.0_{1.4}$   \\
\LaGoNN \ & $\textbf{78.6}_{2.2}$  & $\textbf{81.2}_{1.4}$ & $81.6_{1.1}$  & $80.8_{0.9}$  \\
\midrule
Probe\   & $52.3_{2.0}$  & $64.1_{1.8}$ & $67.2_{1.4}$ & $63.1_{4.3}$  \\
\LaGoNNcheap\  & $47.3_{3.4}$  & $60.7_{1.5}$ & $65.2_{1.4}$ & $59.5_{5.2}$ \\
\bottomrule
\end{tabular}
\end{adjustbox}
\caption{On the first step, \LaGoNN, \LaGoNNlite, and \LaGoNNexp start out the strongest. On the 5$^{th}$ step, \LaGoNN is the most performant method while on the 10$^{th}$ step it is SetFit$_{lite}$. However, \LaGoNNlite is the best overall method based on the average.}
\end{table}

\begin{table}[H]
\centering
\begin{adjustbox}{width=0.5\textwidth}
\begin{tabular}{l c c c c}
\textbf{Method}& & \textbf{Amazon Counterfactual} &  & \\
\textit{Balanced} & $1^{st}$ & $5^{th}$ & $10^{th}$ & Average \\
\midrule
RoBERTa$_{full}$\ & $73.6_{2.1}$ & $78.6_{3.9}$ & $82.4_{1.1}$ & $78.9_{2.2}$  \\
SetFit$_{exp}$\  & $73.8_{4.4}$ & $69.8_{4.0}$ & $64.1_{4.6}$  & $69.6_{3.6}$  \\
\LaGoNNexp \ & $\textbf{76.0}_{3.0}$  & $73.4_{2.6}$  & $72.3_{2.9}$  & $72.5_{3.4}$  \\
\midrule
SetFit$_{lite}$\ & $73.8_{4.4}$  & $\textbf{80.4}_{1.8}$ & $82.4_{0.8}$  & $78.3_{4.3}$   \\
\LaGoNNlite\  & $\textbf{76.0}_{3.0}$  & $80.0_{1.3}$ & $\textbf{82.5}_{0.9}$ & $79.2_{3.2}$ \\

\midrule
RoBERTa$_{freeze}$\ & $73.6_{2.1}$  & $76.8_{1.6}$  & $77.9_{1.0}$ & $76.5_{1.3}$  \\
$k$NN\     & $41.7_{3.4}$  & $57.9_{3.3}$ & $58.3_{3.3}$  & $56.8_{5.1}$   \\

SetFit\ & $73.8_{4.4}$  & $79.2_{1.9}$ & $80.1_{1.0}$  & $78.6_{1.8}$   \\
\LaGoNN \ & $\textbf{76.0}_{3.0}$  & $80.1_{2.0}$ & $81.4_{1.1}$  & $\textbf{79.8}_{1.4}$  \\
\midrule
Probe\   & $52.4_{3.4}$  & $64.7_{2.5}$ & $67.5_{0.4}$ & $63.4_{4.4}$  \\
\LaGoNNcheap\  & $48.1_{3.4}$  & $62.0_{2.0}$ & $65.3_{0.8}$ & $60.5_{5.0}$ \\
\bottomrule
\end{tabular}
\end{adjustbox}
\caption{On the first step, \LaGoNN, \LaGoNNlite, and \LaGoNNexp start out the strongest. On the 5$^{th}$ step,  SetFit$_{lite}$ pulls ahead slightly, yet on the 10$^{th}$ step \LaGoNNlite is the best performer. Overall, \LaGoNN is the best method based on the average.}
\end{table}

\begin{table}[H]
\centering
\begin{adjustbox}{width=0.5\textwidth}
\begin{tabular}{l c c c c}
\textbf{Method}& & \textbf{Toxic Conversations} &  & \\
\textit{Extreme} & $1^{st}$ & $5^{th}$ & $10^{th}$ & Average \\
\midrule
RoBERTa$_{full}$\ & $7.9_{0.5}$ & $21.2_{3.7}$ & $33.8_{5.5}$ & $21.9_{9.3}$  \\
SetFit$_{exp}$\  & $8.8_{1.2}$ & $18.1_{3.4}$ & $24.7_{4.1}$  & $17.6_{5.5}$  \\
\LaGoNNexp \ & $8.9_{1.7}$  & $17.4_{6.6}$  & $26.4_{5.2}$  & $17.9_{6.0}$  \\
\midrule
SetFit$_{lite}$\ & $8.8_{1.2}$  & $15.9_{4.8}$ & $18.0_{3.9}$  & $14.9_{3.2}$   \\
\LaGoNNlite\  & $8.9_{1.7}$  & $16.1_{5.9}$ & $19.8_{6.0}$ & $15.5_{3.7}$ \\

\midrule
RoBERTa$_{freeze}$\ & $7.9_{0.5}$  & $12.8_{2.4}$  & $19.1_{3.2}$ & $13.5_{3.5}$  \\
$k$NN\     & $7.9_{0.0}$  & $8.7_{0.4}$ & $8.7_{0.2}$  & $8.5_{0.3}$   \\

SetFit\ & $8.8_{1.2}$  & $13.1_{2.5}$ & $16.3_{3.0}$  & $13.0_{2.6}$   \\
\LaGoNN \ & $8.9_{1.7}$  & $13.8_{3.9}$ & $17.1_{4.8}$  & $13.4_{2.6}$  \\
\midrule
Probe\   & $\textbf{13.1}_{2.8}$  & $\textbf{24.6}_{2.6}$ & $\textbf{30.1}_{2.1}$ & $\textbf{23.9}_{5.6}$  \\
\LaGoNNcheap\  & $11.3_{2.2}$  & $21.7_{2.7}$ & $27.4_{2.3}$ & $21.3_{5.3}$ \\
\bottomrule
\end{tabular}
\end{adjustbox}
\caption{Probe is most performant method on all steps and the overall strongest performer. We note, however, that \LaGoNN-based methods tend to outperform their SetFit-based counterparts.}
\end{table}

\begin{table}[H]
\centering
\begin{adjustbox}{width=0.5\textwidth}
\begin{tabular}{l c c c c}
\textbf{Method}& & \textbf{Toxic Conversations} &  & \\
\textit{Imbalanced} & $1^{st}$ & $5^{th}$ & $10^{th}$ & Average \\
\midrule
RoBERTa$_{full}$\ & $\textbf{24.1}_{5.6}$ & $43.1_{3.4}$ & $52.1_{2.5}$ & $42.4_{8.2}$  \\
SetFit$_{exp}$\  & $21.8_{6.6}$ & $44.5_{4.1}$ & $51.4_{1.9}$  & $42.1_{9.3}$  \\
\LaGoNNexp \ & $22.7_{9.8}$  & $\textbf{49.1}_{5.6}$  & $\textbf{53.4}_{2.3}$  & $\textbf{45.6}_{9.8}$  \\
\midrule
SetFit$_{lite}$\ & $21.8_{6.6}$  & $41.4_{4.4}$ & $44.8_{3.1}$  & $39.0_{7.0}$   \\
\LaGoNNlite\  & $22.7_{9.8}$  & $47.0_{6.3}$ & $50.2_{5.4}$ & $43.7_{8.6}$ \\

\midrule
RoBERTa$_{freeze}$\ & $\textbf{24.1}_{5.6}$  & $31.2_{4.4}$  & $34.0_{4.0}$ & $30.5_{3.1}$  \\
$k$NN\     & $11.5_{2.5}$  & $14.7_{4.0}$ & $15.3_{3.2}$  & $14.6_{1.1}$   \\

SetFit \ & $21.8_{6.6}$  & $26.7_{5.3}$ & $30.2_{4.0}$  & $26.6_{2.7}$   \\
\LaGoNN \ & $22.7_{9.8}$  & $27.6_{8.9}$ & $30.3_{8.7}$  & $27.4_{2.4}$  \\
\midrule
Probe\   & $23.3_{2.7}$  & $33.0_{2.8}$ & $37.1_{1.8}$ & $32.5_{4.2}$  \\
\LaGoNNcheap\  & $20.5_{3.2}$  & $31.1_{3.2}$ & $35.6_{1.8}$ & $30.5_{4.6}$ \\
\bottomrule
\end{tabular}
\end{adjustbox}
\caption{RoBERTa$_{full}$ and RoBERTa$_{freeze}$ start out as the strongest classifiers on the first step, but are overtaken on subsequent steps by \LaGoNNexp, which ends up as strongest method overall.}
\end{table}

\begin{table}[H]
\centering
\begin{adjustbox}{width=0.5\textwidth}
\begin{tabular}{l c c c c}
\textbf{Method}& & \textbf{Toxic Conversations} &  & \\
\textit{Moderate} & $1^{st}$ & $5^{th}$ & $10^{th}$ & Average \\
\midrule
RoBERTa$_{full}$\ & $34.2_{3.4}$ & $45.5_{1.9}$ & $52.4_{3.3}$ & $45.7_{5.6}$  \\
SetFit$_{exp}$\  & $33.6_{2.9}$ & $47.2_{2.2}$ & $46.6_{3.3}$  & $44.3_{4.3}$  \\
\LaGoNNexp \ & $\textbf{36.6}_{4.2}$  & $48.2_{2.7}$  & $49.9_{3.7}$  & $48.0_{4.4}$  \\
\midrule
SetFit$_{lite}$\ & $33.6_{2.9}$  & $52.6_{2.0}$ & $55.1_{1.6}$  & $48.8_{7.3}$   \\
\LaGoNNlite\  & $\textbf{36.6}_{4.2}$  & $\textbf{56.1}_{1.5}$ & $\textbf{57.7}_{1.4}$ & $\textbf{52.3}_{6.8}$ \\

\midrule
RoBERTa$_{freeze}$\ & $34.2_{3.4}$  & $38.4_{2.1}$  & $39.5_{1.8}$ & $38.0_{1.5}$  \\
$k$NN\     & $19.4_{1.9}$  & $21.5_{3.4}$ & $22.4_{2.9}$  & $21.6_{0.8}$   \\

SetFit\ & $33.6_{2.9}$  & $39.2_{2.9}$ & $41.6_{2.7}$  & $38.6_{2.4}$   \\
\LaGoNN \ & $\textbf{36.6}_{4.2}$  & $42.7_{3.7}$ & $45.0_{3.5}$  & $42.0_{2.5}$  \\
\midrule
Probe\   & $29.0_{2.7}$  & $36.1_{1.2}$ & $39.1_{1.5}$ & $35.5_{3.3}$  \\
\LaGoNNcheap\  & $26.1_{2.7}$  & $34.3_{1.3}$ & $37.5_{1.8}$ & $33.6_{3.6}$ \\
\bottomrule
\end{tabular}
\end{adjustbox}
\caption{On the first step, \LaGoNN, \LaGoNNlite, and \LaGoNNexp start out the strongest, but it is \LaGoNNlite that remains performant for all other steps. \LaGoNNlite is also the strongest overall method based on the average.}
\end{table}

\begin{table}[H]
\centering
\begin{adjustbox}{width=0.5\textwidth}
\begin{tabular}{l c c c c}
\textbf{Method}& & \textbf{Toxic Conversations} &  & \\
\textit{Balanced} & $1^{st}$ & $5^{th}$ & $10^{th}$ & Average \\
\midrule
RoBERTa$_{full}$\ & $32.3_{1.1}$ & $42.7_{1.8}$ & $54.1_{3.4}$ & $43.8_{6.3}$  \\
SetFit$_{exp}$\  & $35.7_{3.4}$ & $32.6_{6.2}$ & $37.4_{2.7}$  & $36.5_{1.9}$  \\
\LaGoNNexp \ & $\textbf{40.4}_{4.4}$  & $40.2_{6.6}$  & $39.8_{7.5}$  & $40.0_{1.2}$  \\
\midrule
SetFit$_{lite}$\ & $35.7_{3.4}$  & $52.7_{2.5}$ & $53.9_{2.2}$  & $46.8_{7.8}$   \\
\LaGoNNlite\  & $\textbf{40.4}_{4.4}$  & $\textbf{52.9}_{2.6}$ & $\textbf{54.0}_{2.3}$ & $\textbf{48.3}_{6.4}$ \\

\midrule
RoBERTa$_{freeze}$\ & $32.3_{1.1}$  & $39.2_{1.5}$  & $41.0_{0.6}$ & $38.5_{2.4}$  \\
$k$NN\     & $17.4_{0.8}$  & $23.7_{2.6}$ & $24.3_{2.7}$  & $23.1_{2.0}$   \\

SetFit\ & $35.7_{3.4}$  & $44.5_{2.9}$ & $46.1_{2.8}$  & $43.6_{2.9}$   \\
\LaGoNN \ & $\textbf{40.4}_{4.4}$  & $46.6_{2.7}$ & $48.1_{2.2}$  & $46.1_{2.2}$  \\
\midrule
Probe\   & $29.5_{2.4}$  & $35.9_{0.9}$ & $40.2_{0.9}$ & $36.1_{3.5}$  \\
\LaGoNNcheap\  & $26.8_{2.7}$  & $34.5_{1.3}$ & $38.5_{0.8}$ & $34.4_{3.7}$ \\
\bottomrule
\end{tabular}
\end{adjustbox}
\caption{On the first step, \LaGoNN, \LaGoNNlite, and \LaGoNNexp start out the strongest, but it is \LaGoNNlite that remains performant for all other steps. \LaGoNNlite is also the strongest overall method based on the average.}
\end{table}

\begin{table}[H]
\centering
\begin{adjustbox}{width=0.5\textwidth}
\begin{tabular}{l c c c c}
\textbf{Method}& & \textbf{Hate Speech Offensive} &  & \\
\textit{Extreme} & $1^{st}$ & $5^{th}$ & $10^{th}$ & Average \\
\midrule
RoBERTa$_{full}$\ & $30.2_{1.4}$ & $43.5_{2.5}$ & $\textbf{51.2}_{2.2}$ & $\textbf{44.3}_{7.4}$  \\
SetFit$_{exp}$\  & $30.3_{0.8}$ & $\textbf{44.0}_{1.3}$ & $51.1_{2.0}$  & $43.8_{6.5}$  \\
\LaGoNNexp \ & $30.3_{0.7}$  & $40.7_{2.9}$  & $49.1_{4.4}$  & $42.2_{6.2}$  \\
\midrule
SetFit$_{lite}$\ & $30.3_{0.8}$  & $43.4_{2.5}$ & $45.5_{3.4}$  & $41.6_{4.6}$   \\
\LaGoNNlite\  & $30.3_{0.7}$  & $40.9_{3.4}$ & $41.5_{4.8}$ & $39.1_{3.6}$ \\

\midrule
RoBERTa$_{freeze}$\ & $30.2_{1.4}$  & $33.5_{3.1}$  & $34.4_{3.4}$ & $33.1_{1.4}$  \\
$k$NN\     & $\textbf{31.5}_{1.2}$  & $35.9_{2.7}$ & $37.4_{2.0}$  & $35.8_{1.7}$   \\

SetFit\ & $30.3_{0.8}$  & $38.4_{2.5}$ & $41.1_{1.5}$  & $37.8_{3.3}$   \\
\LaGoNN \ & $30.3_{0.7}$  & $35.7_{2.6}$ & $39.1_{2.4}$  & $35.6_{2.7}$  \\
\midrule
Probe\   & $29.0_{0.2}$  & $34.7_{1.5}$ & $40.1_{2.1}$ & $35.1_{3.8}$  \\
\LaGoNNcheap\  & $29.0_{0.1}$  & $36.9_{1.8}$ & $40.5_{2.1}$ & $36.2_{3.7}$ \\
\bottomrule
\end{tabular}
\end{adjustbox}
\caption{$k$NN is the strongest method at first, but is overtaken by SetFit$_{exp}$ on the $5^{th}$ step, which is then overtaken by RoBERTa$_{full}$ on the $10^{th}$ step. RoBERTa$_{full}$ is overall most performant system based on the average.}
\end{table}

\begin{table}[H]
\centering
\begin{adjustbox}{width=0.5\textwidth}
\begin{tabular}{l c c c c}
\textbf{Method}& & \textbf{Hate Speech Offensive} &  & \\
\textit{Imbalanced} & $1^{st}$ & $5^{th}$ & $10^{th}$ & Average \\
\midrule
RoBERTa$_{full}$\ & $50.6_{3.0}$ & $65.2_{3.9}$ & $\textbf{70.3}_{1.2}$ & $64.2_{5.3}$  \\
SetFit$_{exp}$\  & $54.4_{4.3}$ & $66.3_{1.8}$ & $68.9_{2.0}$  & $64.3_{4.5}$  \\
\LaGoNNexp \ & $\textbf{57.0}_{5.2}$  & $\textbf{67.0}_{4.4}$  & $69.8_{2.1}$  & $\textbf{64.9}_{4.6}$  \\
\midrule
SetFit$_{lite}$\ & $54.4_{4.3}$  & $65.5_{3.0}$ & $65.9_{3.5}$  & $63.5_{3.9}$   \\
\LaGoNNlite\  & $\textbf{57.0}_{5.2}$  & $66.6_{2.6}$ & $66.6_{1.9}$ & $64.3_{4.1}$ \\

\midrule
RoBERTa$_{freeze}$\ & $50.6_{3.0}$  & $54.1_{1.6}$  & $55.3_{2.3}$ & $54.1_{1.3}$  \\
$k$NN\     & $55.6_{4.8}$  & $57.3_{2.3}$ & $58.8_{3.6}$  & $57.4_{1.1}$   \\

SetFit \ & $54.4_{4.3}$  & $57.0_{3.9}$ & $58.2_{3.8}$  & $57.2_{1.1}$   \\
\LaGoNN \ & $\textbf{57.0}_{5.2}$  & $58.2_{4.1}$ & $58.3_{3.4}$  & $58.3_{0.6}$  \\
\midrule
Probe\   & $46.5_{2.2}$  & $57.8_{1.7}$ & $60.3_{1.2}$ & $56.5_{4.5}$  \\
\LaGoNNcheap\  & $47.1_{1.3}$  & $56.5_{2.2}$ & $59.5_{2.5}$ & $55.6_{3.8}$ \\
\bottomrule
\end{tabular}
\end{adjustbox}
\caption{On the first step, \LaGoNN, \LaGoNNlite, and \LaGoNNexp start out the strongest, and \LaGoNNexp continues to be performant, but is overtaken on the $10^{th}$ step by RoBERTa$_{full}$. \LaGoNNexp is the strongest overall method based on the average.}
\end{table}

\begin{table}[H]
\centering
\begin{adjustbox}{width=0.5\textwidth}
\begin{tabular}{l c c c c}
\textbf{Method}& & \textbf{Hate Speech Offensive} &  & \\
\textit{Moderate} & $1^{st}$ & $5^{th}$ & $10^{th}$ & Average \\
\midrule
RoBERTa$_{full}$\ & $61.9_{3.4}$ & $70.8_{1.0}$ & $\textbf{72.5}_{1.4}$ & $69.9_{3.2}$  \\
SetFit$_{exp}$\  & $\textbf{64.3}_{4.2}$ & $70.6_{2.4}$ & $72.4_{0.5}$  & $69.8_{2.8}$  \\
\LaGoNNexp \ & $63.8_{4.9}$  & $\textbf{71.0}_{2.1}$  & $72.3_{1.0}$  & $\textbf{70.0}_{3.0}$  \\
\midrule
SetFit$_{lite}$\ & $64.3_{4.2}$  & $70.3_{2.2}$ & $71.2_{2.1}$  & $69.3_{2.3}$   \\
\LaGoNNlite\  & $63.8_{4.9}$  & $70.7_{1.4}$ & $71.4_{1.0}$ & $69.4_{2.5}$ \\

\midrule
RoBERTa$_{freeze}$\ & $61.9_{3.4}$  & $63.2_{4.1}$  & $64.1_{4.5}$ & $63.2_{0.6}$  \\
$k$NN\     & $\textbf{64.3}_{4.0}$  & $63.3_{2.9}$ & $63.9_{2.5}$  & $63.7_{0.4}$   \\

SetFit \ & $\textbf{64.3}_{4.2}$  & $67.3_{3.2}$ & $67.6_{2.3}$  & $66.9_{1.1}$   \\
\LaGoNN \ & $63.8_{4.9}$  & $65.0_{5.3}$ & $66.7_{5.9}$  & $65.3_{0.9}$  \\
\midrule
Probe\   & $55.6_{1.7}$  & $63.8_{0.8}$ & $66.1_{0.3}$ & $63.2_{3.0}$  \\
\LaGoNNcheap\  & $56.0_{3.6}$  & $62.2_{1.4}$ & $66.0_{0.9}$ & $62.3_{2.9}$ \\
\bottomrule
\end{tabular}
\end{adjustbox}
\caption{Similar to the imbalanced setting, on the first step, \LaGoNN, \LaGoNNlite, and \LaGoNNexp start out the strongest, and \LaGoNNexp continues to be performant, but is overtaken on the $10^{th}$ step by RoBERTa$_{full}$. \LaGoNNexp is the strongest overall method based on the average.}
\end{table}

\begin{table}[H]
\centering
\begin{adjustbox}{width=0.5\textwidth}
\begin{tabular}{l c c c c}
\textbf{Method}& & \textbf{Hate Speech Offensive} &  & \\
\textit{Balanced} & $1^{st}$ & $5^{th}$ & $10^{th}$ & Average \\
\midrule
RoBERTa$_{full}$\ & $59.7_{3.5}$ & $66.9_{1.2}$ & $\textbf{69.2}_{1.8}$ & $\textbf{66.4}_{2.7}$  \\
SetFit$_{exp}$\  & $60.7_{1.3}$ & $66.3_{1.6}$ & $67.5_{0.9}$  & $65.9_{2.2}$  \\
\LaGoNNexp \ & $\textbf{61.5}_{1.7}$  & $66.4_{1.4}$  & $67.7_{0.9}$  & $66.1_{1.8}$  \\
\midrule
SetFit$_{lite}$\ & $60.7_{1.3}$  & $66.3_{2.0}$ & $66.5_{0.9}$  & $65.1_{1.7}$   \\
\LaGoNNlite\  & $\textbf{61.5}_{1.7}$  & $\textbf{67.1}_{1.1}$ & $67.3_{0.8}$ & $66.0_{1.7}$ \\

\midrule
RoBERTa$_{freeze}$\ & $59.7_{3.5}$  & $60.4_{2.7}$  & $63.1_{2.3}$ & $61.0_{1.3}$  \\
$k$NN\     & $60.7_{1.3}$  & $59.6_{2.8}$ & $59.5_{2.5}$  & $59.5_{0.5}$   \\

SetFit \ & $60.7_{1.3}$  & $62.5_{0.7}$ & $63.4_{1.0}$  & $62.3_{1.0}$   \\
\LaGoNN \ & $\textbf{61.5}_{1.7}$  & $62.8_{1.5}$ & $64.2_{1.0}$  & $63.0_{0.9}$  \\
\midrule
Probe\   & $54.9_{1.4}$  & $58.5_{0.9}$ & $60.9_{0.4}$ & $58.7_{1.7}$  \\
\LaGoNNcheap\  & $54.2_{2.3}$  & $58.6_{0.6}$ & $60.6_{0.5}$ & $58.5_{1.8}$ \\
\bottomrule
\end{tabular}
\end{adjustbox}
\caption{Similar to the moderate setting, on the first step, \LaGoNN, \LaGoNNlite, and \LaGoNNexp start out the strongest, but RoBERTa$_{full}$ overtakes \LaGoNNlite by the $10^{th}$ step. RoBERTa$_{full}$ slightly outperforms \LaGoNNlite and \LaGoNNexp as the overall strongest method based on the average.}
\end{table}
\clearpage

\begin{table}[H]
\centering
\begin{adjustbox}{width=0.5\textwidth}
\begin{tabular}{l c c c c}
\textbf{Method}& & \textbf{LIAR} &  & \\
\textit{Extreme} & $1^{st}$ & $5^{th}$ & $10^{th}$ & Average \\
\midrule
RoBERTa$_{full}$\ & $\textbf{32.0}_{2.7}$ & $\textbf{34.7}_{2.9}$ & $35.1_{4.3}$ & $33.7_{1.0}$  \\
SetFit$_{exp}$\  & $31.2_{3.8}$ & $30.4_{3.1}$ & $31.8_{2.9}$  & $31.5_{0.7}$  \\
\LaGoNNexp \ & $30.6_{4.7}$  & $30.3_{2.0}$  & $31.3_{2.0}$  & $31.1_{0.6}$  \\
\midrule
SetFit$_{lite}$\ & $31.2_{3.8}$  & $32.7_{3.8}$ & $33.5_{4.2}$  & $32.7_{0.8}$   \\
\LaGoNNlite\  & $30.6_{4.7}$  & $31.8_{3.9}$ & $32.4_{2.7}$ & $31.6_{0.6}$ \\

\midrule
RoBERTa$_{freeze}$\ & $\textbf{32.0}_{2.7}$  & $32.8_{4.5}$  & $34.2_{5.0}$ & $33.2_{0.7}$  \\
$k$NN\     & $27.0_{0.5}$  & $27.3_{0.8}$ & $27.9_{0.8}$  & $27.4_{0.3}$   \\

SetFit \ & $31.2_{3.8}$  & $33.7_{5.1}$ & $\textbf{35.7}_{5.1}$  & $\textbf{34.3}_{1.6}$   \\
\LaGoNN \ & $30.6_{4.7}$  & $32.0_{4.6}$ & $33.7_{5.4}$  & $32.6_{0.9}$  \\
\midrule
Probe\   & $30.7_{2.0}$  & $30.6_{3.9}$ & $31.7_{2.9}$ & $31.1_{0.4}$  \\
\LaGoNNcheap\  & $30.7_{2.0}$  & $30.5_{3.8}$ & $31.4_{2.6}$ & $31.0_{0.4}$ \\
\bottomrule
\end{tabular}
\end{adjustbox}
\caption{RoBERTa$_{freeze}$ and RoBERTa$_{full}$ start out performant and RoBERTa$_{full}$ continues to be until the $10^{th}$ step where it is overtaken by SetFit, which ends up being the strongest overall method.}
\end{table}

\begin{table}[H]
\centering
\begin{adjustbox}{width=0.5\textwidth}
\begin{tabular}{l c c c c}
\textbf{Method}& & \textbf{LIAR} &  & \\
\textit{Imbalanced} & $1^{st}$ & $5^{th}$ & $10^{th}$ & Average \\
\midrule
RoBERTa$_{full}$\ & $31.4_{3.2}$ & $35.8_{2.6}$ & $\textbf{40.0}_{4.3}$ & $36.2_{2.4}$  \\
SetFit$_{exp}$\  & $\textbf{32.3}_{4.5}$ & $35.9_{3.1}$ & $36.4_{2.2}$  & $35.2_{1.1}$  \\
\LaGoNNexp \ & $\textbf{32.3}_{4.6}$  & $35.7_{3.4}$  & $36.5_{2.3}$  & $35.7_{1.4}$  \\
\midrule
SetFit$_{lite}$\ & $\textbf{32.3}_{4.5}$  & $35.6_{2.7}$ & $37.4_{2.6}$  & $35.8_{1.6}$   \\
\LaGoNNlite\  & $\textbf{32.3}_{4.6}$  & $35.2_{2.4}$ & $36.6_{2.7}$ & $35.5_{1.3}$ \\

\midrule
RoBERTa$_{freeze}$\ & $31.4_{3.2}$  & $34.1_{2.6}$  & $35.6_{3.2}$ & $34.0_{1.4}$  \\
$k$NN\     & $27.0_{0.2}$  & $28.5_{1.0}$ & $29.0_{1.0}$  & $28.7_{0.7}$   \\

SetFit\ & $\textbf{32.3}_{4.5}$  & $\textbf{36.5}_{3.1}$ & $38.5_{3.4}$  & $\textbf{36.3}_{2.0}$   \\
\LaGoNN \ & $\textbf{32.3}_{4.6}$  & $34.9_{2.2}$ & $36.9_{2.5}$  & $35.3_{1.4}$  \\
\midrule
Probe\   & $30.7_{3.0}$  & $32.8_{1.8}$ & $35.0_{1.6}$ & $33.5_{1.5}$  \\
\LaGoNNcheap\  & $30.4_{3.0}$  & $32.9_{1.8}$ & $35.4_{1.7}$ & $33.5_{1.7}$ \\
\bottomrule
\end{tabular}
\end{adjustbox}
\caption{\LaGoNN, \LaGoNNlite, \LaGoNNexp, SetFit, SetFit$_{lite}$, and SetFit$_{exp}$ start out as the most performant, but SetFit is the strongest on the $5^{th}$ step and RoBERTa$_{full}$ on the $10^{th}$. Overall, SetFit is strongest method based on the average over all steps.}
\end{table}

\begin{table}[H]
\centering
\begin{adjustbox}{width=0.5\textwidth}
\begin{tabular}{l c c c c}
\textbf{Method}& & \textbf{LIAR} &  & \\
\textit{Moderate} & $1^{st}$ & $5^{th}$ & $10^{th}$ & Average \\
\midrule
RoBERTa$_{full}$\ & $33.9_{3.1}$ & $38.4_{2.7}$ & $\textbf{43.9}_{2.2}$ & $\textbf{39.5}_{3.0}$  \\
SetFit$_{exp}$\  & $33.0_{2.6}$ & $37.2_{1.8}$ & $38.7_{1.5}$  & $37.4_{1.6}$  \\
\LaGoNNexp \ & $\textbf{34.1}_{3.4}$  & $\textbf{38.7}_{2.3}$  & $39.0_{1.8}$  & $37.8_{1.5}$  \\
\midrule
SetFit$_{lite}$\ & $33.0_{2.6}$  & $38.5_{1.3}$ & $40.4_{2.0}$  & $38.2_{2.1}$   \\
\LaGoNNlite\  & $\textbf{34.1}_{3.4}$  & $38.4_{2.0}$ & $39.6_{1.5}$ & $37.9_{1.6}$ \\

\midrule
RoBERTa$_{freeze}$\ & $33.9_{3.1}$  & $35.3_{2.6}$  & $36.8_{2.2}$ & $35.4_{1.0}$  \\
$k$NN\     & $29.2_{0.8}$  & $29.7_{1.5}$ & $30.0_{0.6}$  & $29.8_{0.3}$   \\

SetFit \ & $33.0_{2.6}$  & $37.2_{3.9}$ & $39.4_{3.5}$  & $37.0_{1.8}$   \\
\LaGoNN \ & $\textbf{34.1}_{3.4}$  & $37.0_{3.1}$ & $38.6_{3.0}$  & $36.8_{1.3}$  \\
\midrule
Probe\   & $31.6_{1.1}$  & $34.7_{2.5}$ & $37.0_{2.5}$ & $34.9_{1.7}$  \\
\LaGoNNcheap\  & $31.4_{0.9}$  & $35.3_{2.3}$ & $37.6_{2.0}$ & $35.3_{1.9}$ \\
\bottomrule
\end{tabular}
\end{adjustbox}
\caption{\LaGoNN, \LaGoNNlite,  and \LaGoNNexp are the most performant classifiers on the first step, while \LaGoNNexp remains strong until the $10^{th}$ step where it is overtaken by RoBERTa$_{full}$. RoBERTa$_{full}$ is the overally strongest method if we aggregate over all steps.}
\end{table}

\begin{table}[H]
\centering
\begin{adjustbox}{width=0.5\textwidth}
\begin{tabular}{l c c c c}
\textbf{Method}& & \textbf{LIAR} &  & \\
\textit{Balanced} & $1^{st}$ & $5^{th}$ & $10^{th}$ & Average \\
\midrule
RoBERTa$_{full}$\ & $33.8_{2.1}$ & $\textbf{39.4}_{2.4}$ & $\textbf{43.5}_{1.7}$ & $\textbf{40.2}_{3.2}$  \\
SetFit$_{exp}$\  & $\textbf{34.4}_{2.3}$ & $36.7_{1.7}$ & $37.0_{1.3}$  & $36.5_{1.1}$  \\
\LaGoNNexp \ & $33.8_{1.8}$  & $34.2_{2.7}$  & $37.2_{1.9}$  & $36.2_{1.4}$  \\
\midrule
SetFit$_{lite}$\ & $\textbf{34.4}_{2.3}$  & $38.7_{2.3}$ & $40.3_{2.8}$  & $38.0_{2.1}$   \\
\LaGoNNlite\  & $33.8_{1.8}$  & $37.6_{2.0}$ & $39.4_{2.8}$ & $37.2_{1.9}$ \\

\midrule
RoBERTa$_{freeze}$\ & $33.8_{2.1}$  & $36.6_{1.6}$  & $38.6_{1.5}$ & $36.7_{1.5}$  \\
$k$NN\     & $30.1_{0.4}$  & $31.3_{2.1}$ & $30.6_{1.1}$  & $30.9_{0.4}$   \\

SetFit\ & $\textbf{34.4}_{2.3}$  & $38.3_{2.5}$ & $40.0_{2.0}$  & $37.9_{1.6}$   \\
\LaGoNN \ & $33.8_{1.8}$  & $38.3_{1.3}$ & $40.6_{0.6}$  & $38.1_{2.0}$  \\
\midrule
Probe\   & $32.1_{1.9}$  & $35.2_{1.4}$ & $37.2_{2.5}$ & $35.2_{1.7}$  \\
\LaGoNNcheap\  & $31.9_{1.9}$  & $36.0_{1.0}$ & $37.5_{2.5}$ & $35.7_{1.8}$ \\
\bottomrule
\end{tabular}
\end{adjustbox}
\caption{SetFit, SetFit$_{lite}$, and SetFit$_{exp}$ start out the strongest on the first step, but are overtaken by RoBERTa$_{full}$ on the $5^{th}$ which remains the most performant on the $10^{th}$ step and if we consider the average over all steps.}
\end{table}

\newpage
\subsection{Additional results: general text classification}
\label{sec:generalappendix}
\begin{figure*}[h!]
    \centering
    \includegraphics[scale=0.44]{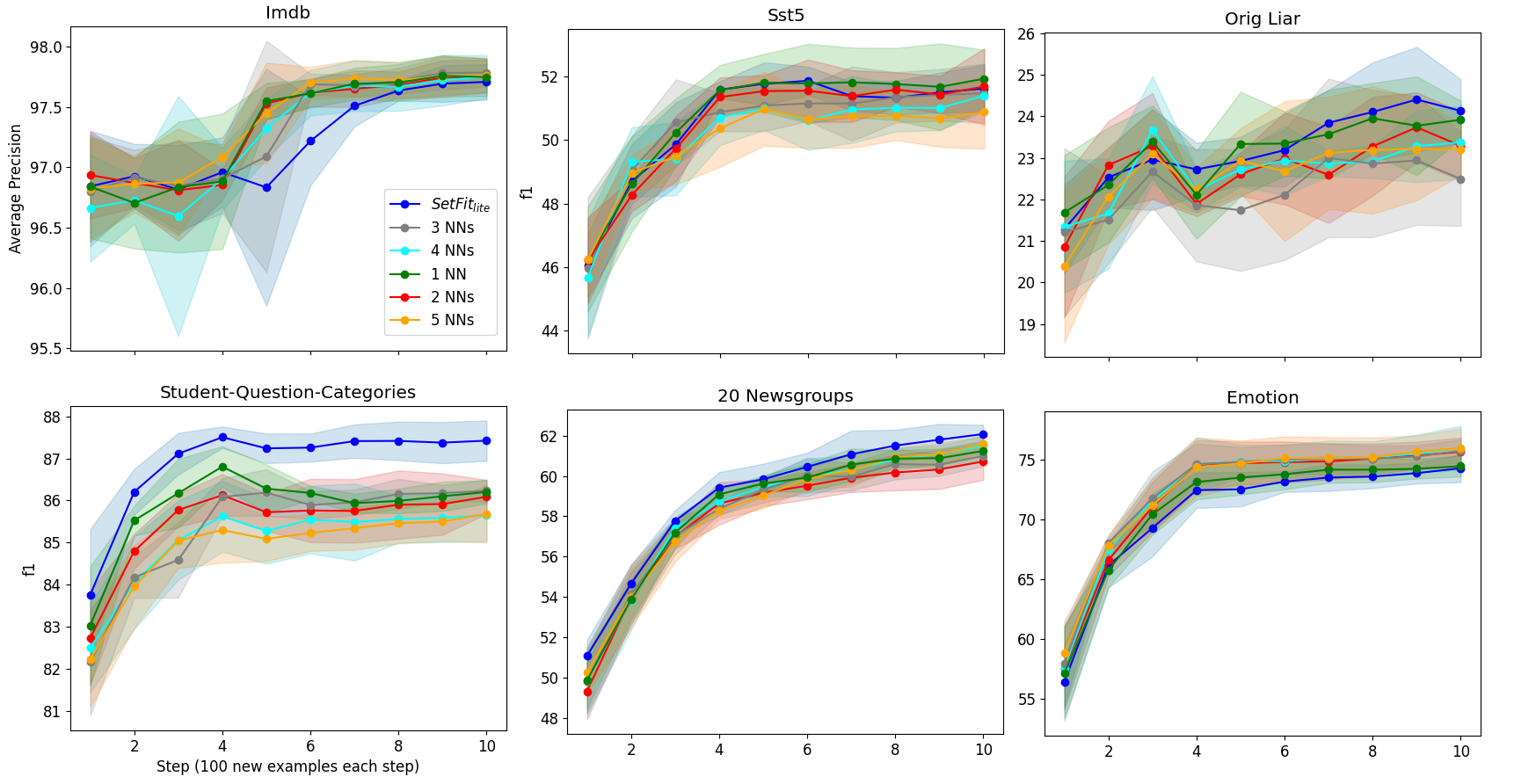}
    \captionof{figure}{SetFit$_{lite}$ performance compared against one to five neighbors for \LaGoNNlite \labdist. The measure is average precision for IMDB, macro-F1 elsewhere.}
    \label{generallite_appendix}
\end{figure*}

\begin{figure*}[h!]
    \centering
    \includegraphics[scale=0.44]{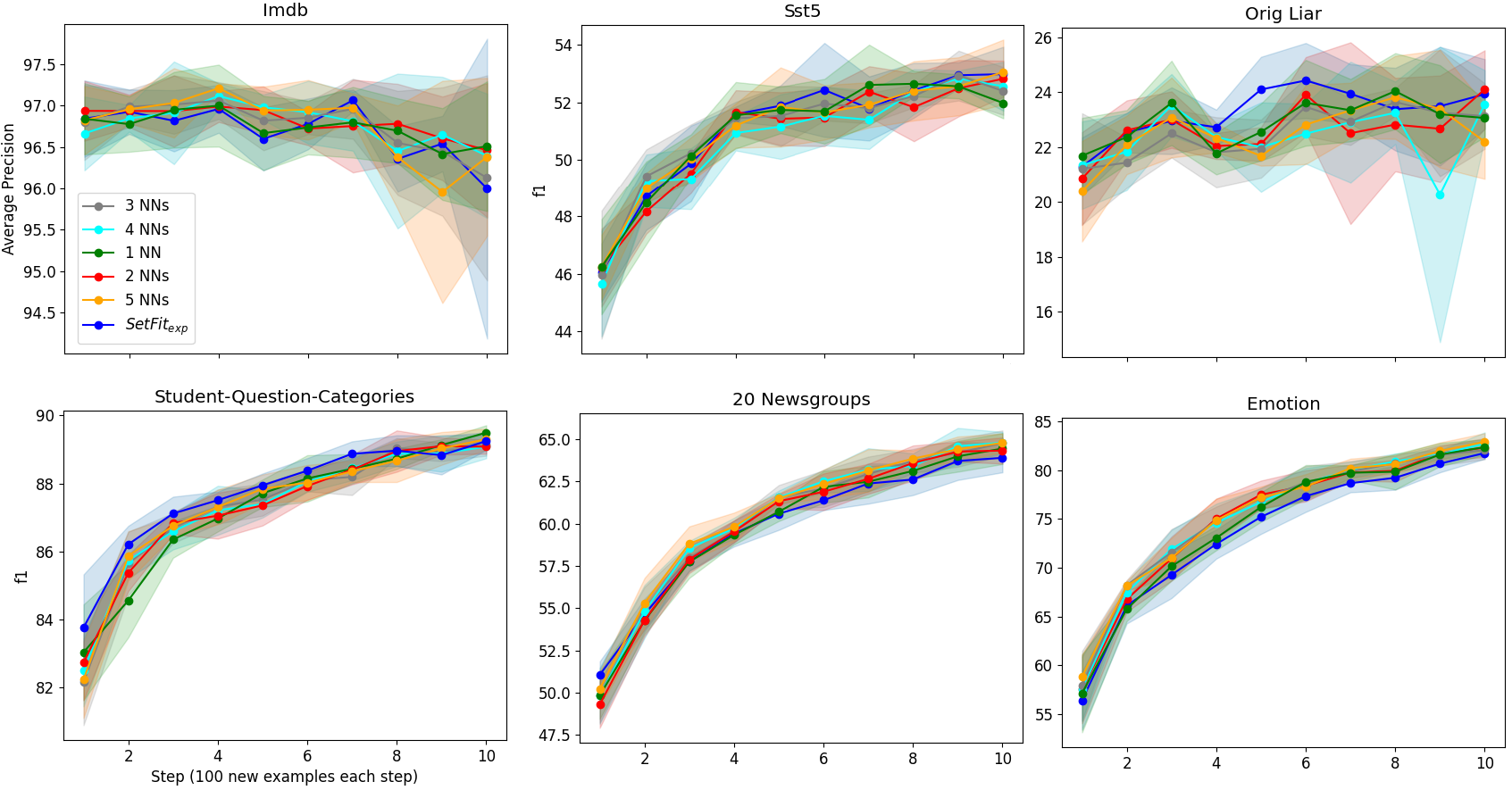}
    \captionof{figure}{SetFit$_{exp}$ performance compared against one to five neighbors for \LaGoNNexp \labdist. The measure is average precision for IMDB, macro-F1 elsewhere.}
    \label{generalexp_appendix}
\end{figure*}

In this Appendix section, we provide additional results from our general text classification experiments in the main text, Section \ref{general_main}. Here we show results comparing \LaGoNNlite against SetFit$_{lite}$ and \LaGoNNexp against SetFit$_{exp}$,  but we include results for one to five neighbors with \LaGoNN \labdist, Figures \ref{generallite_appendix} and \ref{generalexp_appendix}, respectively. The measure is average precision for IMDB, macro-F1 elsewhere.

In general, the number of neighbors we consider does not appear to have a large impact on \LaGoNN’s predictive power and our method continues to be a more stable classifier than SetFit and can generally be expected to improve SetFit’s performance. We also see that continued fine-tuning with the embedding model is only helpful for cases when the dataset has a relatively large number of labels.  One exception to this is the case of Student Question Categories, where there are four labels. While it is clear that SetFit$_{lite}$ is a stronger model than \LaGoNN lite, if we consider the more expensive alternatives, the story changes; if we continue to fine-tune, the prediction curves are essentially the same, and \LaGoNNexp seems to have a slight edge on SetFit$_{exp}$ as we add training data.

LIAR, both the collapsed version we considered in our content moderation experiments and the original version (Orig Liar) we examine in our general text classification experiments here, seems to be a very difficult dataset. Adding examples or increased fine-tuning does not appear to consistently increase model performance. We observed this across all experimental settings and balanced regimes and is a sensible finding, as it should be very difficult to determine the truth of a specific statement without additional context.

\clearpage
\newpage

\subsection{Additional results: multilingual text classification}
\label{sec:multilingualappendix}

In this Appendix section, we provide multilingual text classification results from experiments where we compare SetFit$_{exp}$ and SetFit$_{lite}$ against \LaGoNNexp and \LaGoNNlite respectively. For these experiments, we used the Multilingual Amazon Reviews Corpus \cite{marc_reviews}, which has five labels, where each label is a star rating in Chinese, English, French, German, Japanese, or Spanish.\footnote{\url{https://huggingface.co/datasets/amazon_reviews_multi}} To create the mapping from label to text, we used code from the ADAPET \cite{tam-etal-2021-improving} port in the official SetFit repository.\footnote{\url{https://github.com/huggingface/setfit/blob/main/scripts/adapet/ADAPET/utilcode.py}} In these experiments, we used the same multilingual pretrained Sentence Transformer for all models under the balanced sampling regime.\footnote{\url{https://huggingface.co/sentence-transformers/paraphrase-multilingual-mpnet-base-v2}} In the case of \LaGoNNexp and \LaGoNNlite, we use \labdist and search over one to five neighbors, reporting all results.

\begin{figure*}[t]
    \centering
    \includegraphics[scale=0.43]{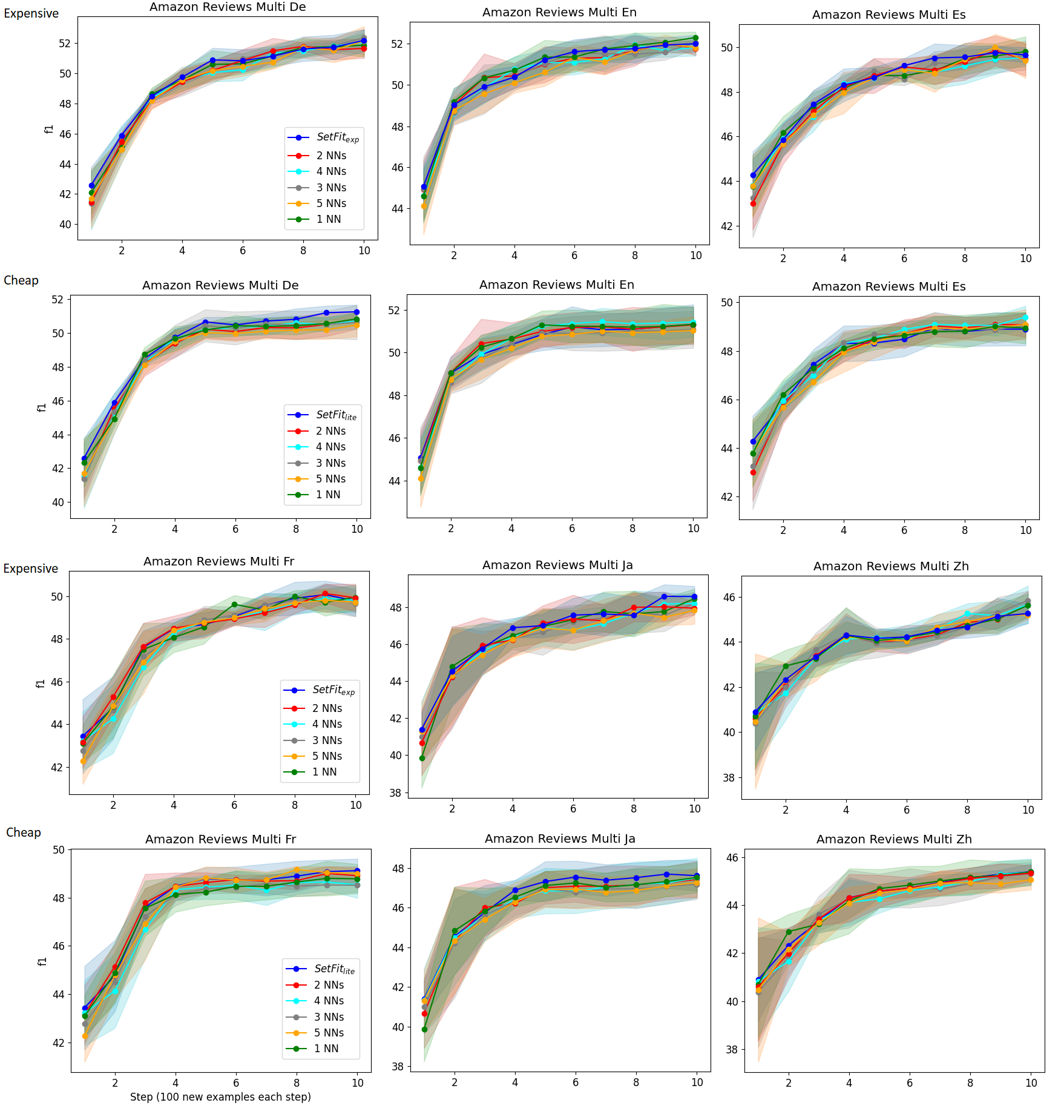}
    \captionof{figure}{Multilingual classification experiments. In the first row, we display results from expensive models on German, English, Spanish data, with their cheaper counterparts in the following row. In the third and fourth row, we do the same but for French, Japanese, and Chinese. The measure is macro-F1 in all cases.}
    \label{multilingual}
\end{figure*}

Figure \ref{multilingual} shows our results for expensive and inexpensive models. We note in all cases all models perform similarly. This supports our assertion in Section \ref{general_main} that when the training data is balanced and we have only a handful of labels or less, it is sufficient to fine-tune the Sentence Transformer on only a subset of available training data. A classifier can then be fit on all available data, encoded with the fine-tuned ST. We observed this for SST-5 and observe it again here, especially clearly on the Chinese subset of this dataset. SetFit$_{exp}$ plateaus on the fifth step and stops learning, with different versions of \LaGoNNexp outperforming it on later steps. However, if we move down on row, we see that all cheaper models continue to learn on all steps.

\clearpage
\newpage
\subsection{Ablations}
In this Appendix section, we perform ablation studies with \LaGoNN to support our findings in the main text.

\subsubsection{Ablation: \LaGoNN configurations}
\label{sec:appendix_configs}
Here, we provide an in-depth comparison between all \LaGoNN configurations, \labell, \dist, \labdist, \textt, and \all (see Table \ref{table:LaGoNNtypes}) for all content moderation datasets, balances, and levels of expense. The evaluation measure is the mean average precision ($\times 100$) over five seeds in all cases except for Hate Speech Offensive where the measure is the macro-F1. 

\begin{figure*}[h!]
    \centering
    \includegraphics[scale=0.24]{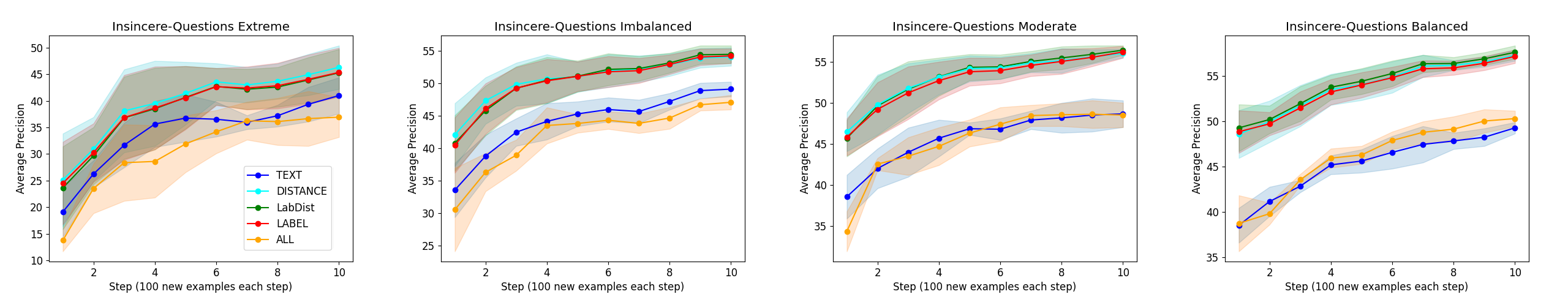}
    \captionof{figure}{\LaGoNNcheap performance for all configurations and balance regimes on the Insincere Questions dataset. The relevant balance is in the title of each panel.}
    \label{lagonn_cheap_iq}
\end{figure*}

\begin{figure*}[h]
    \centering
    \includegraphics[scale=0.24]{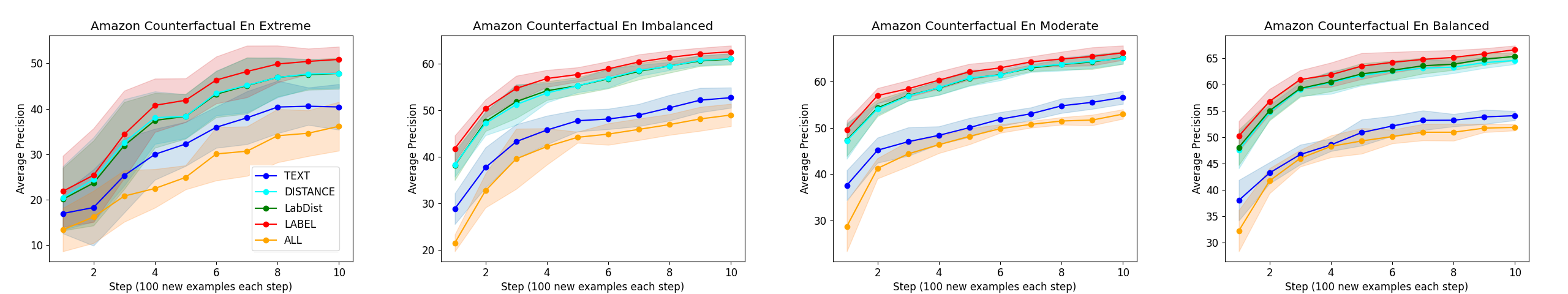}
    \captionof{figure}{\LaGoNNcheap performance for all configurations and balance regimes on the Amazon Counterfactual dataset. The relevant balance is in the title of each panel.}
    \label{lagonn_cheap_ac}
\end{figure*}

\begin{figure*}[h]
    \centering
    \includegraphics[scale=0.24]{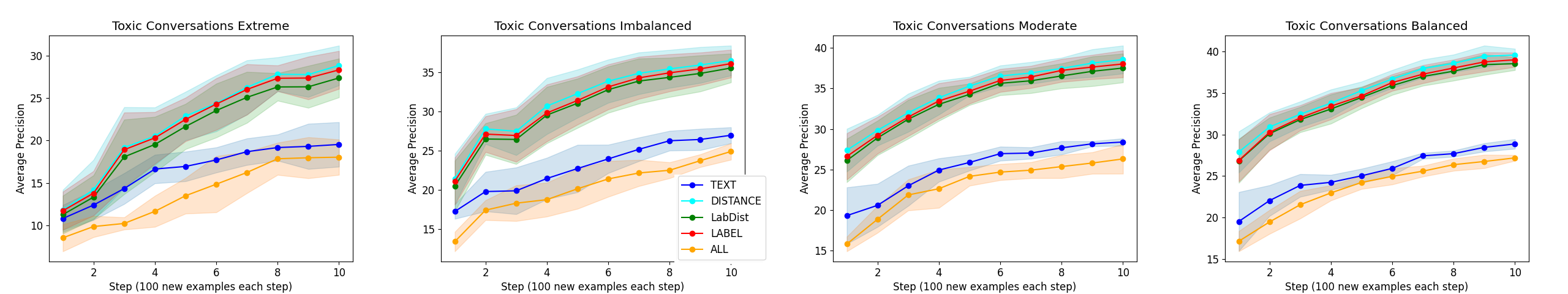}
    \captionof{figure}{\LaGoNNcheap performance for all configurations and balance regimes on the Toxic Conversations dataset. The relevant balance is in the title of each panel.}
    \label{lagonn_cheap_tx}
\end{figure*}

\begin{figure*}[h]
    \centering
    \includegraphics[scale=0.24]{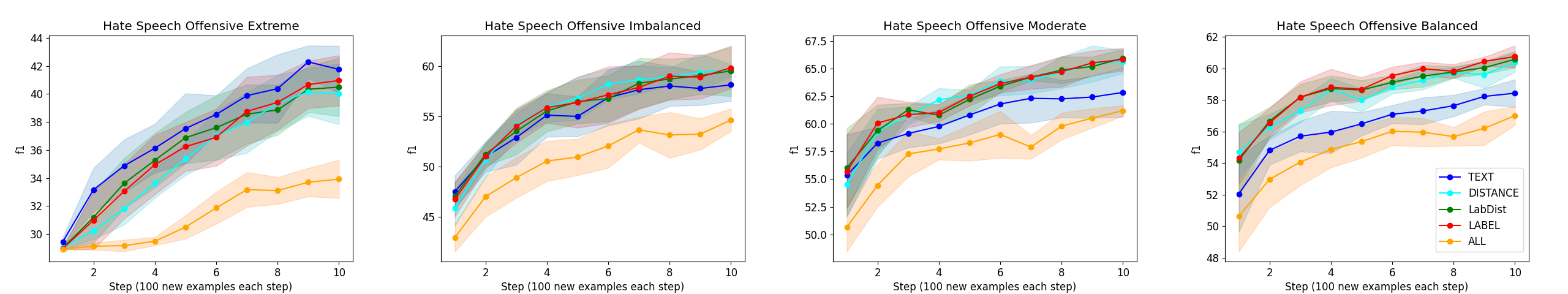}
    \captionof{figure}{\LaGoNNcheap performance for all configurations and balance regimes on the Hate Speech Offensive dataset. The relevant balance is in the title of each panel.}
    \label{lagonn_cheap_hs}
\end{figure*}

\begin{figure*}[h]
    \centering
    \includegraphics[scale=0.24]{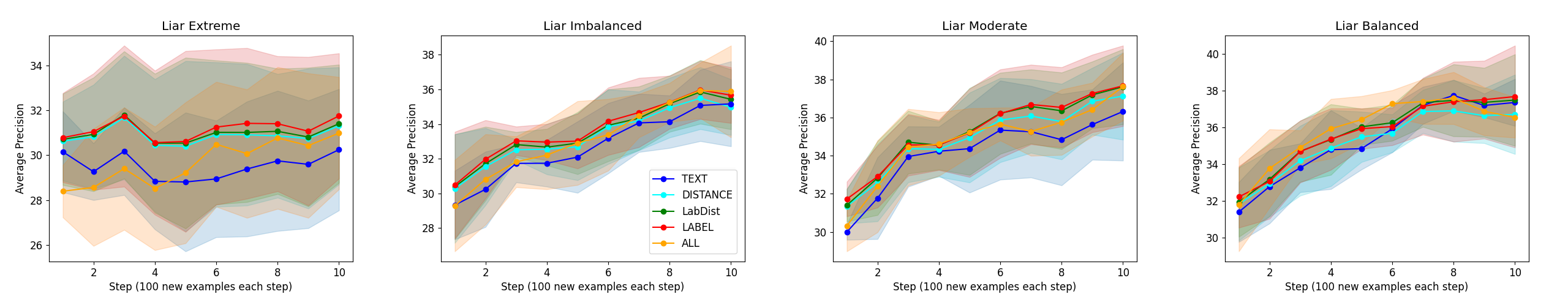}
    \captionof{figure}{\LaGoNNcheap performance for all configurations and balance regimes on the LIAR dataset. The relevant balance is in the title of each panel.}
    \label{lagonn_cheap_LIAR}
\end{figure*}

\begin{figure*}[h!]
    \centering
    \includegraphics[scale=0.24]{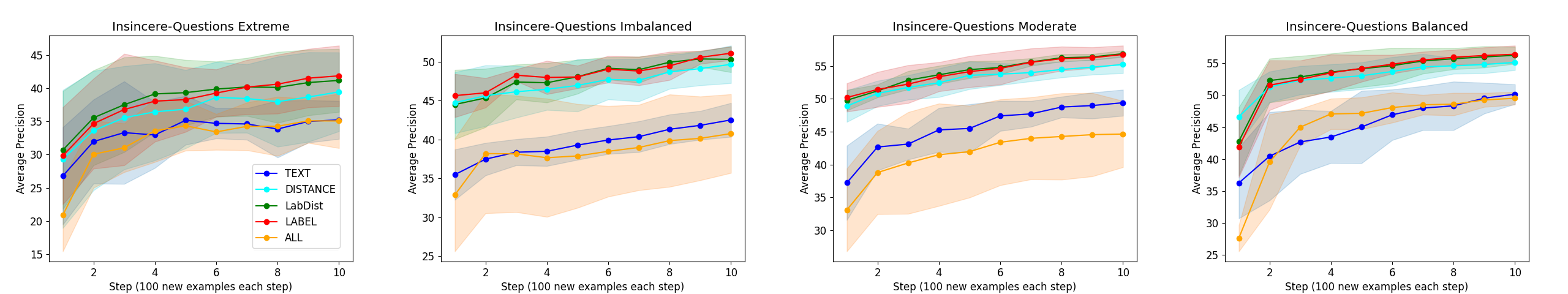}
    \captionof{figure}{\LaGoNN performance for all configurations and balance regimes on the Insincere Questions dataset. The relevant balance is in the title of each panel.}
    \label{lagonn_iq}
\end{figure*}

\begin{figure*}[h]
    \centering
    \includegraphics[scale=0.24]{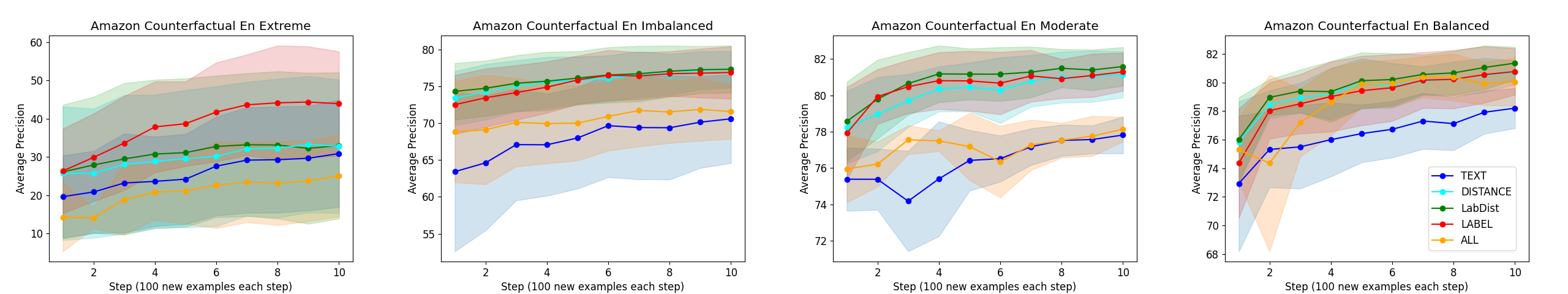}
    \captionof{figure}{\LaGoNN performance for all configurations and balance regimes on the Amazon Counterfactual dataset. The relevant balance is in the title of each panel.}
    \label{lagonn_ac}
\end{figure*}

\begin{figure*}[h]
    \centering
    \includegraphics[scale=0.24]{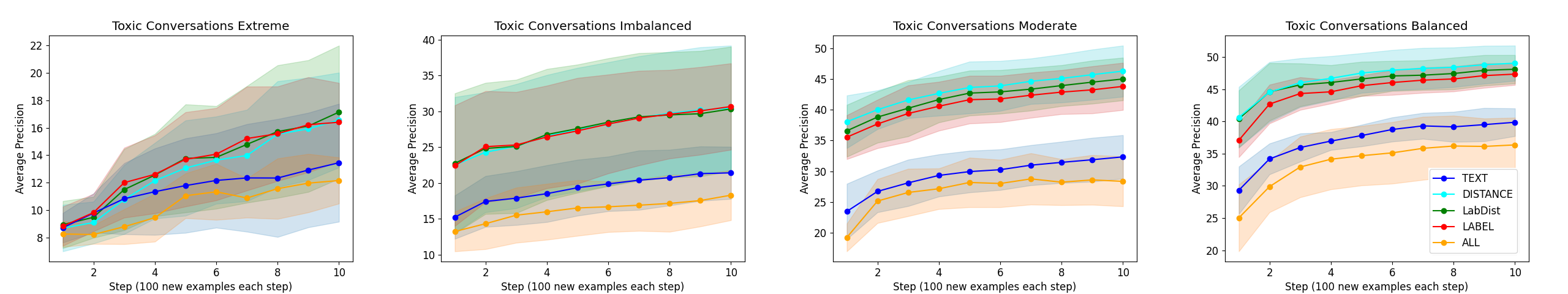}
    \captionof{figure}{\LaGoNN performance for all configurations and balance regimes on the Toxic Conversations dataset. The relevant balance is in the title of each panel.}
    \label{lagonn_tx}
\end{figure*}

\begin{figure*}[h]
    \centering
    \includegraphics[scale=0.24]{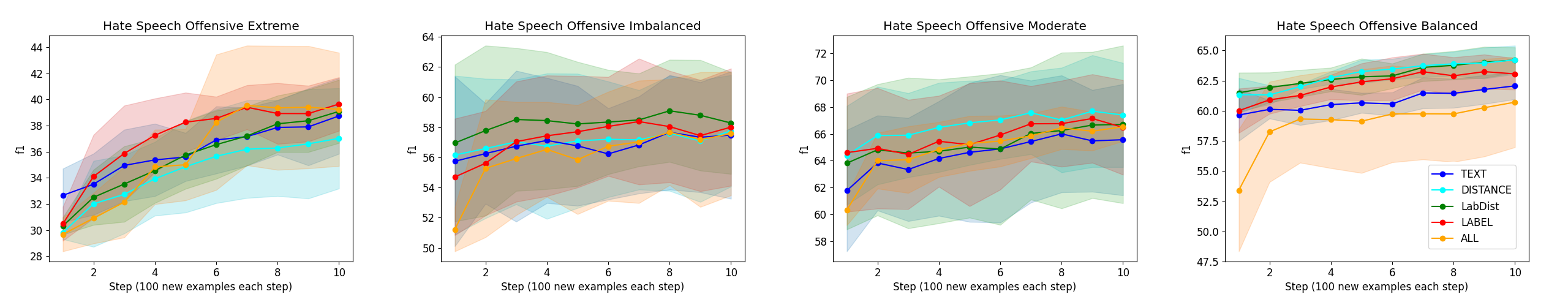}
    \captionof{figure}{\LaGoNN performance for all configurations and balance regimes on the Hate Speech Offensive dataset. The relevant balance is in the title of each panel.}
    \label{lagonn_hs}
\end{figure*}

\begin{figure*}[h]
    \centering
    \includegraphics[scale=0.24]{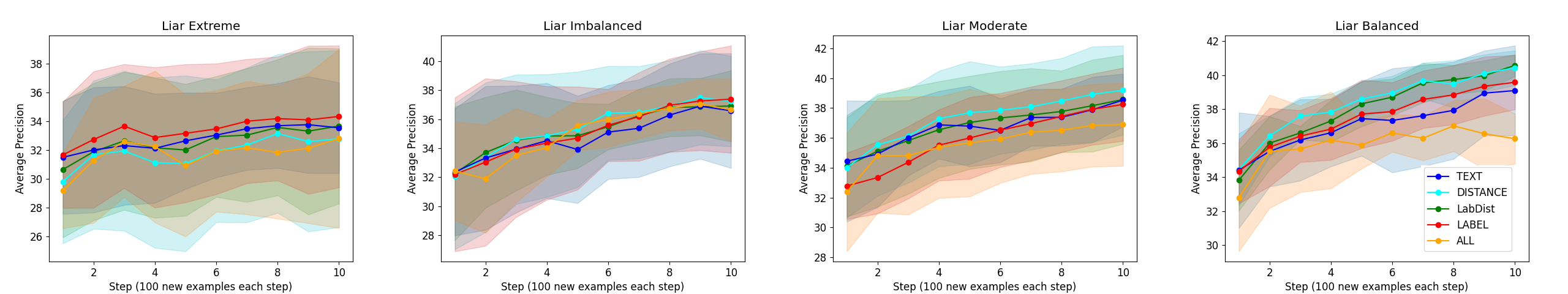}
    \captionof{figure}{\LaGoNN performance for all configurations and balance regimes on the LIAR dataset. The relevant balance is in the title of each panel.}
    \label{lagonn_LIAR}
\end{figure*}

\begin{figure*}[h!]
    \centering
    \includegraphics[scale=0.24]{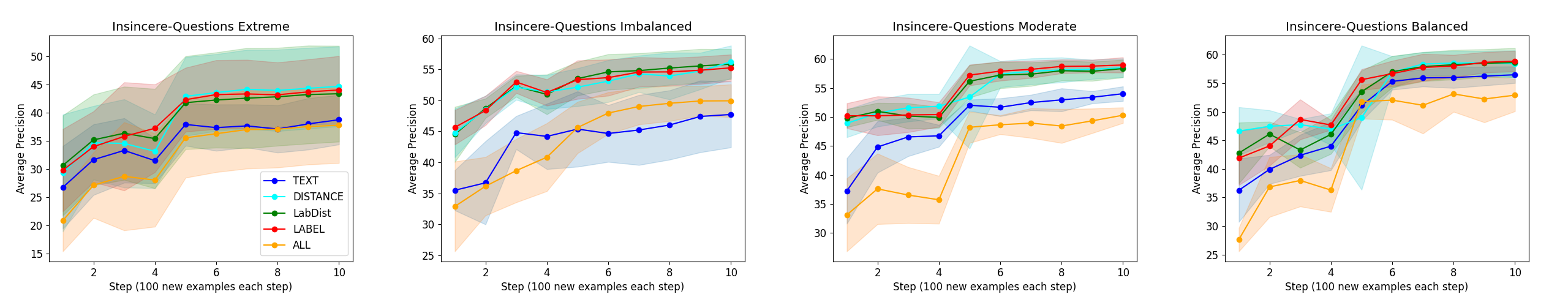}
    \captionof{figure}{\LaGoNNlite performance for all configurations and balance regimes on the Insincere Questions dataset. The relevant balance is in the title of each panel.}
    \label{lagonn_lite_iq}
\end{figure*}

\begin{figure*}[h]
    \centering
    \includegraphics[scale=0.24]{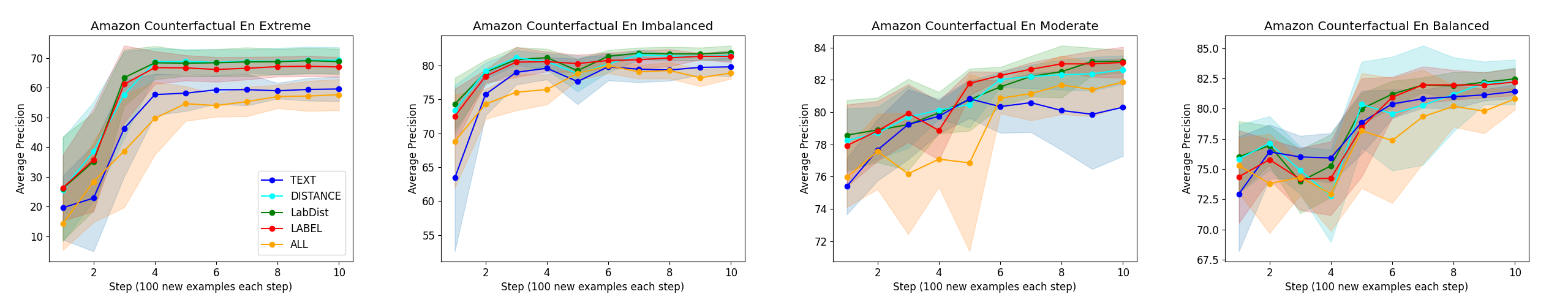}
    \captionof{figure}{\LaGoNNlite performance for all configurations and balance regimes on the Amazon Counterfactual dataset. The relevant balance is in the title of each panel.}
    \label{lagonn_lite_ac}
\end{figure*}

\begin{figure*}[h]
    \centering
    \includegraphics[scale=0.24]{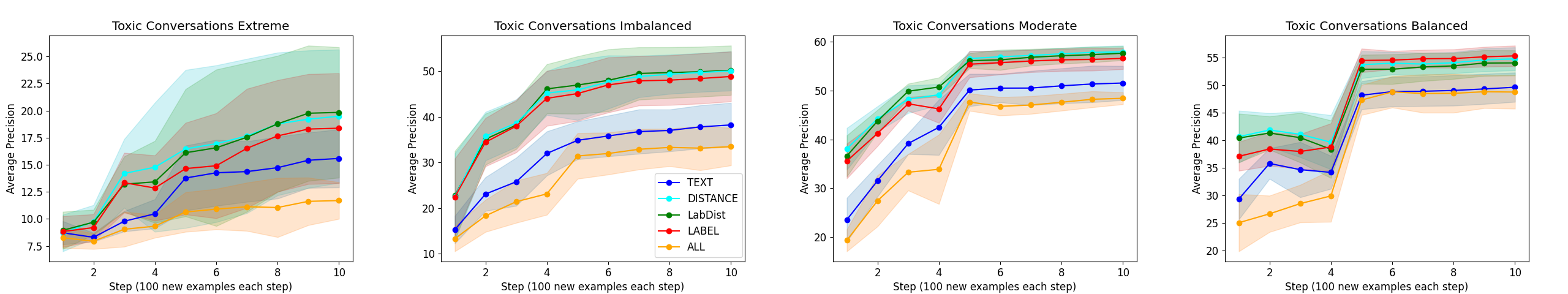}
    \captionof{figure}{\LaGoNNlite performance for all configurations and balance regimes on the Toxic Conversations dataset. The relevant balance is in the title of each panel.}
    \label{lagonn_lite_tx}
\end{figure*}

\begin{figure*}[h]
    \centering
    \includegraphics[scale=0.24]{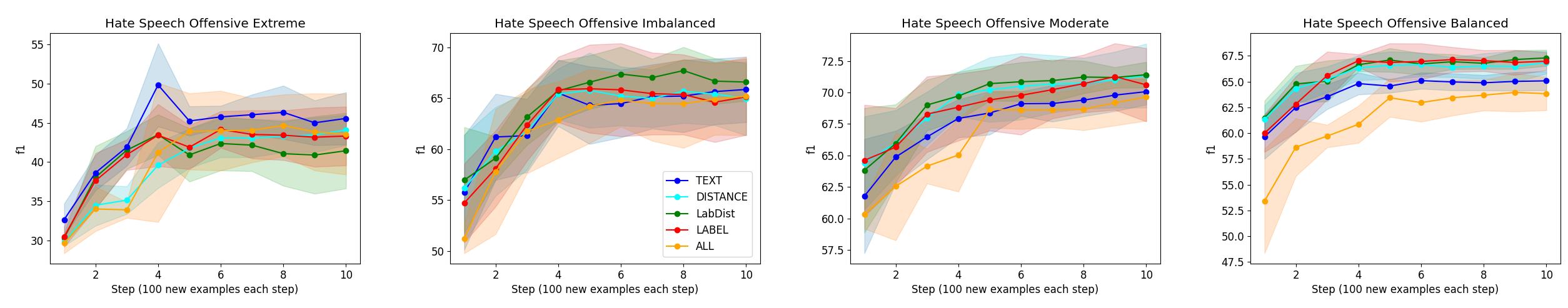}
    \captionof{figure}{\LaGoNNlite performance for all configurations and balance regimes on the Hate Speech Offensive dataset. The relevant balance is in the title of each panel.}
    \label{lagonn_lite_hs}
\end{figure*}

\begin{figure*}[h]
    \centering
    \includegraphics[scale=0.24]{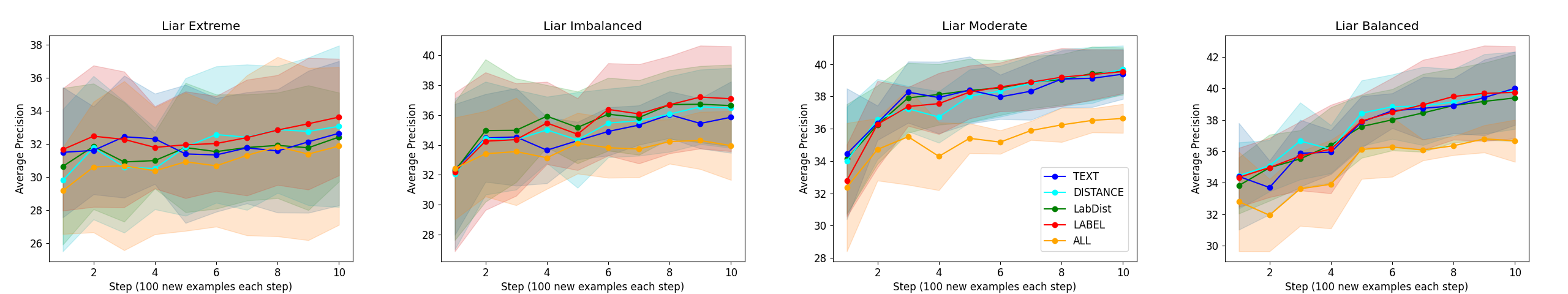}
    \captionof{figure}{\LaGoNNlite performance for all configurations and balance regimes on the LIAR dataset. The relevant balance is in the title of each panel.}
    \label{lagonn_lite_LIAR}
\end{figure*}

\begin{figure*}[h!]
    \centering
    \includegraphics[scale=0.24]{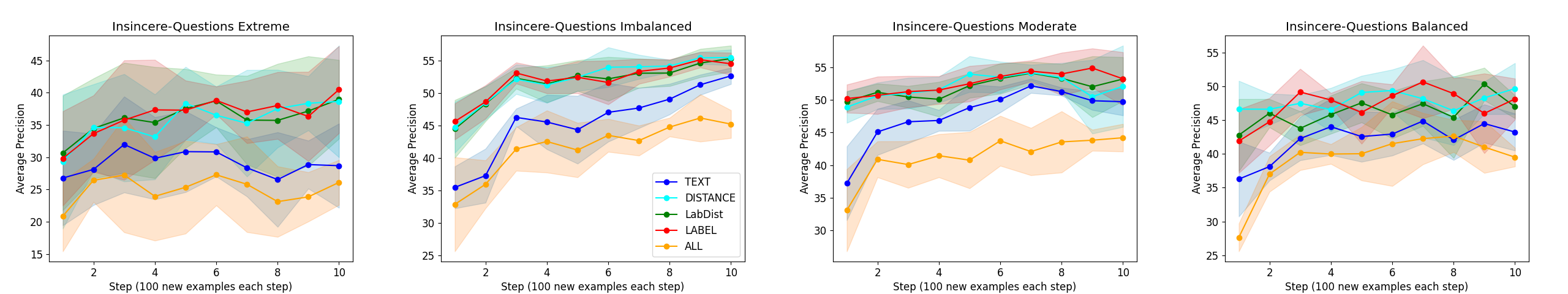}
    \captionof{figure}{\LaGoNNexp performance for all configurations and balance regimes on the Insincere Questions dataset. The relevant balance is in the title of each panel.}
    \label{lagonn_exp_iq}
\end{figure*}

\begin{figure*}[h]
    \centering
    \includegraphics[scale=0.24]{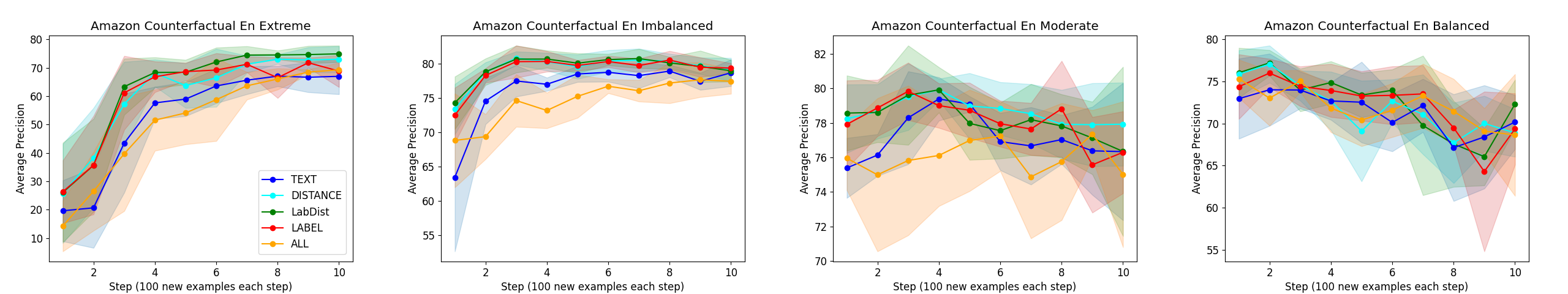}
    \captionof{figure}{\LaGoNNexp performance for all configurations and balance regimes on the Amazon Counterfactual dataset. The relevant balance is in the title of each panel.}
    \label{lagonn_exp_ac}
\end{figure*}

\begin{figure*}[h]
    \centering
    \includegraphics[scale=0.24]{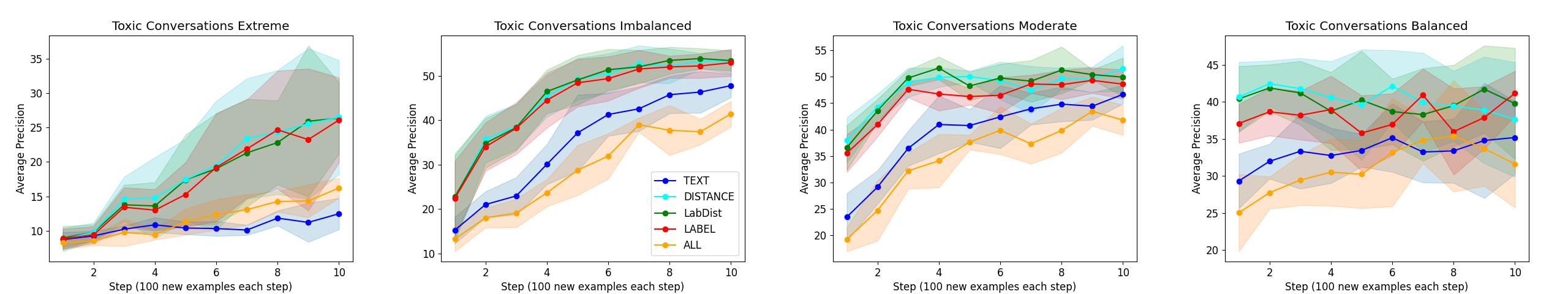}
    \captionof{figure}{\LaGoNNexp performance for all configurations and balance regimes on the Toxic Conversations dataset. The relevant balance is in the title of each panel.}
    \label{lagonn_exp_tx}
\end{figure*}

\begin{figure*}[h]
    \centering
    \includegraphics[scale=0.24]{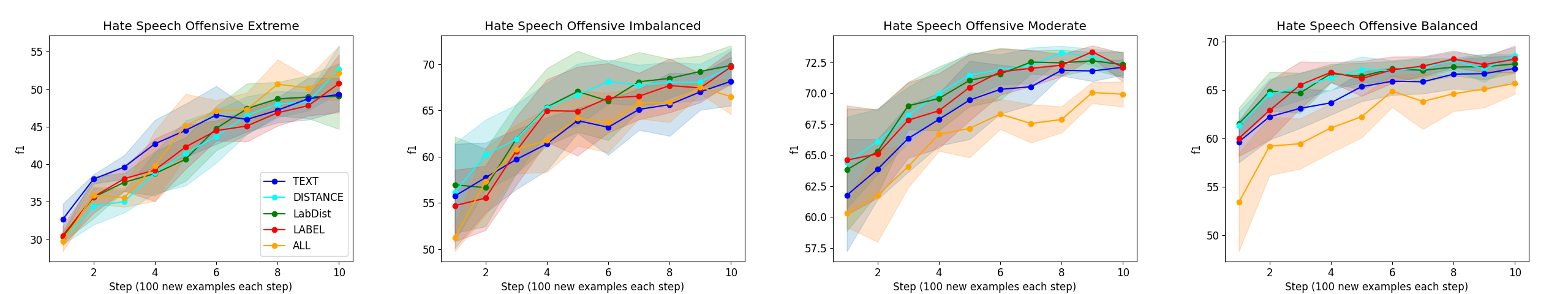}
    \captionof{figure}{\LaGoNNexp performance for all configurations and balance regimes on the Hate Speech Offensive dataset. The relevant balance is in the title of each panel.}
    \label{lagonn_exp_hs}
\end{figure*}

\begin{figure*}[h]
    \centering
    \includegraphics[scale=0.24]{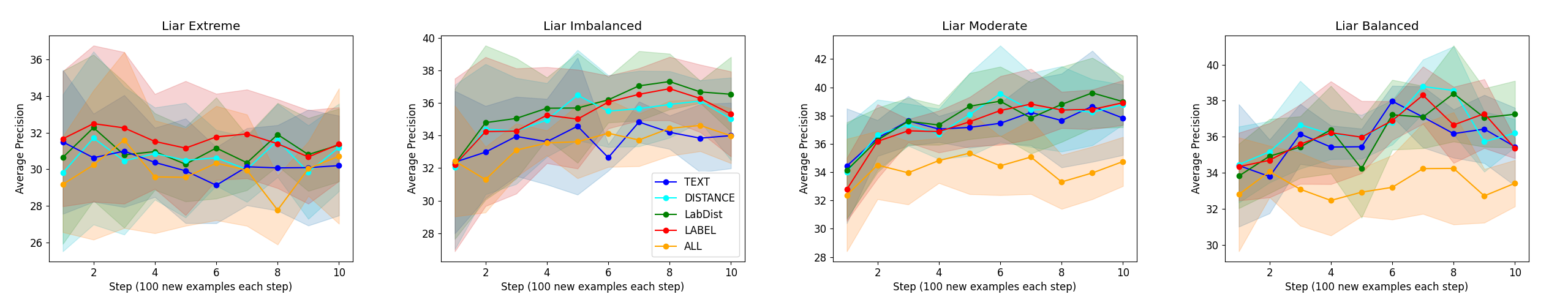}
    \captionof{figure}{\LaGoNNexp performance for all configurations and balance regimes on the LIAR dataset. The relevant balance is in the title of each panel.}
    \label{lagonn_exp_LIAR}
\end{figure*}

Below, Figures \ref{lagonn_cheap_iq} through \ref{lagonn_cheap_LIAR} are the results for the \LaGoNNcheap training strategy, Figures \ref{lagonn_iq} through \ref{lagonn_LIAR} are the results for \LaGoNN, Figures \ref{lagonn_lite_iq} through \ref{lagonn_lite_LIAR} are the results for \LaGoNNlite, and Figures \ref{lagonn_exp_iq} through \ref{lagonn_exp_LIAR} are the results for \LaGoNNexp. We place the figures on a new page for ease of viewing. 

In the case of \LaGoNNcheap, if we do not fine-tune the embedding model we see little variation in the standard deviation bands, with the exception of the LIAR dataset, which seems to be a very difficult dataset. When we do fine-tune, we see a great deal of variation, especially in cases of label imbalance, which is expected as the representations are altered more.
The performance of \textt and \all is very unstable, often being the worst performers, while sometimes being the best. Interestingly, we note that \dist, \labell, and \labdist often show very similar performance. In our opinion. \labdist seems to be the most consistent and stable performer, especially in cases when the embedding model is fine-tuned, \LaGoNN, \LaGoNNlite, and \LaGoNNexp.

Overall, we believe that \labdist is the most performant/stable configuration of \LaGoNN, and it is about this version that we present results in the main text. We note that we could have presented the best performer for each evaluation scenario, however, this is not in the spirit of our work as it adds yet another hyperparameter to configure, standing in the way of practical usage and convoluting our analysis. However, in our codebase, we hope that we have made it easy for one to change these configurations for their own usage, be it scientific or otherwise.




\clearpage
\newpage
\subsubsection{Ablation: \LaGoNN $k$ nearest neighbors}
\label{nns_lagonn}
Here, at the suggestion of an anonymous reviewer, we present ablation results and analysis of searching over one to five nearest neighbors when modifying input via \LaGoNN. We present results over all \LaGoNN configurations under the \LaGoNNlite fine-tuning strategy and with all balance regimes for the content moderation datasets. For the general text classification setting, we present results for both \LaGoNNlite and \LaGoNNexp fine-tuning under the balanced regime for all datasets with the \labdist and \textt configurations.

\begin{figure*}[h!]
    \centering
    \includegraphics[scale=0.3]{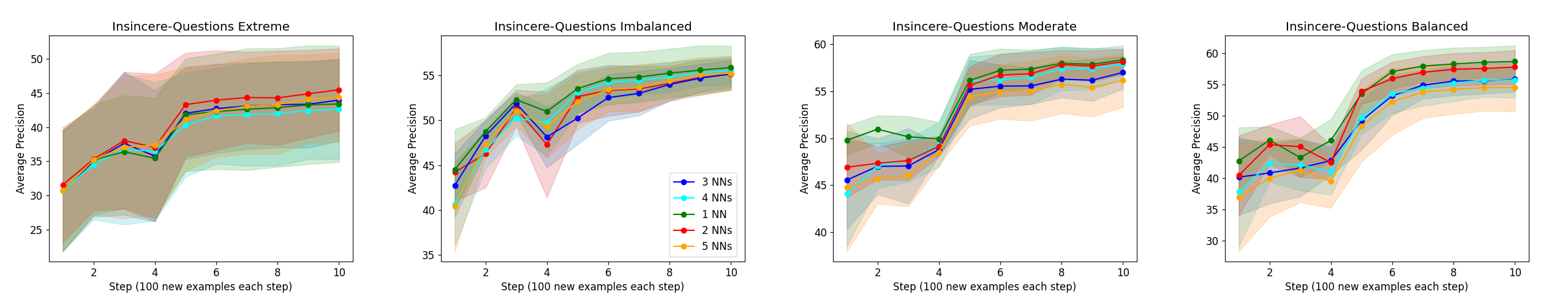}
    \captionof{figure}{\labdist results for one to five neighbors under the \LaGoNNlite fine-tuning strategy on the Insincere Questions dataset. The relevant balance is in the title of each panel.}
    \label{nn_label_iq}
\end{figure*}

\begin{figure*}[h]
    \centering
    \includegraphics[scale=0.3]{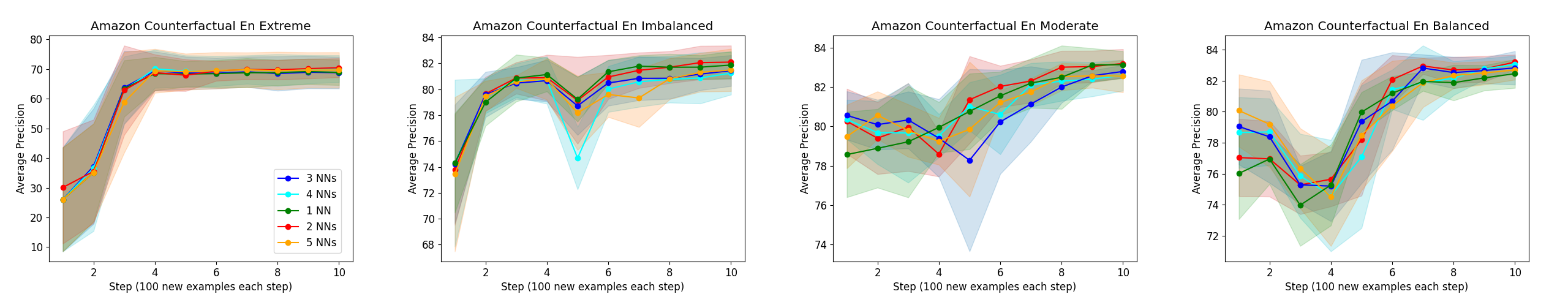}
    \captionof{figure}{\labdist results for one to five neighbors under the \LaGoNNlite fine-tuning strategy on the the Amazon Counterfactual dataset. The relevant balance is in the title of each panel.}
    \label{nn_label_ac}
\end{figure*}

\begin{figure*}[h]
    \centering
    \includegraphics[scale=0.3]{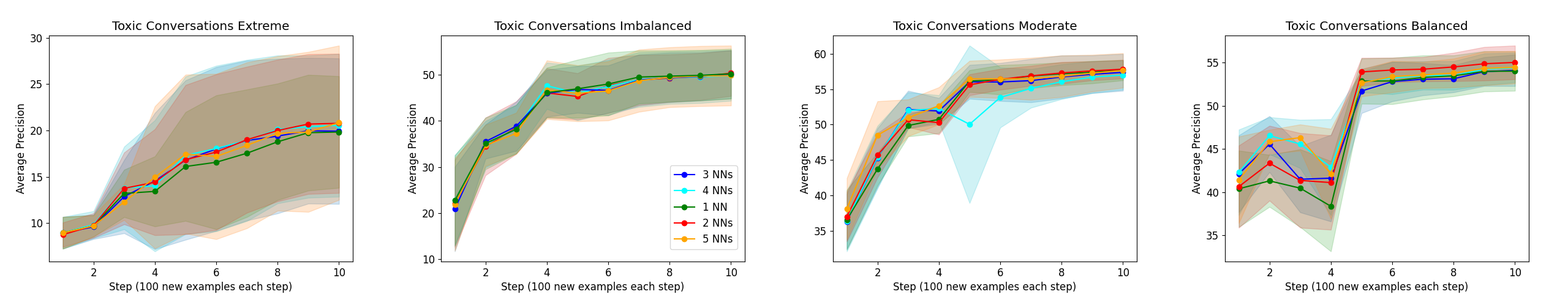}
    \captionof{figure}{\labdist results for one to five neighbors under the \LaGoNNlite fine-tuning strategy on the Toxic Conversations dataset. The relevant balance is in the title of each panel.}
    \label{nn_label_tx}
\end{figure*}

\begin{figure*}[h]
    \centering
    \includegraphics[scale=0.3]{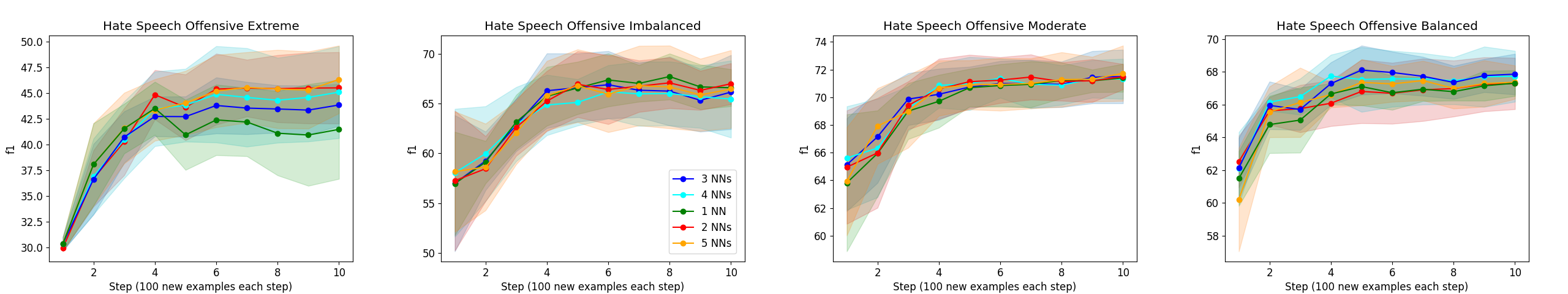}
    \captionof{figure}{\labdist results for one to five neighbors under the \LaGoNNlite fine-tuning strategy on the Hate Speech Offensive dataset. The relevant balance is in the title of each panel.}
    \label{nn_label_hs}
\end{figure*}

\begin{figure*}[h]
    \centering
    \includegraphics[scale=0.3]{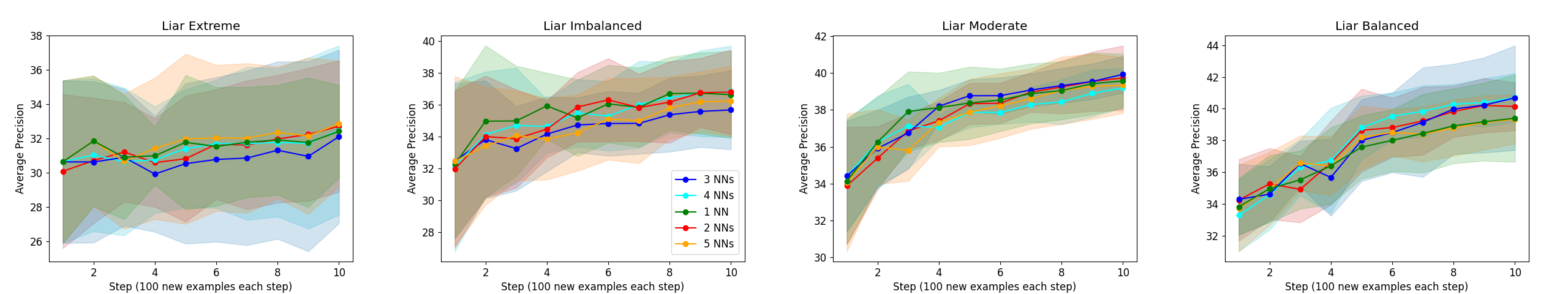}
    \captionof{figure}{\labdist results for one to five neighbors under the \LaGoNNlite fine-tuning strategy on the LIAR dataset. The relevant balance is in the title of each panel.}
    \label{nn_label_LIAR}
\end{figure*}

\begin{figure*}[h!]
    \centering
    \includegraphics[scale=0.3]{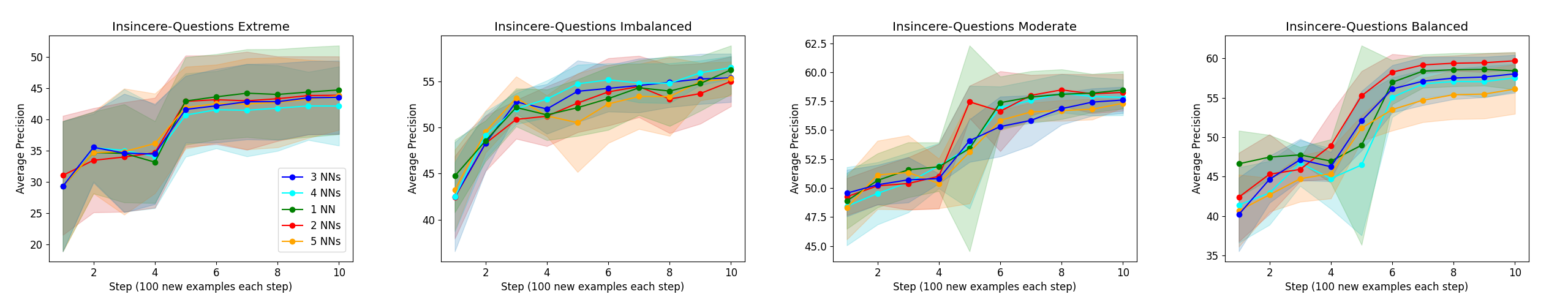}
    \captionof{figure}{\dist results for one to five neighbors under the \LaGoNNlite fine-tuning strategy on the the Insincere Questions dataset. The relevant balance is in the title of each panel.}
    \label{dist_NN_iq}
\end{figure*}

\begin{figure*}[h]
    \centering
    \includegraphics[scale=0.3]{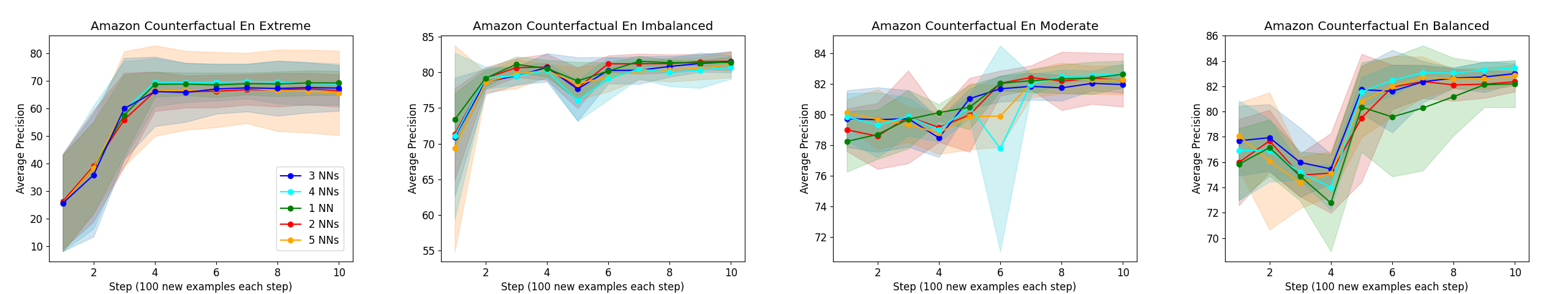}
    \captionof{figure}{\dist results for one to five neighbors under the \LaGoNNlite fine-tuning strategy on the Amazon Counterfactual dataset. The relevant balance is in the title of each panel.}
    \label{dist_nn_ac}
\end{figure*}

\begin{figure*}[h]
    \centering
    \includegraphics[scale=0.3]{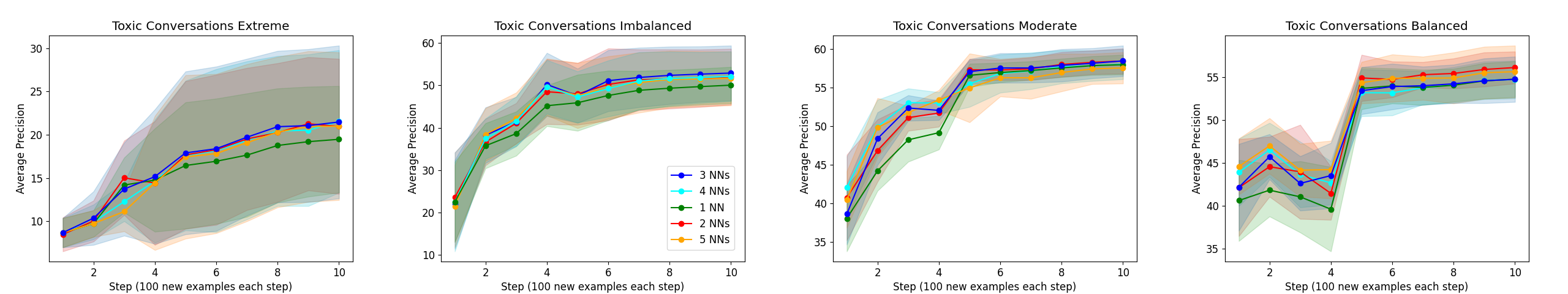}
    \captionof{figure}{\dist results for one to five neighbors under the \LaGoNNlite fine-tuning strategy on the Toxic Conversations dataset. The relevant balance is in the title of each panel.}
    \label{dist_nn_tx}
\end{figure*}

\begin{figure*}[h]
    \centering
    \includegraphics[scale=0.3]{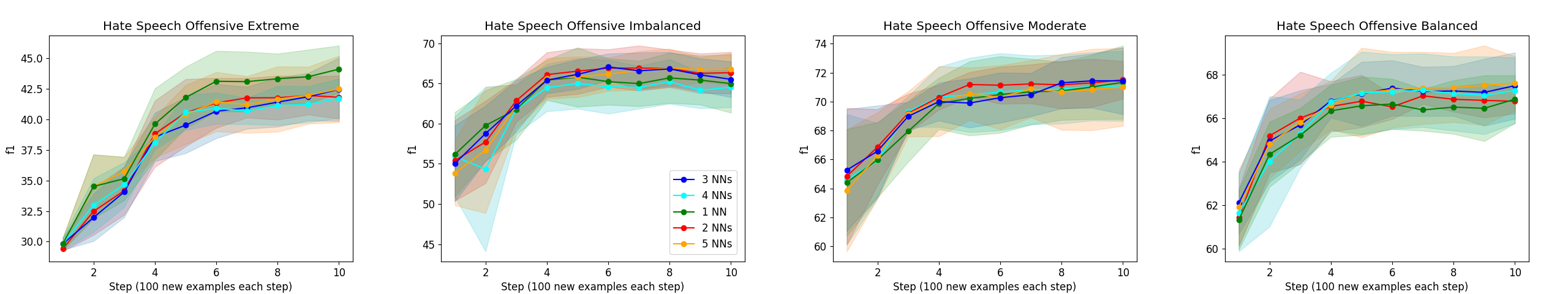}
    \captionof{figure}{\dist results for one to five neighbors under the \LaGoNNlite fine-tuning strategy on the Hate Speech Offensive dataset. The relevant balance is in the title of each panel.}
    \label{dist_nn_hs}
\end{figure*}

\begin{figure*}[h]
    \centering
    \includegraphics[scale=0.3]{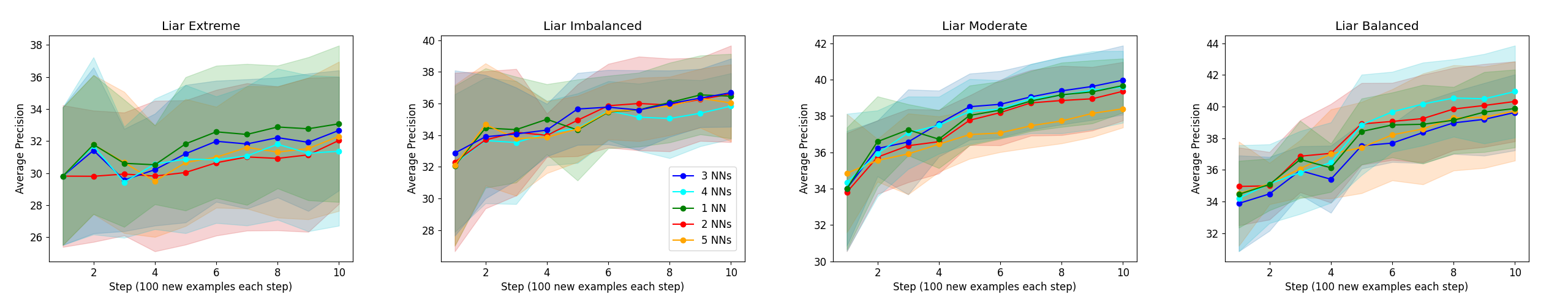}
    \captionof{figure}{\dist results for one to five neighbors under the \LaGoNNlite fine-tuning strategy on the LIAR dataset. The relevant balance is in the title of each panel.}
    \label{dist_nn_LIAR}
\end{figure*}

\begin{figure*}[h!]
    \centering
    \includegraphics[scale=0.3]{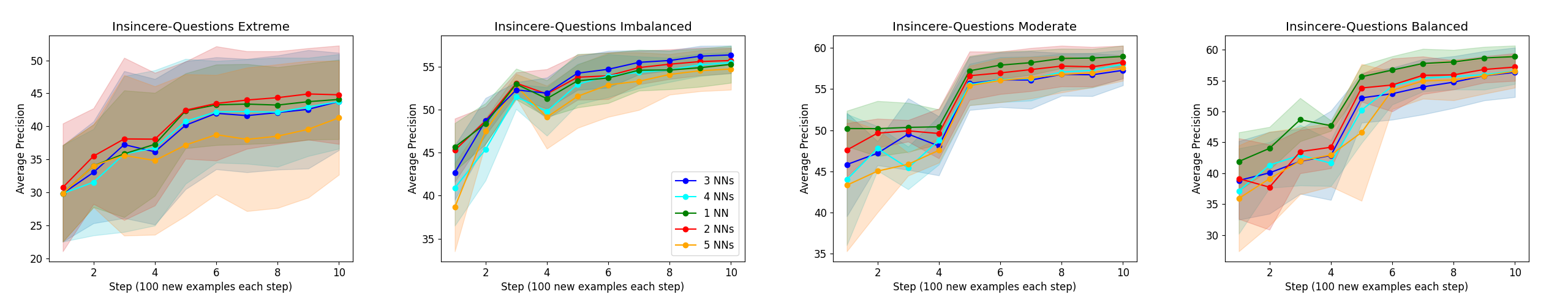}
    \captionof{figure}{\labell results for one to five neighbors under the \LaGoNNlite fine-tuning strategy on the Insincere Questions dataset. The relevant balance is in the title of each panel.}
    \label{only_label_NN_iq}
\end{figure*}

\begin{figure*}[h]
    \centering
    \includegraphics[scale=0.3]{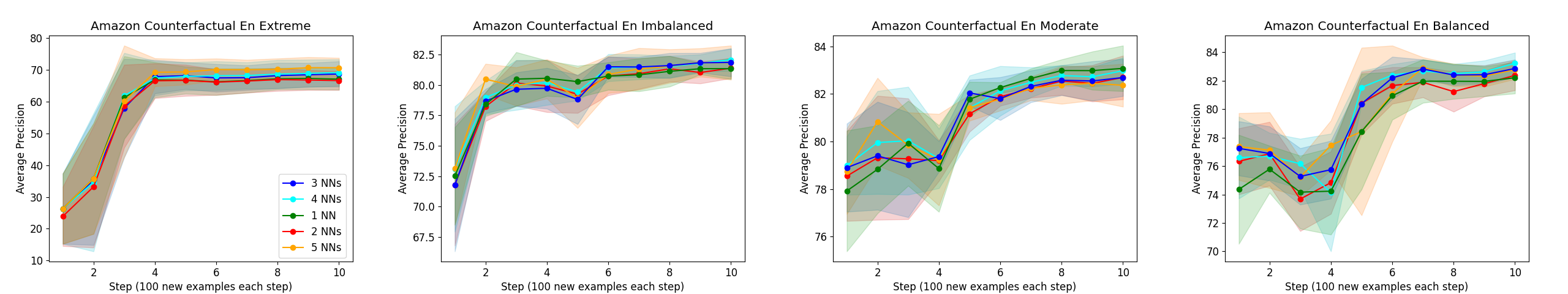}
    \captionof{figure}{\labell results for one to five neighbors under the \LaGoNNlite fine-tuning strategy on the Amazon Counterfactual dataset. The relevant balance is in the title of each panel.}
    \label{only_label_NN_ac}
\end{figure*}

\begin{figure*}[h]
    \centering
    \includegraphics[scale=0.3]{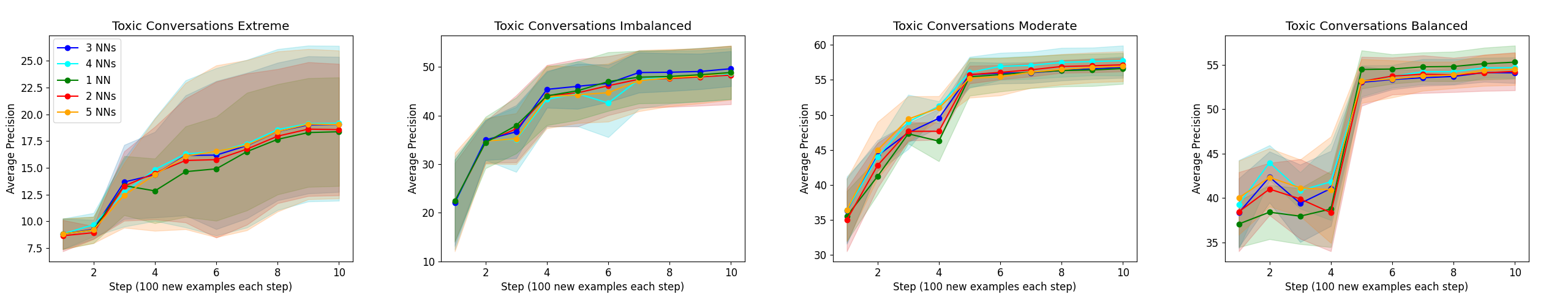}
    \captionof{figure}{\labell results for one to five neighbors under the \LaGoNNlite fine-tuning strategy on the Toxic Conversations dataset. The relevant balance is in the title of each panel.}
    \label{only_label_NN_tx}
\end{figure*}

\begin{figure*}[h]
    \centering
    \includegraphics[scale=0.3]{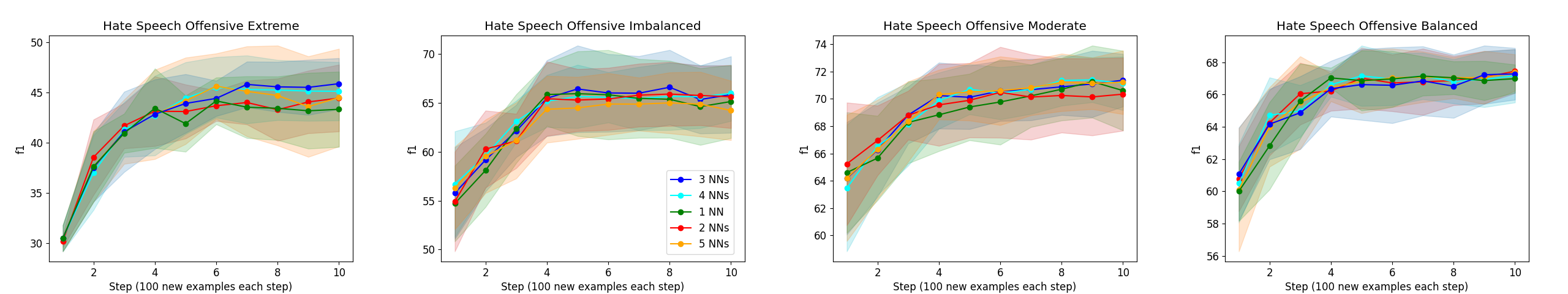}
    \captionof{figure}{\labell results for one to five neighbors under the \LaGoNNlite fine-tuning strategy on the Hate Speech Offensive dataset. The relevant balance is in the title of each panel.}
    \label{only_label_NN_hs}
\end{figure*}

\begin{figure*}[h]
    \centering
    \includegraphics[scale=0.3]{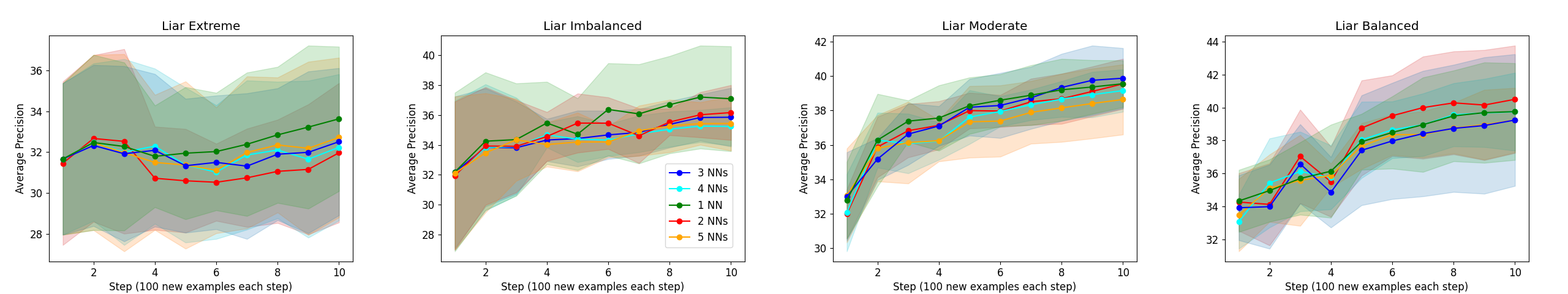}
    \captionof{figure}{\labell results for one to five neighbors under the \LaGoNNlite fine-tuning strategy on the LIAR dataset. The relevant balance is in the title of each panel.}
    \label{only_label_NN_LIAR}
\end{figure*}

\begin{figure*}[h!]
    \centering
    \includegraphics[scale=0.3]{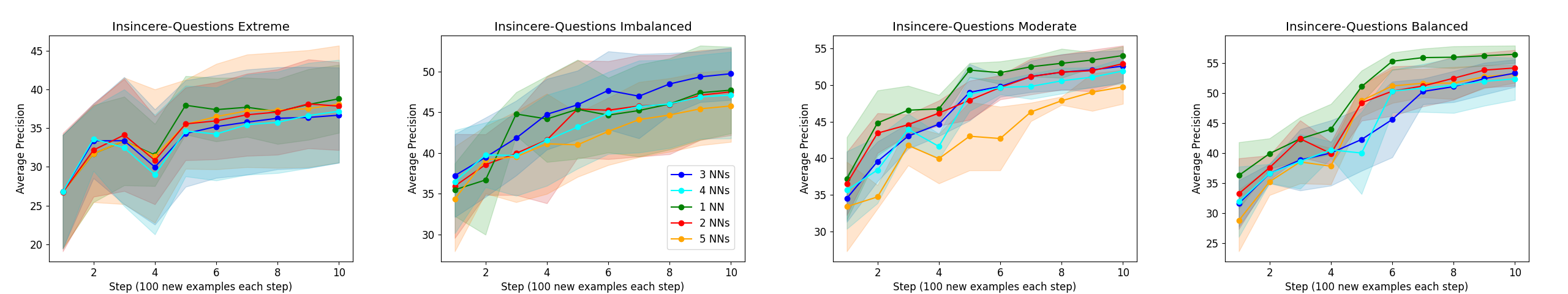}
    \captionof{figure}{\textt results for one to five neighbors under the \LaGoNNlite fine-tuning strategy on the Insincere Questions dataset. The relevant balance is in the title of each panel.}
    \label{text_nn_iq}
\end{figure*}

\begin{figure*}[h]
    \centering
    \includegraphics[scale=0.3]{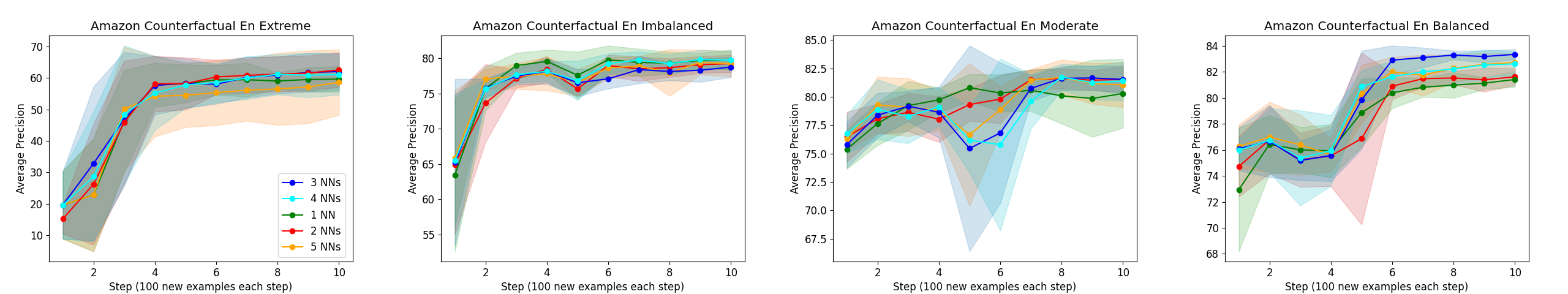}
    \captionof{figure}{\textt results for one to five neighbors under the \LaGoNNlite fine-tuning strategy on the Amazon Counterfactual dataset. The relevant balance is in the title of each panel.}
    \label{text_nn_ac}
\end{figure*}

\begin{figure*}[h]
    \centering
    \includegraphics[scale=0.3]{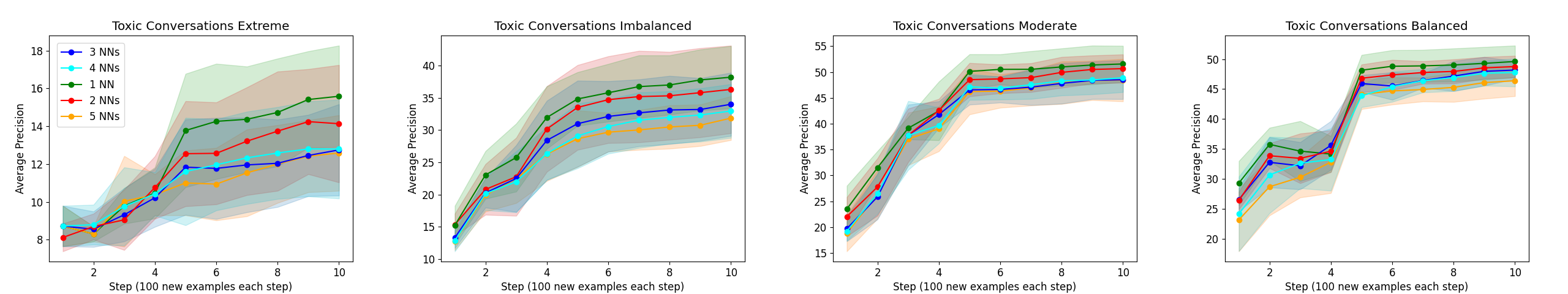}
    \captionof{figure}{\textt results for one to five neighbors under the \LaGoNNlite fine-tuning strategy on the Toxic Conversations dataset. The relevant balance is in the title of each panel.}
    \label{text_nn_tx}
\end{figure*}

\begin{figure*}[h]
    \centering
    \includegraphics[scale=0.3]{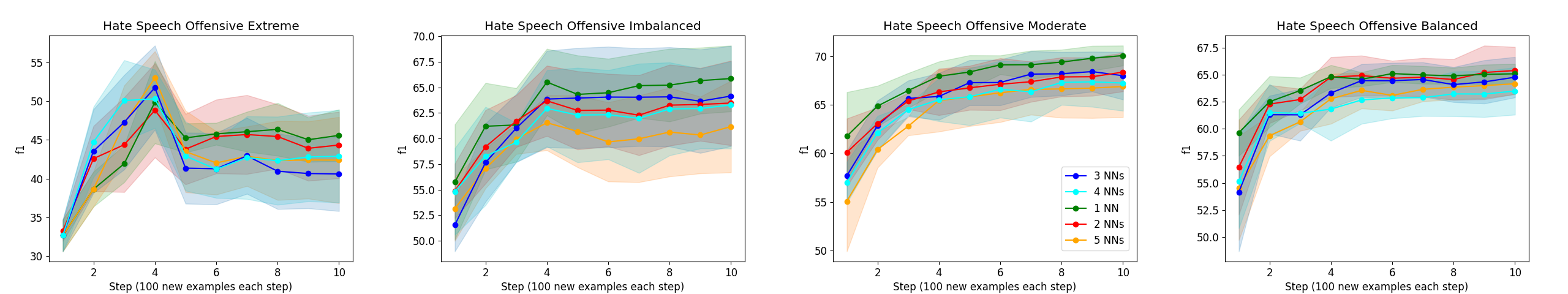}
    \captionof{figure}{\textt results for one to five neighbors under the \LaGoNNlite fine-tuning strategy on the Hate Speech Offensive dataset. The relevant balance is in the title of each panel.}
    \label{text_nn_hs}
\end{figure*}

\begin{figure*}[h]
    \centering
    \includegraphics[scale=0.3]{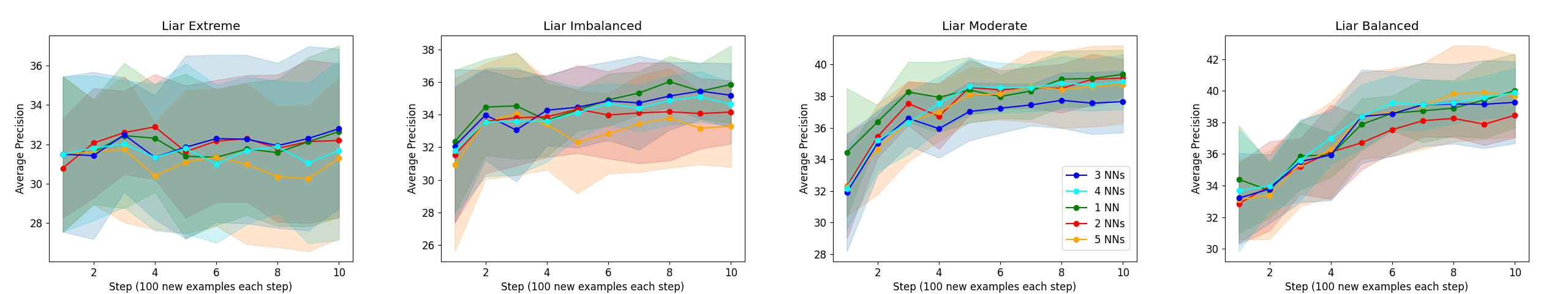}
    \captionof{figure}{\textt results for one to five neighbors under the \LaGoNNlite fine-tuning strategy on the LIAR dataset. The relevant balance is in the title of each panel.}
    \label{ltext_nn_LIAR}
\end{figure*}

\begin{figure*}[h!]
    \centering
    \includegraphics[scale=0.3]{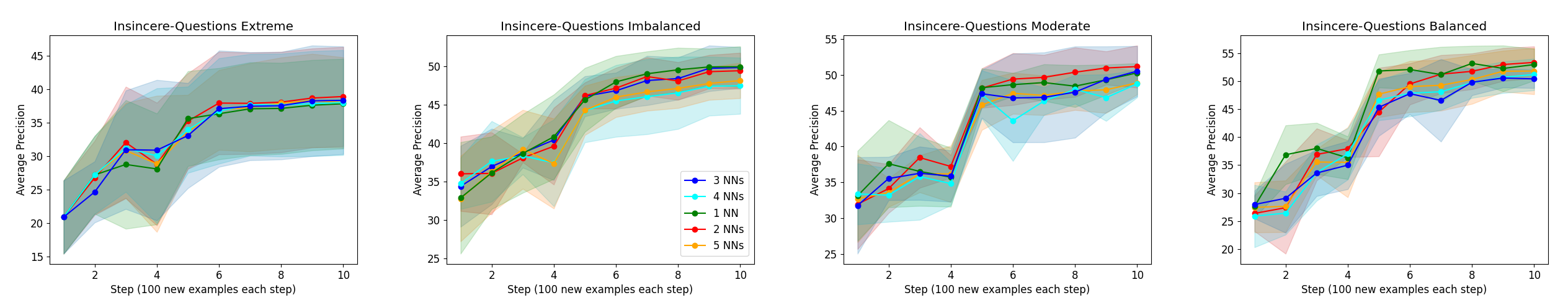}
    \captionof{figure}{\all results for one to five neighbors under the \LaGoNNlite fine-tuning strategy on the Insincere Questions dataset. The relevant balance is in the title of each panel.}
    \label{BOTH__nn_iq}
\end{figure*}

\begin{figure*}[h]
    \centering
    \includegraphics[scale=0.3]{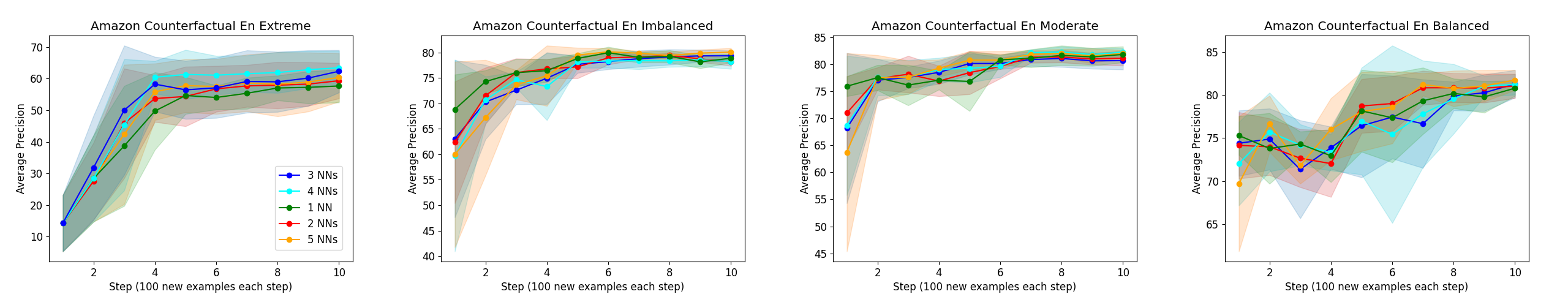}
    \captionof{figure}{\all results for one to five neighbors under the \LaGoNNlite fine-tuning strategy on the Amazon Counterfactual dataset. The relevant balance is in the title of each panel.}
    \label{BOTH__nn_ac}
\end{figure*}

\begin{figure*}[h]
    \centering
    \includegraphics[scale=0.3]{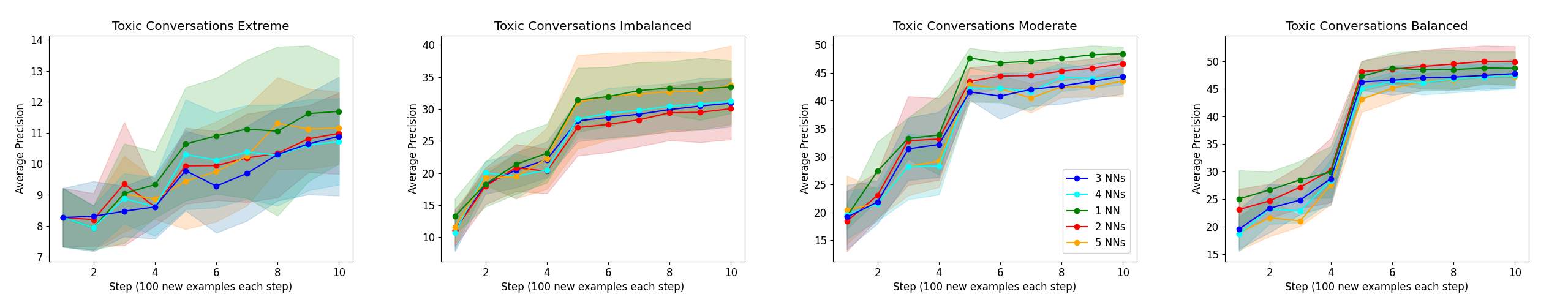}
    \captionof{figure}{\all results for one to five neighbors under the \LaGoNNlite fine-tuning strategy on the Toxic Conversations dataset. The relevant balance is in the title of each panel.}
    \label{BOTH__nn_tx}
\end{figure*}

\begin{figure*}[h]
    \centering
    \includegraphics[scale=0.3]{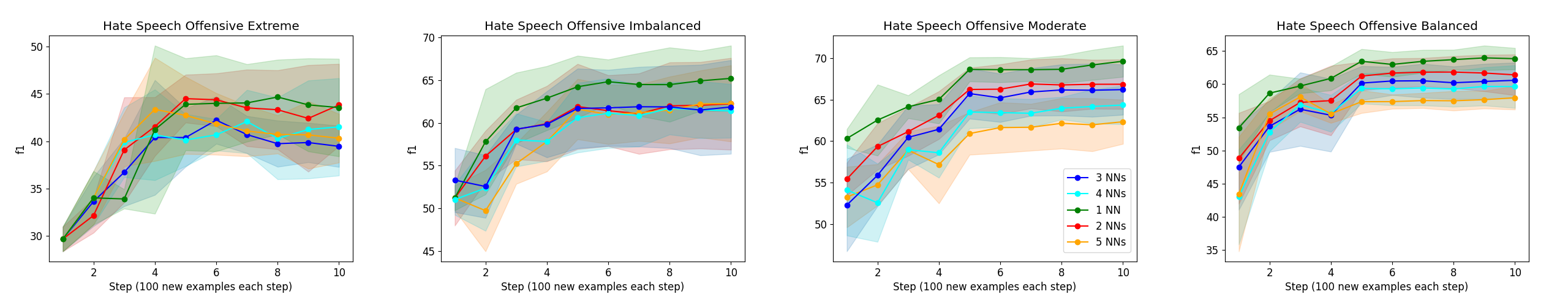}
    \captionof{figure}{\all results for one to five neighbors under the \LaGoNNlite fine-tuning strategy on the Hate Speech Offensive dataset. The relevant balance is in the title of each panel.}
    \label{BOTH__nn_hs}
\end{figure*}

\begin{figure*}[h]
    \centering
    \includegraphics[scale=0.3]{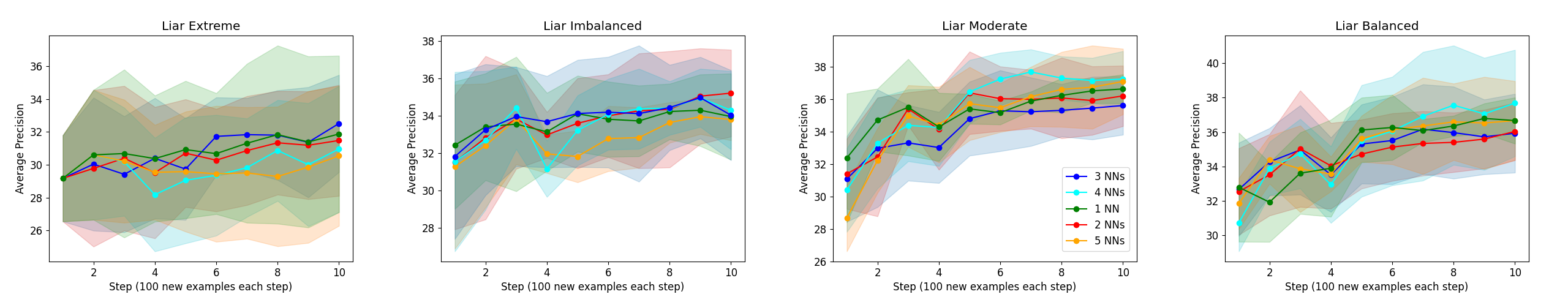}
    \captionof{figure}{\all results for one to five neighbors under the \LaGoNNlite fine-tuning strategy on the LIAR dataset. The relevant balance is in the title of each panel.}
    \label{BOTH__nn_LIAR}
\end{figure*}

If we consider all \LaGoNN configurations and balance regimes in the case content moderation, Figures \ref{nn_label_iq} through \ref{BOTH__nn_LIAR}, the number of neighbors does not appear to be an important hyperparameter; the learning curves for a given dataset and balance regime are very similar. While there is variation, the trend appears to be that the first NN results in the stablest, most performant, and most consistent model. 

However, if we only focus on \labdist (Figures \ref{nn_label_iq} through \ref{nn_label_LIAR}), the default \LaGoNN configuration, we see that it can be a very important hyperparameter to consider in cases of extreme imbalance or when we have balanced data but few data points. For example, performance is boosted by up to five points for Hate Speech Offensive by the tenth step (1000 examples) with five neighbors under the extreme balance regime, yet for the balanced regime, the performance curves are roughly the same. For Toxic Conversations, in the balanced regime, we see that we can increase performance by up to seven points on the second step (200 examples) by considering more neighbors.

Turning our attention now to the general classification experiments, we see that the number of neighbors for both the \labdist and \textt configurations continues to consistently not really make much of a difference, with all models showing very similar performance curves for all datasets. We note however that \labdist appears to be the most performant configuration of our method. While continued fine-tuning on datasets with a large number of labels does increase performance, we observe essentially the same boost for all neighbors. We also observe similar instability and performance degradation when we fine-tune on a large number of examples in cases when we have few labels.




\begin{figure*}[h]
    \centering
    \includegraphics[scale=0.43]{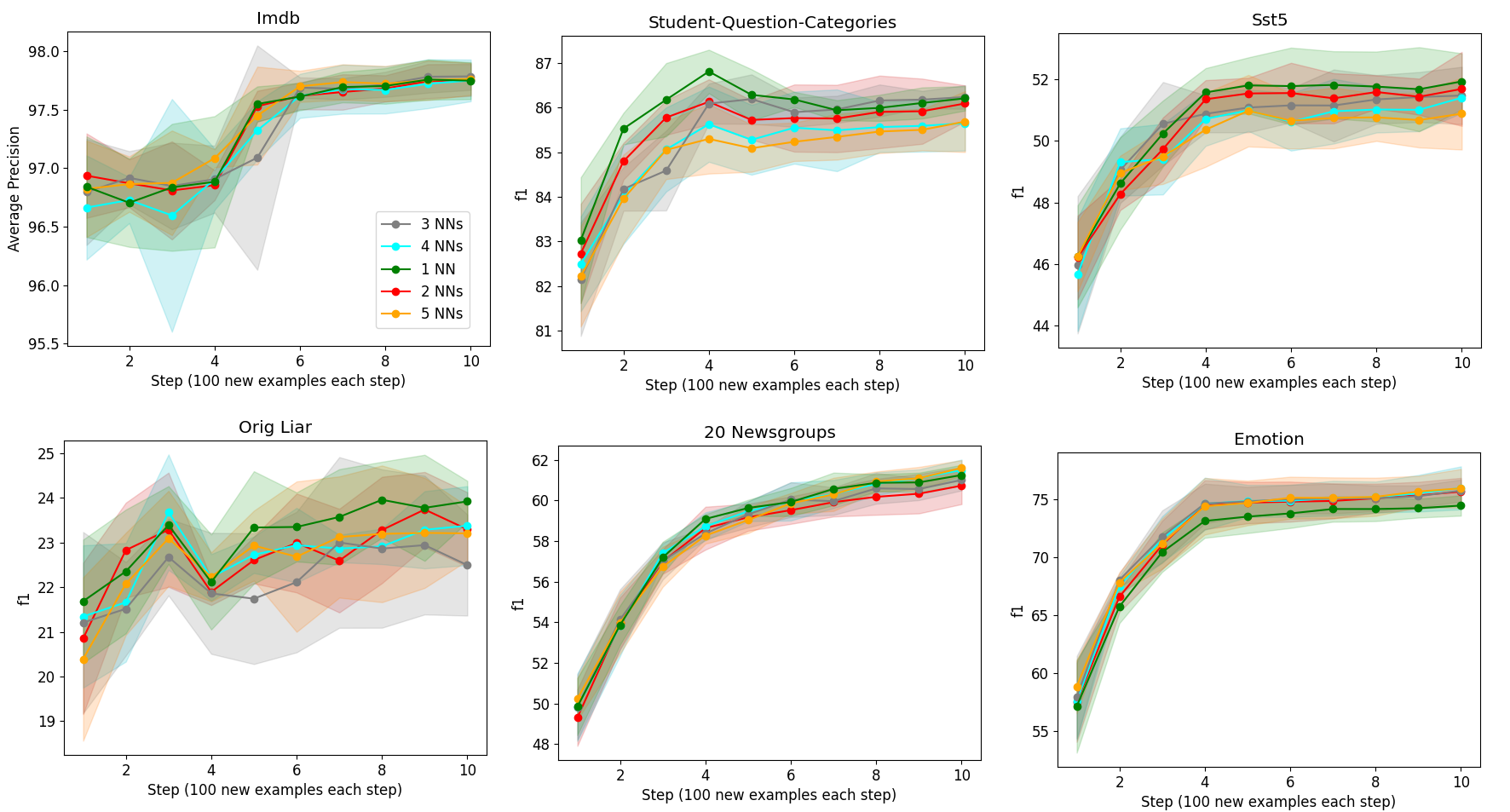}
    \captionof{figure}{\labdist results for one to five neighbors under the \LaGoNNlite fine-tuning strategy over all six general classification datasets. Results are for the balanced sampling regime and the measure is average precision for IMDB, macro-F1 elsewhere.}
    \label{nn_label_lite}
\end{figure*}
\begin{figure*}[h]
    \centering
    \includegraphics[scale=0.43]{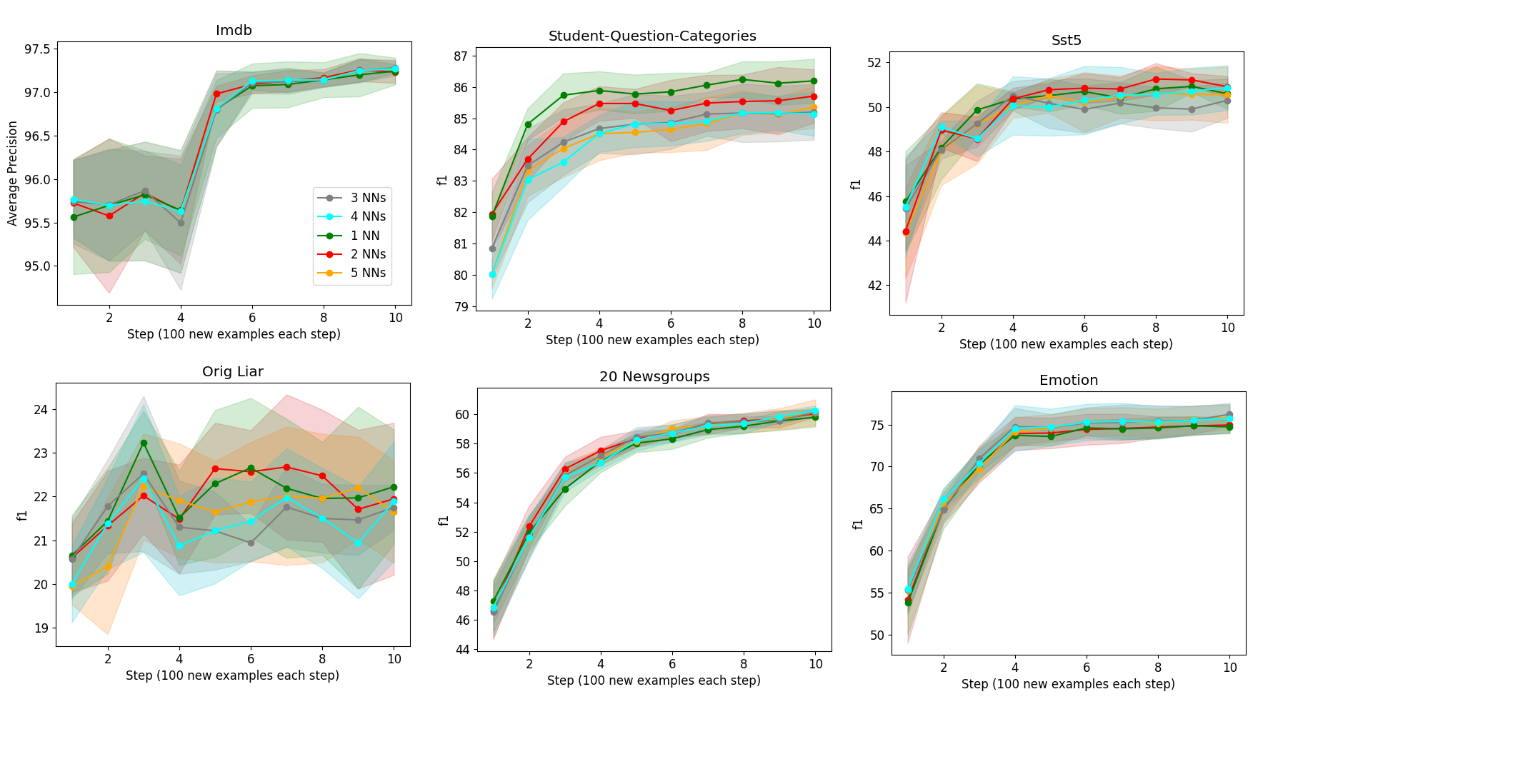}
    \captionof{figure}{\textt results for one to five neighbors under the \LaGoNNlite fine-tuning strategy over all six general classification datasets. Results are for the balanced sampling regime and the measure is average precision for IMDB, macro-F1 elsewhere.}
    \label{nn_text_lite}
\end{figure*}
\begin{figure*}[h]
    \centering
    \includegraphics[scale=0.43]{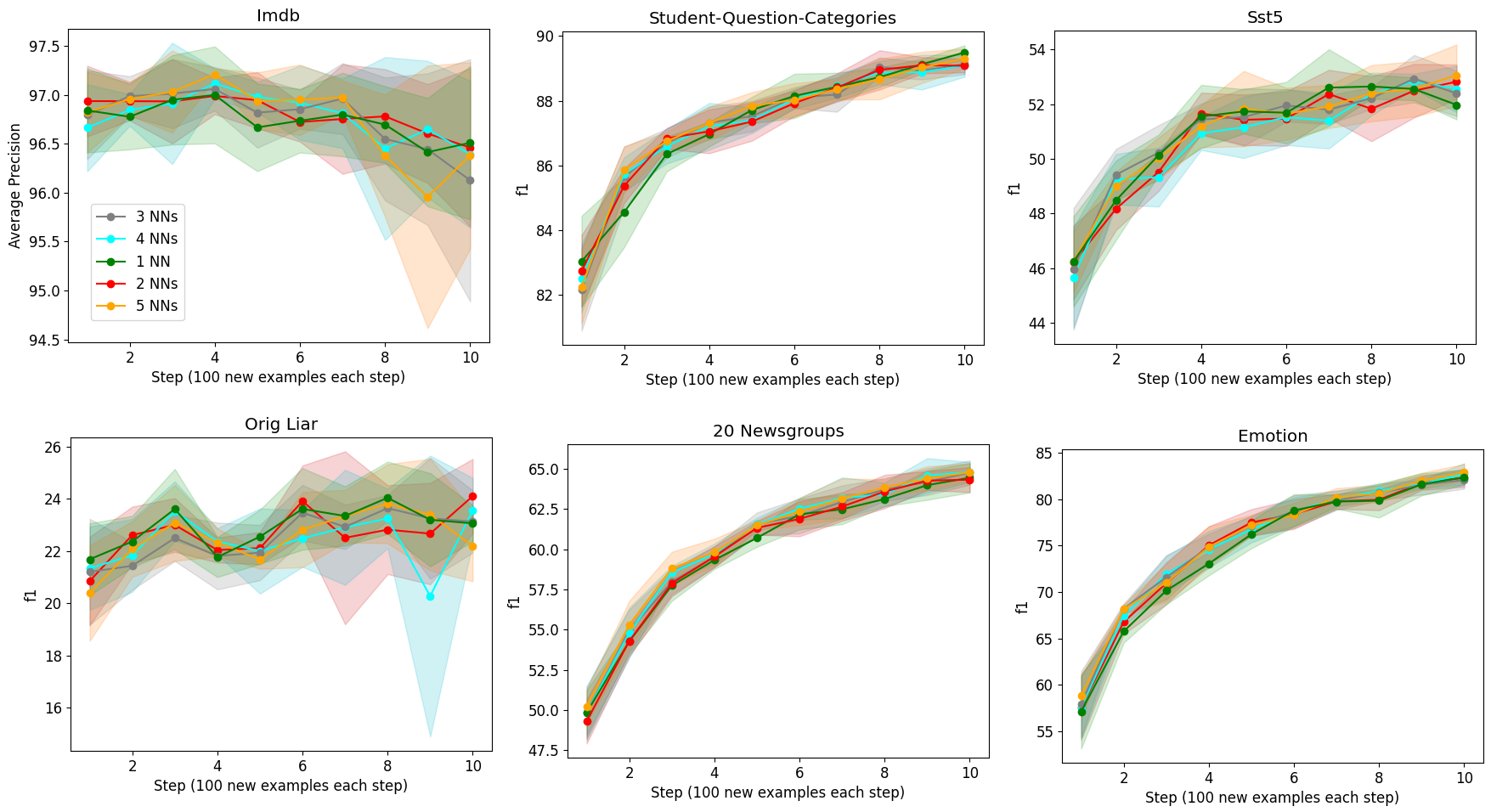}
    \captionof{figure}{\labdist results for one to five neighbors under the \LaGoNNexp fine-tuning strategy over all six general classification datasets. Results are for the balanced sampling regime and the measure is average precision for IMDB, macro-F1 elsewhere.}
    \label{nn_label_exp}
\end{figure*}
\begin{figure*}[h]
    \centering
    \includegraphics[scale=0.43]{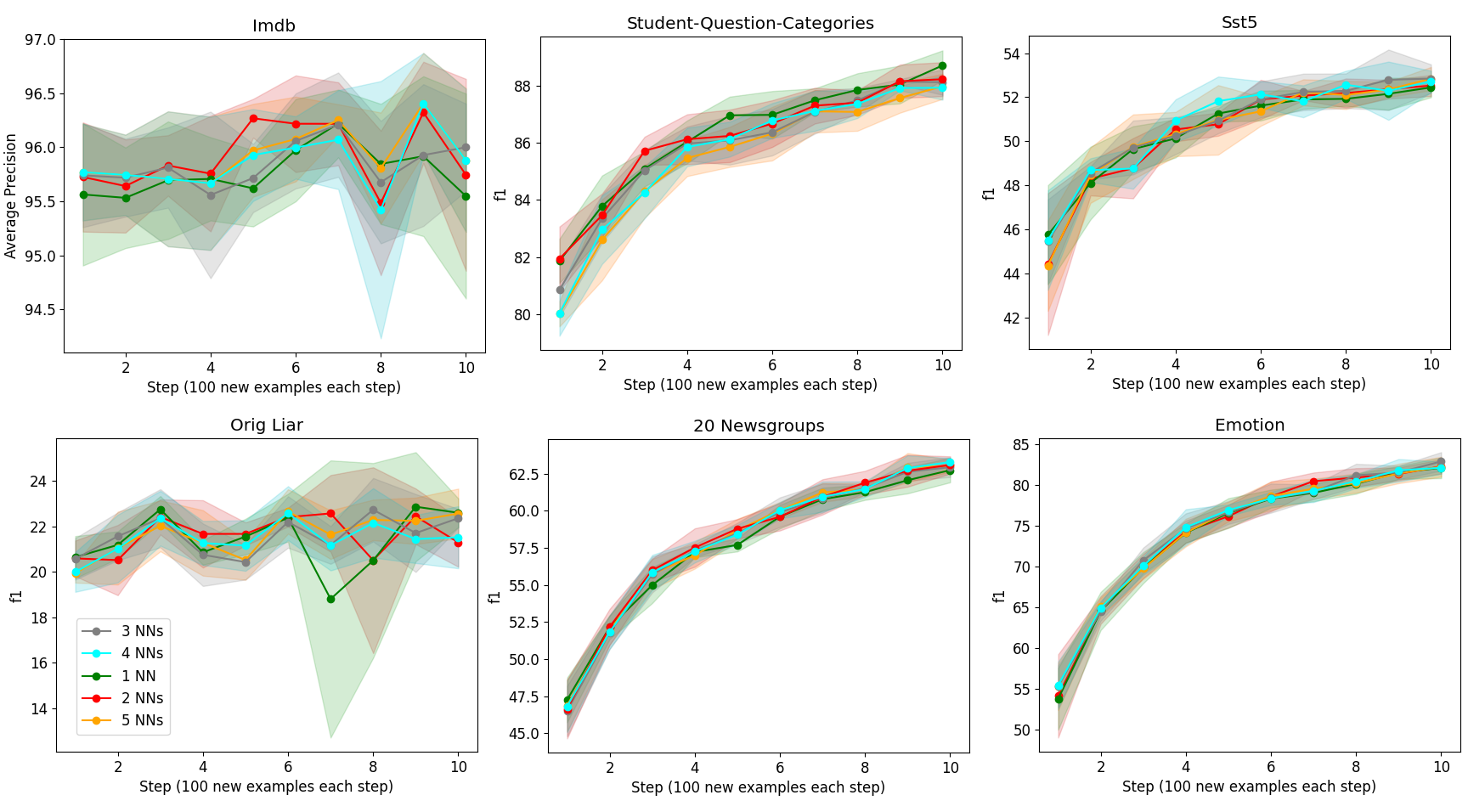}
    \captionof{figure}{\textt results for one to five neighbors under the \LaGoNNexp fine-tuning strategy over all six general classification datasets. Results are for the balanced sampling regime and the measure is average precision for IMDB, macro-F1 elsewhere.}
    \label{nn_text_exp}
\end{figure*}
\clearpage
\newpage

\subsubsection{Ablation: the effect of encoding distance}
\label{encodedist}
Here, at the suggestion of an anonymous reviewer, we present ablation results and analysis of how encoding distance affects \LaGoNN, because PLMs often struggle to understand numbers. Note that during our development stage, we ensured that our tokenizer was capable of encoding floats with trailing digits. To examine the effect of trailing digits on \LaGoNN, we consider the \dist configuration (see Table \ref{table:LaGoNNtypes}), where we append only the Euclidean distance to the input text. In this ablation, however, we round to different levels of precision. For example, if the distance were a float of 0.123456789, we round it to the nearest whole number, 0.0, single digit float, 0.1, three digit float, 0.123, six digit float, 0.123457, and finally keep it unrounded, that is, the original \dist configuration, 0.123456789. The below results are only for the \LaGoNNlite training strategy. We chose \LaGoNNlite for this ablation because it provides insight into both how distance affects full-model fine-tuning and only refitting the classification head. The results can be seen below in Figures \ref{dist_iq} through \ref{dist_LIAR}. We place the figures on a new page for ease of viewing.

Interestingly, we tend to observe very similar performance curves for all rounding precisions. The exceptions to this would perhaps be Amazon Counterfactual and Hate Speech Offensive in the balanced regime where \dist and rounding to the third trailing digit respectively exhibit large instability.

Although not always the case, it appears that providing the model with the distance rounded to the nearest whole number tends to result in the strongest and stablest performer, however, we emphasize that in general there does not seem to a dramatic difference between the rounding precisions we considered. Longer digits slightly worsen model performance and the model might learn the most from simpler or abbreviated representations of distance. This finding motivated us to consider the ablation in Appendix \ref{supportlabdist}.

\begin{figure*}[h!]
    \centering
    \includegraphics[scale=0.24]{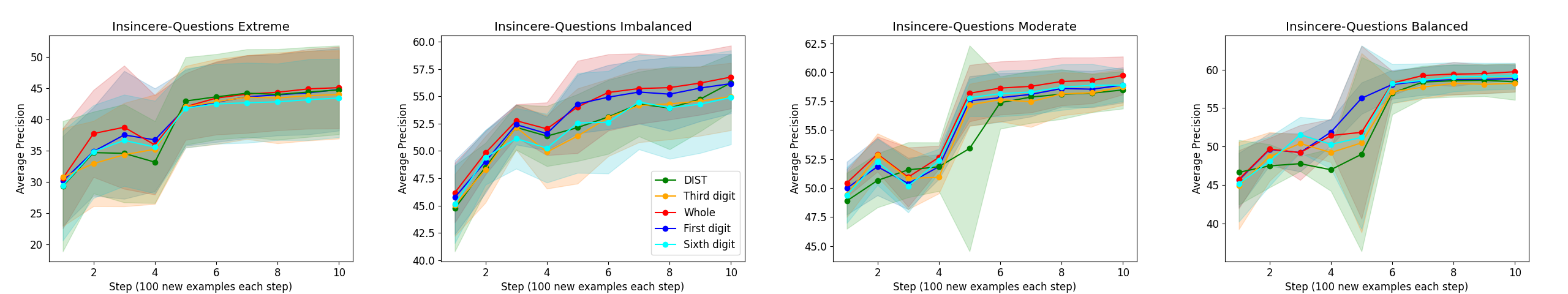}
    \captionof{figure}{\LaGoNNlite performance when considering different rounding precisions for the Euclidean distance before appending it to a modified instance. We consider all balance regimes on the Insincere Questions dataset and the relevant balance is in the title of each panel.}
    \label{dist_iq}
\end{figure*}
\begin{figure*}[h]
    \centering
    \includegraphics[scale=0.24]{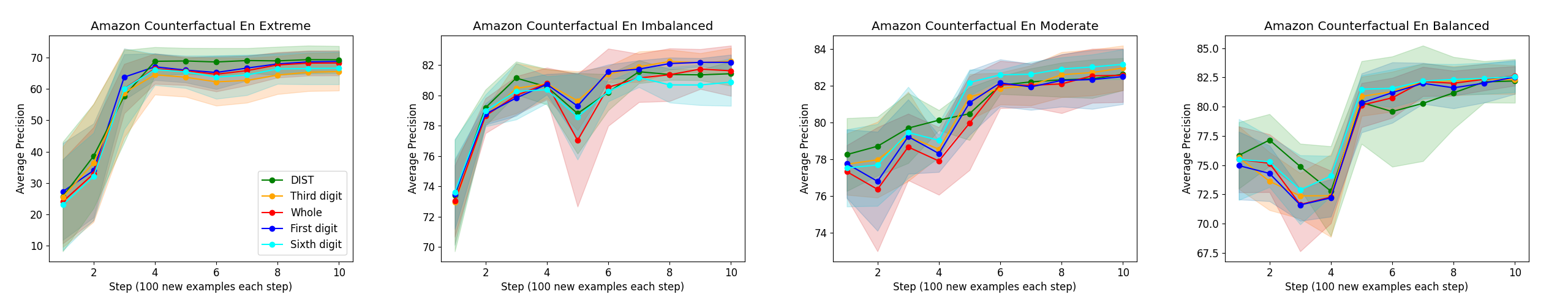}
    \captionof{figure}{\LaGoNNlite performance when considering different rounding precisions for the Euclidean distance before appending it to a modified instance. We consider all balance regimes on the Amazon Counterfactual dataset and the relevant balance is in the title of each panel.}
    \label{dist_ac}
\end{figure*}
\begin{figure*}[h]
    \centering
    \includegraphics[scale=0.24]{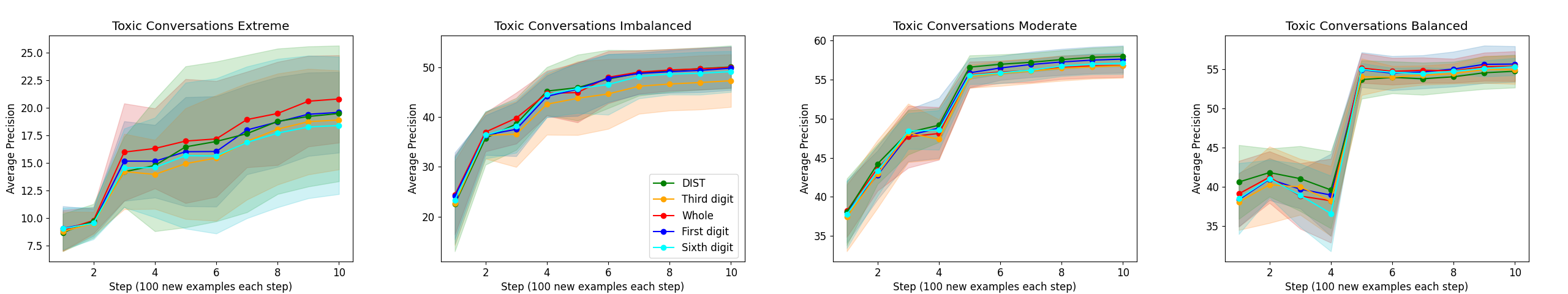}
    \captionof{figure}{\LaGoNNlite performance when considering different rounding precisions for the Euclidean distance before appending it to a modified instance. We consider all balance regimes on the Toxic Conversations dataset and the relevant balance is in the title of each panel.}
    \label{dist_tx}
\end{figure*}
\begin{figure*}[h]
    \centering
    \includegraphics[scale=0.24]{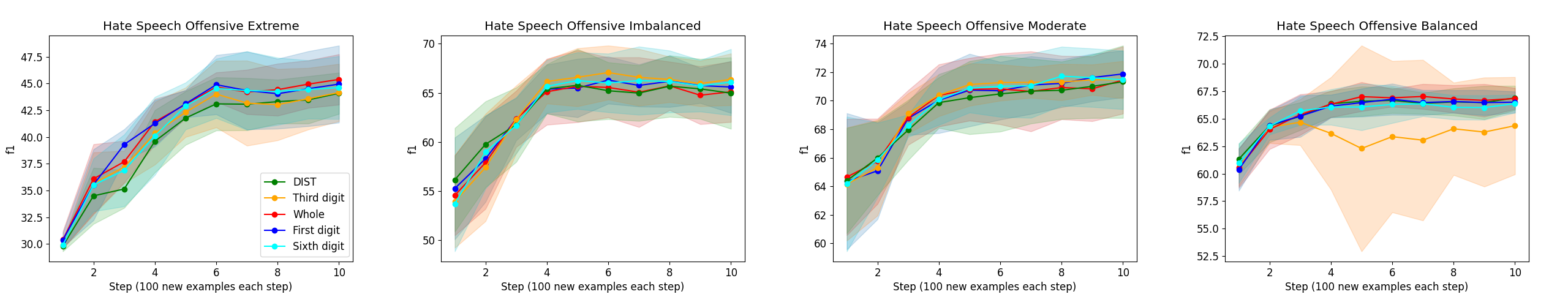}
    \captionof{figure}{\LaGoNNlite performance when considering different rounding precisions for the Euclidean distance before appending it to a modified instance. We consider all balance regimes on the Hate Speech Offensive dataset and the relevant balance is in the title of each panel.}
    \label{dist_hs}
\end{figure*}
\begin{figure*}[h]
    \centering
    \includegraphics[scale=0.24]{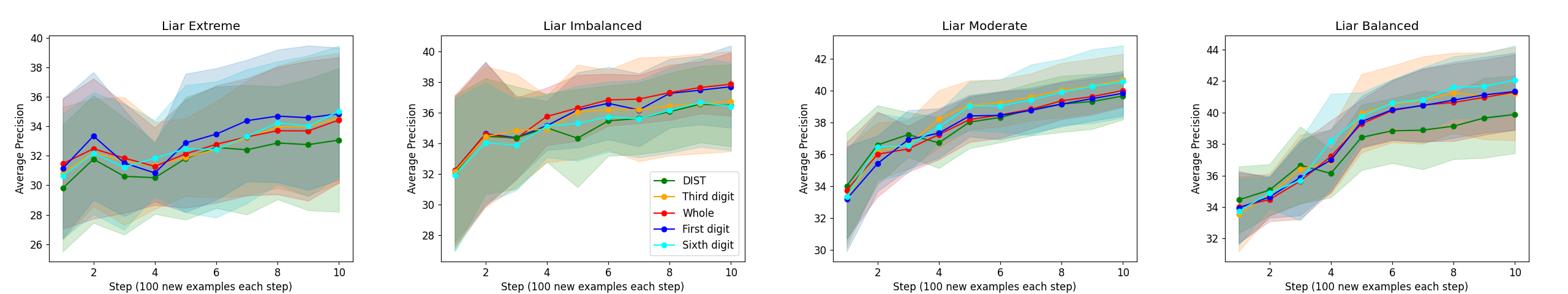}
    \captionof{figure}{\LaGoNNlite performance when considering different rounding precisions for the Euclidean distance before appending it to a modified instance. We consider all balance regimes on the LIAR dataset and the relevant balance is in the title of each panel.}
    \label{dist_LIAR}
\end{figure*}
\clearpage
\newpage

\subsubsection{Ablation: support for \labdist}
\label{supportlabdist}
The results from the ablation in Appendix \ref{encodedist} suggest that rounding the distance to the nearest whole number results in a stronger classifier than appending the unrounded distance. Thus far, we have asserted that \labdist, where we append both the gold label of the NN and unrounded distance is the most performant version of \LaGoNN (see Table \ref{table:LaGoNNtypes}). To demonstrate that this is reasonable, in this ablation study, we compare the original \labdist configuration against three models, namely the \labell configuration, distance rounded to near whole number (Whole), and finally a new configuration similar to \labdist, but where we append the gold label and distance rounded to a whole number, which we refer to as \labround. As in Appendix \ref{encodedist}, in this ablation we consider only the \LaGoNNlite fine-tuning strategy. We chose \LaGoNNlite for this ablation because it provides insight into both how the different configurations affect full-model fine-tuning and only re-fitting the classification head. The results can be seen below in Figures \ref{labdist_iq} through \ref{labdist_LIAR}. We place the figures on a new page for ease of viewing.

In general, we note very similar performance curves for these four models. In the case of Insincere Questions, appending the distance after rounding it to the nearest whole number (Whole, the red curve), is a strong model, except in the balanced regime where we note large instability.  The results for Amazon Counterfactual tell a different story, where rounding the Euclidean distance to the nearest whole number causes large instability and even degrades performance on the fifth step. 

For the other evaluation scenarios, it is unclear what is the strongest method as sometimes \labdist is the best performer and sometimes it is Whole (the red curve). However, we believe that in general \labdist is the most stable model while also often being the most performant. We therefore choose it as our default \LaGoNN configuration as a compromise between strength and stability. It is about this configuration which we report results in the main text. Our interpretation of this is that passing the model both a discrete prediction (the gold label of the NN) and a truly continuous measure of similarity (the unrounded Euclidean distance) gives it the most consistent and dependable reasoning ability.

We note, as we did in Appendix \ref{sec:appendix_configs}, that we could have presented the best performer for each evaluation scenario, however, it is not the goal of our work to create even more hyperparameters that must be iterated over. However, we hope that our codebase has made it easy for one to change these configurations for their own purposes.

\begin{figure*}[h!]
    \centering
    \includegraphics[scale=0.24]{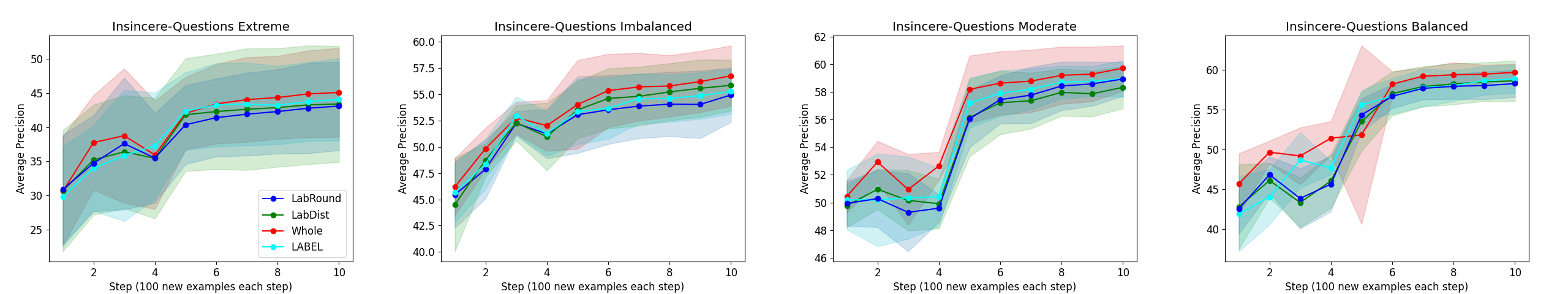}
    \captionof{figure}{\LaGoNNlite performance where we compare the \labdist against \labell, \labround, and rounding the distance to the nearest whole number.
 We consider all balance regimes on the Insincere Questions dataset and the relevant balance is in the title of each panel.}
    \label{labdist_iq}
\end{figure*}
\begin{figure*}[h]
    \centering
    \includegraphics[scale=0.24]{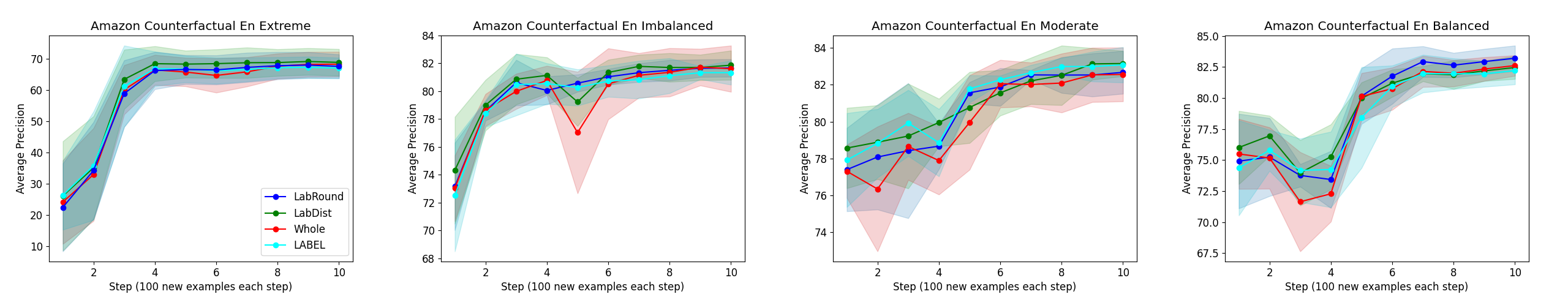}
    \captionof{figure}{\LaGoNNlite performance where we compare the \labdist against \labell, \labround, and rounding the distance to the nearest whole number.
 We consider all balance regimes on the Amazon Counterfactual dataset and the relevant balance is in the title of each panel.}
    \label{labdist_ac}
\end{figure*}
\begin{figure*}[h]
    \centering
    \includegraphics[scale=0.24]{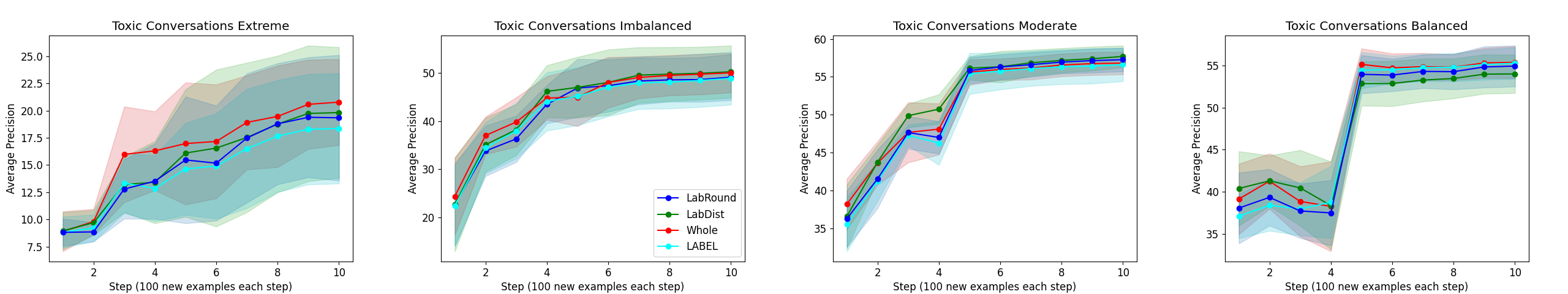}
    \captionof{figure}{\LaGoNNlite performance where we compare the \labdist against \labell, \labround, and rounding the distance to the nearest whole number.
 We consider all balance regimes on the Toxic Conversations dataset and the relevant balance is in the title of each panel.}
    \label{labdist_tx}
\end{figure*}
\begin{figure*}[h]
    \centering
    \includegraphics[scale=0.24]{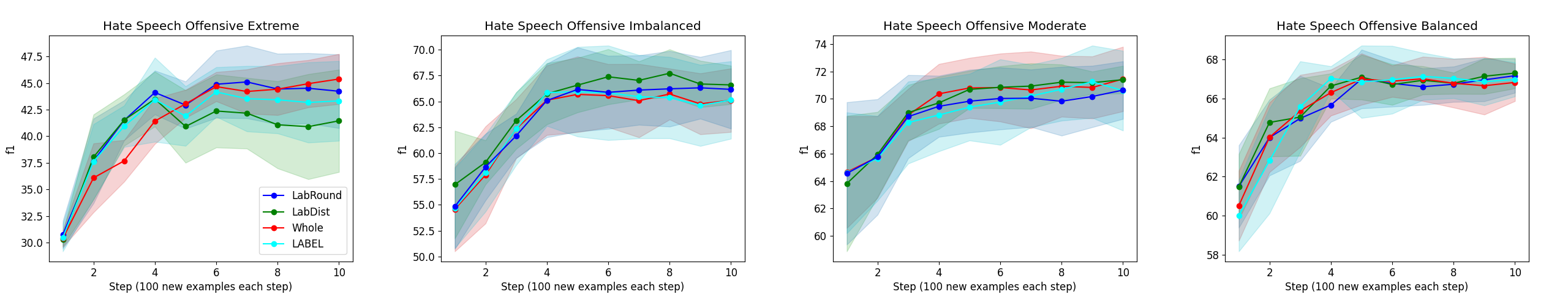}
    \captionof{figure}{\LaGoNNlite performance where we compare the \labdist against \labell, \labround, and rounding the distance to the nearest whole number. We consider all balance regimes on the Hate Speech Offensive dataset and the relevant balance is in the title of each panel.}
    \label{labdist_hs}
\end{figure*}
\begin{figure*}[h]
    \centering
    \includegraphics[scale=0.24]{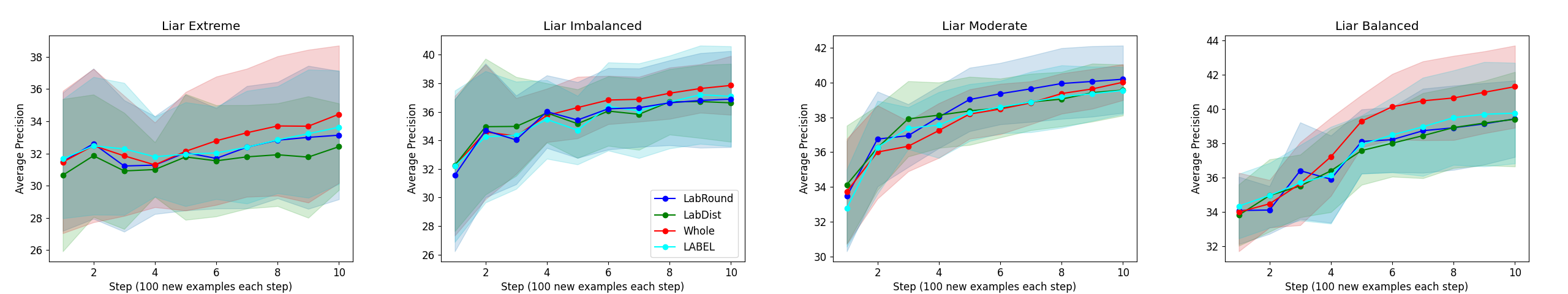}
    \captionof{figure}{\LaGoNNlite performance where we compare the \labdist against \labell, \labround, and rounding the distance to the nearest whole number.
 We consider all balance regimes on the LIAR dataset and the relevant balance is in the title of each panel.}
    \label{labdist_LIAR}
\end{figure*}
\clearpage
\newpage

\subsection{Examples of \LaGoNN modified text}
\label{sec:appendixmod}

\textbf{WARNING:} Some of the examples below are of an offensive nature. Please view with caution.

In this section, we provide examples of how \LaGoNNexp modifies test text from the content moderation datasets we studied under the \all configuration. We choose this configuration because the information it appends from a NN in the training data to a test instance encapsulates all configurations. \LaGoNNexp was trained under a balanced distribution and five examples per label were chosen randomly on the first, fifth, and tenth step to demonstrate how the same test instance might be decorated with different training examples as the training data grow. We have made the .csv files available with our code and data files. In order not to break our .pdf generator, we were forced to remove a handful of symbols from the below text, but the original modifications remain in-tact in the .csv files included with our code files. Note that MPNET's separator token is </s>, not [SEP].

\newpage

\textbf{Insincere Questions Step 1}

\textbf{Test Modified} What rapper still relevant and popular today has the best rhyme schemes? </s> <insincere question 3.859471321105957> What would be a good nickname for Trump, Donald Dumbck, and President Spankovich? </s> <valid question 4.124274253845215> What are after class 12 courses in commerce stream to choose from? I have completed my class 12 (expexted 90+) and aim to do business (not aim to do job).

\textbf{Label} valid question

\textbf{Test Modified} Which books do you suggest to someone who get a free time and will help him stay motivated? </s> <valid question 3.9509353637695312> What are the best online courses to learn data science? </s> <insincere question 4.300448417663574> What are the more steps in Career Oriented Education?

\textbf{Label} valid question

\textbf{Test Modified} How will you feel if someone talks badly about Kunti? </s> <insincere question 3.5063605308532715> Why are the UK government and the media (especially the BBC and the Guardian) demonising ordinary British people, manipulating buzz words like “alt-right”, “Islamaphobia”, “racist” to suppress legitimate outrage at Muslim grooming gangs? </s> <valid question 3.6699037551879883> How do Israelis and Palestinians view Nuseir Yassin?

\textbf{Label} valid question

\textbf{Test Modified} Why is equine HYPP inherited? </s> <valid question 4.066534996032715> Can you share some of the pics of hostel of Indira Gandhi medical college, Shimla, Himachal Pradesh? </s> <insincere question 4.231775760650635> I am an experienced programmer and in my high school my teacher tried to make me use python so I said, "No; Trust me, python is just a language for beginners, thereby making it not for me." I got sent out. Did I do anything wrong?

\textbf{Label} valid question

\textbf{Test Modified} How do the Valerie Stevens leather jackets achieve their quality during the manufacturing process? </s> <valid question 3.9721384048461914> How are the Lancaster leather sofas manufactured? </s> <insincere question 4.3559441566467285> I am an experienced programmer and in my high school my teacher tried to make me use python so I said, "No; Trust me, python is just a language for beginners, thereby making it not for me." I got sent out. Did I do anything wrong?

\textbf{Label} valid question

\textbf{Test Modified} Is Ariana Grande really as mean and bitchy as she seems? </s> <insincere question 3.572277545928955> Why is Alia Bhatt so dumb? </s> <valid question 3.924571990966797> Do you agree with Congressman Steve King's comments on immigrant children in detention centers?

\textbf{Label} insincere question

\textbf{Test Modified} Do you guys know that aliens are real and all those satellites we send up in space work as a sort of tracking device for them so in a few years it will be too late for Earth? </s> <insincere question 3.6094439029693604> Have you noticed how conservatives are capturing the English language and modifying the definitions of political words? </s> <valid question 3.6901655197143555> Do you agree with Congressman Steve King's comments on immigrant children in detention centers?

\textbf{Label} insincere question

\textbf{Test Modified} Is it politically incorrect to say female privilege, but it is a more accurate term to say, white female privilege? </s> <insincere question 3.323280096054077> Why are the UK government and the media (especially the BBC and the Guardian) demonising ordinary British people, manipulating buzz words like “alt-right”, “Islamaphobia”, “racist” to suppress legitimate outrage at Muslim grooming gangs? </s> <valid question 3.986680269241333> Do you agree with Congressman Steve King's comments on immigrant children in detention centers?

\textbf{Label} insincere question

\textbf{Test Modified} On Mother’s Day, is it reasonable to reflect there is some truth in the unfashionable notion than women are more driven by emotion and men more driven by reason? </s> <insincere question 3.499204158782959> Why are the UK government and the media (especially the BBC and the Guardian) demonising ordinary British people, manipulating buzz words like “alt-right”, “Islamaphobia”, “racist” to suppress legitimate outrage at Muslim grooming gangs? </s> <valid question 3.771740198135376> Do you agree with Congressman Steve King's comments on immigrant children in detention centers?

\textbf{Label} insincere question

\textbf{Test Modified} If the U.S. president is a role model, is it acceptable for children to say "go fuck yourself," and use the word "pussy" to describe women? </s> <insincere question 3.497847080230713> Why are the UK government and the media (especially the BBC and the Guardian) demonising ordinary British people, manipulating buzz words like “alt-right”, “Islamaphobia”, “racist” to suppress legitimate outrage at Muslim grooming gangs? </s> <valid question 3.845909357070923> Do you agree with Congressman Steve King's comments on immigrant children in detention centers?

\textbf{Label} insincere question

\textbf{Insincere Questions Step 5}

\textbf{Test Modified} What rapper still relevant and popular today has the best rhyme schemes? </s> <insincere question 3.871907949447632> What would be a good nickname for Trump, Donald Dumbck, and President Spankovich? </s> <valid question 4.028958797454834> Why does Dancing with the Stars not include Bachata as one their dance styles?

\textbf{Label} valid question

\textbf{Test Modified} Which books do you suggest to someone who get a free time and will help him stay motivated? </s> <valid question 3.6081225872039795> What is a good degree to get at community college if you want to explore different subjects and figure out your career path? </s> <insincere question 3.8502604961395264> What are the more steps in Career Oriented Education?

\textbf{Label} valid question

\textbf{Test Modified} How will you feel if someone talks badly about Kunti? </s> <valid question 3.5355563163757324> How do I stop feeling bad after a girl had a crush on me? </s> <insincere question 3.689171075820923> Why Indian girls go crazy about marrying Shri. Rahul Gandhi ji?

\textbf{Label} valid question

\textbf{Test Modified} Why is equine HYPP inherited? </s> <insincere question 3.6035702228546143> Can female animals with male humans sex? </s> <valid question 3.7413032054901123> How long do guinea pigs live for?

\textbf{Label} valid question

\textbf{Test Modified} How do the Valerie Stevens leather jackets achieve their quality during the manufacturing process? </s> <valid question 2.747288227081299> How are the Lancaster leather sofas manufactured? </s> <insincere question 3.944884777069092> Why don't all Trump supporters buy only made in USA goods, e.g. many of them have their cars of Asian/European companies, shop in places where more than 70 of items are not made in USA, eat multi-national cuisine or otherwise stop their hypocrisy?

\textbf{Label} valid question

\textbf{Test Modified} Is Ariana Grande really as mean and bitchy as she seems? </s> <insincere question 3.3252298831939697> Why is Alia Bhatt so dumb? </s> <valid question 3.7413415908813477> How do I stop feeling bad after a girl had a crush on me?

\textbf{Label} insincere question

\textbf{Test Modified} Do you guys know that aliens are real and all those satellites we send up in space work as a sort of tracking device for them so in a few years it will be too late for Earth? </s> <insincere question 3.0673365592956543> Isn't it obvious now that walking on the moon by the Americans was a hoax, because walking on the bright side of the moon, even in a space suit would be fatal? </s> <valid question 3.1978228092193604> Why do we weunch satellites?

\textbf{Label} insincere question

\textbf{Test Modified} Is it politically incorrect to say female privilege, but it is a more accurate term to say, white female privilege? </s> <insincere question 2.9176812171936035> How does the privilege of being attractive compare to the privilege of being White in the US? </s> <valid question 3.112481117248535> Is the media wrong for enforcing gender stereotypes?

\textbf{Label} insincere question

\textbf{Test Modified} On Mother’s Day, is it reasonable to reflect there is some truth in the unfashionable notion than women are more driven by emotion and men more driven by reason? </s> <insincere question 3.102353811264038> Do women look down on men who are single, even if the man is more successful in other aspects of his life? </s> <valid question 3.1890125274658203> Why are some women uninterested in sex?

\textbf{Label} insincere question

\textbf{Test Modified} If the U.S. president is a role model, is it acceptable for children to say "go fuck yourself," and use the word "pussy" to describe women? </s> <insincere question 3.163693904876709> Is it wrong to take your retarded son to a hooker for his 21st birthday? </s> <valid question 3.456286907196045> Do you agree with Congressman Steve King's comments on immigrant children in detention centers?

\textbf{Label} insincere question

\textbf{Insincere Questions Step 10}

\textbf{Test Modified} What rapper still relevant and popular today has the best rhyme schemes? </s> <valid question 3.7103171348571777> What is the oldest fashion trends running yet? </s> <insincere question 3.871907949447632> What would be a good nickname for Trump, Donald Dumbck, and President Spankovich?

\textbf{Label} valid question

\textbf{Test Modified} Which books do you suggest to someone who get a free time and will help him stay motivated? </s> <valid question 3.1401429176330566> How can I stay motivated when learning something new? </s> <insincere question 3.7235560417175293> I'm hungry and I'm too lazy too get out of bed, should I get a psychologist or ask you questions?

\textbf{Label} valid question

\textbf{Test Modified} How will you feel if someone talks badly about Kunti? </s> <insincere question 3.4893462657928467> Does Tamil Isai Soundarajan support Vijayendra for disrespecting the Tamil Anthem? </s> <valid question 3.5355563163757324> How do I stop feeling bad after a girl had a crush on me?

\textbf{Label} valid question

\textbf{Test Modified} Why is equine HYPP inherited? </s> <valid question 3.5067965984344482> What disadvantages do animals that don't have bones face? </s> <insincere question 3.6035702228546143> Can female animals with male humans sex?

\textbf{Label} valid question

\textbf{Test Modified} How do the Valerie Stevens leather jackets achieve their quality during the manufacturing process? </s> <valid question 2.747288227081299> How are the Lancaster leather sofas manufactured? </s> <insincere question 3.9087233543395996> Are Newport cigarettes designed to selectively destroy black people's DNA?

\textbf{Label} valid question

\textbf{Test Modified} Is Ariana Grande really as mean and bitchy as she seems? </s> <valid question 3.183567762374878> I like this girl who used to be quite rude and would run through boyfriends very fast. But now that school started again, she seems to have gotten a lot nicer throughout Summer. Is she faking her politeness, and is it worth pursuing her? </s> <insincere question 3.3253660202026367> Why is Alia Bhatt so dumb?

\textbf{Label} insincere question

\textbf{Test Modified} Do you guys know that aliens are real and all those satellites we send up in space work as a sort of tracking device for them so in a few years it will be too late for Earth? </s> <insincere question 3.0673365592956543> Isn't it obvious now that walking on the moon by the Americans was a hoax, because walking on the bright side of the moon, even in a space suit would be fatal? </s> <valid question 3.1978228092193604> Why do we weunch satellites?

\textbf{Label} insincere question

\textbf{Test Modified} Is it politically incorrect to say female privilege, but it is a more accurate term to say, white female privilege? </s> <insincere question 2.9176158905029297> How does the privilege of being attractive compare to the privilege of being White in the US? </s> <valid question 3.112481117248535> Is the media wrong for enforcing gender stereotypes?

\textbf{Label} insincere question

\textbf{Test Modified} On Mother’s Day, is it reasonable to reflect there is some truth in the unfashionable notion than women are more driven by emotion and men more driven by reason? </s> <insincere question 2.9901626110076904> Do you agree that females think with their brains and males with their testicles? </s> <valid question 3.1890125274658203> Why are some women uninterested in sex?

\textbf{Label} insincere question

\textbf{Test Modified} If the U.S. president is a role model, is it acceptable for children to say "go fuck yourself," and use the word "pussy" to describe women? </s> <insincere question 2.994286298751831> Why do feminists let their daughters have sex with their boyfriend’s at home? </s> <valid question 3.456286907196045> Do you agree with Congressman Steve King's comments on immigrant children in detention centers?

\textbf{Label} insincere question

\textbf{Amazon Counterfactual Step 1}

\textbf{Test Modified} Clings to the wall, doesn't flop around when a bag is pulled out, the mess of bags falling out is gone. </s> <not-counterfactual 3.6492726802825928> Hopes that it will keep it's shape after washing. </s> <counterfactual 4.012346267700195> "Had I reviewed this immediately I would have given this product five stars because It worked."""

\textbf{Label} not-counterfactual

\textbf{Test Modified} I like these jeans they sit low enough without being inappropriate when you sit or bend over. </s> <counterfactual 3.402600049972534> "But oddly enough, the bottoms are a little too loose in the waist (37) and could have used another inch or two in the inseam ( I normally take a 35"" or 36"" in jeans, depending on the brand if this helps).""" </s> <not-counterfactual 3.4201438426971436> These boxer-briefs are very soft, very comfortable, and fit like high-end underwear the likes of which you might get at, oh, say, Calvin Klein for example, but for about half the price.

\textbf{Label} not-counterfactual

\textbf{Test Modified} He was very professional and wish all transactions I make through Amazon were this good. </s> <counterfactual 3.4319908618927> I wish I had had him as an instructor at college. </s> <not-counterfactual 4.054030895233154> I worried that it would be cheap or not fit or...whatever...But WOW!

\textbf{Label} not-counterfactual

\textbf{Test Modified} Well written with a twist I didn't expect. </s> <not-counterfactual 3.3257973194122314> "The crossover from the characters from one novel to others keeps me interested; after all, I do hate to miss a Dee-Ann or Eggie"" appearance.""" </s> <counterfactual 3.6820030212402344> "Had I reviewed this immediately I would have given this product five stars because It worked."""

\textbf{Label} not-counterfactual

\textbf{Test Modified} Doesn't feel like the quality levi's I am used to. </s> <not-counterfactual 3.2773308753967285> However, the fabric is not that great, it's cheap scratchy cotton. </s> <counterfactual 3.746659755706787> The blanket is nice and soft but it is white, so if it doesn't light up it isn't much use!

\textbf{Label} not-counterfactual

\textbf{Test Modified} If we had wall studs, I believe the enclosed hardware would have been sufficient. </s> <counterfactual 3.4338643550872803> i wish the storage compartment was a little bigger and opened up instead of slidding on and off. </s> <not-counterfactual 3.9785308837890625> I worried that it would be cheap or not fit or...whatever...But WOW!

\textbf{Label} counterfactual

\textbf{Test Modified} If this ever turns into a film, I hope they do it justice! </s> <not-counterfactual 3.5291523933410645> "The crossover from the characters from one novel to others keeps me interested; after all, I do hate to miss a Dee-Ann or Eggie"" appearance.""" </s> <counterfactual 3.751143217086792> "Had I reviewed this immediately I would have given this product five stars because It worked."""

\textbf{Label} counterfactual

\textbf{Test Modified} If you don't want a prominent display this rack is too large for most bed or living rooms, it is wider and taller than my tall Broyhill wardrobe style dresser which was the largest piece in the room until this shoe rack. </s> <not-counterfactual 3.865670680999756> "It also validates the incorrect"" assumption that we are alone in the feelings we suppress when we sense the complete garbage that is thrown out into society.""" </s> <counterfactual 4.063361167907715> The blanket is nice and soft but it is white, so if it doesn't light up it isn't much use!

\textbf{Label} counterfactual

\textbf{Test Modified} I wish I could have seen all of the places he recommends! </s> <counterfactual 3.5627076625823975> I wish I had had him as an instructor at college. </s> <not-counterfactual 4.141315937042236> I worried that it would be cheap or not fit or...whatever...But WOW!

\textbf{Label} counterfactual

\textbf{Test Modified} I wish I could replace just that small stupid piece, since there's nothing wrong with the rest of the hose assembly. </s> <counterfactual 3.6057372093200684> i wish the storage compartment was a little bigger and opened up instead of slidding on and off. </s> <not-counterfactual 4.064871311187744> I worried that it would be cheap or not fit or...whatever...But WOW!

\textbf{Label} counterfactual

\textbf{Amazon Counterfactual Step 5}

\textbf{Test Modified} Clings to the wall, doesn't flop around when a bag is pulled out, the mess of bags falling out is gone. </s> <not-counterfactual 3.161406993865967> And the dvd cases were tightly packed to ensure they didn't move around. </s> <counterfactual 3.308583974838257> The case is small, cord seems to always want to stay kinked and coiled, plug should be angled and not straight...which are all items that others have pointed out.

\textbf{Label} not-counterfactual

\textbf{Test Modified} I like these jeans they sit low enough without being inappropriate when you sit or bend over. </s> <counterfactual 2.606198310852051> "But oddly enough, the bottoms are a little too loose in the waist (37) and could have used another inch or two in the inseam ( I normally take a 35"" or 36"" in jeans, depending on the brand if this helps).""" </s> <not-counterfactual 2.6380045413970947> A tad loose but I rather have it fit this way than too tight.

\textbf{Label} not-counterfactual

\textbf{Test Modified} He was very professional and wish all transactions I make through Amazon were this good. </s> <not-counterfactual 3.3291680812835693> This new speaker was just what the doctor ordered and I couldn't be more pleased. </s> <counterfactual 3.4589436054229736> Had the person handling the shipping of this item been at all concerned with the use of the product at the end of the mailing process, the slightest bit of care could have been taken to ensure it's proper delivery.

\textbf{Label} not-counterfactual

\textbf{Test Modified} Well written with a twist I didn't expect. </s> <not-counterfactual 2.651658535003662> The book had some interesting twists that I did see coming and I look forward to reading part two of this series. </s> <counterfactual 2.8373162746429443> Fun read Could have been a little longer with more detail.

\textbf{Label} not-counterfactual

\textbf{Test Modified} Doesn't feel like the quality levi's I am used to. </s> <counterfactual 2.733877182006836> It has the same great comfortable \& flattering features plus the great denim texture that Lee has perfected- smoothing and stretchy without the excessive cling- but I think it must have been designed for people who have a greater surplus of belly fat than I. </s> <not-counterfactual 2.856729745864868> Will keep but won't be that casual sexy top you always want to turn to.

\textbf{Label} not-counterfactual

\textbf{Test Modified} If we had wall studs, I believe the enclosed hardware would have been sufficient. </s> <not-counterfactual 2.6638145446777344> It was a little tricky to find the center of the studs using my stud finder but once I felt comfortable with the lines I had drawn, I drilled the pilot holes and bolted this thing to the wall. </s> <counterfactual 2.879924774169922> The only thing I would have like for it to have a hole in the middle so I can put the stopper in without removing the mat.

\textbf{Label} counterfactual

\textbf{Test Modified} If this ever turns into a film, I hope they do it justice! </s> <not-counterfactual 2.671574354171753> I read this book because of the motion picture that is coming out soon. </s> <counterfactual 3.1458709239959717> Was a good story, though there could have been more to it.

\textbf{Label} counterfactual

\textbf{Test Modified} If you don't want a prominent display this rack is too large for most bed or living rooms, it is wider and taller than my tall Broyhill wardrobe style dresser which was the largest piece in the room until this shoe rack. </s> <counterfactual 2.7353768348693848> I bought this mount because I wanted one that would sit on three studs instead of two because my TV is quite heavy and I would have had a hard time centering it on my wall if I didn't have the wide hanging rail that this one has. </s> <not-counterfactual 2.873617172241211> Good for under the bed shoe storage, IF the wife wants to use it.

\textbf{Label} counterfactual

\textbf{Test Modified} I wish I could have seen all of the places he recommends! </s> <counterfactual 2.799947738647461> I wish I had had him as an instructor at college. </s> <not-counterfactual 3.3013432025909424> And as the ole man isn't any version of slender it was good that he got to try on some shirts before hand.

\textbf{Label} counterfactual

\textbf{Test Modified} I wish I could replace just that small stupid piece, since there's nothing wrong with the rest of the hose assembly. </s> <counterfactual 2.628289222717285> The only thing I would have like for it to have a hole in the middle so I can put the stopper in without removing the mat. </s> <not-counterfactual 2.9200568199157715> The only downside is my laptop does not have the screw holes on it and the screws do not retract far enough back for me to push the connector all the way in, but a simple smash will rid that issue (this thing is durable!)

\textbf{Label} counterfactual

\textbf{Amazon Counterfactual Step 10}

\textbf{Test Modified} Clings to the wall, doesn't flop around when a bag is pulled out, the mess of bags falling out is gone. </s> <not-counterfactual 3.161406993865967> And the dvd cases were tightly packed to ensure they didn't move around. </s> <counterfactual 3.289605140686035> If I had to come up with anything negative, I would say that the attachments don’t seem to stay on the vacuum cleaner when not in use - but that could be me not putting them on properly!

\textbf{Label} not-counterfactual

\textbf{Test Modified} I like these jeans they sit low enough without being inappropriate when you sit or bend over. </s> <not-counterfactual 2.447404623031616> These shorts fit really well and look good too. </s> <counterfactual 2.550638198852539> The top fits great just wish the bottoms fit too.

\textbf{Label} not-counterfactual

\textbf{Test Modified} He was very professional and wish all transactions I make through Amazon were this good. </s> <not-counterfactual 3.3291127681732178> This new speaker was just what the doctor ordered and I couldn't be more pleased. </s> <counterfactual 3.3897111415863037> But the author alleviated my concerns quickly with a few well-timed comments about how it was the man could have known that the arrangement was something Jack wanted.

\textbf{Label} not-counterfactual

\textbf{Test Modified} Well written with a twist I didn't expect. </s> <not-counterfactual 2.557446002960205> "A bit workmanlike, not up to Lord's high standard of A Night to Remember,"" but well-detailed, and a story that not many now know.""" </s> <counterfactual 2.792485475540161> Wow I am really glad I didn't read these reviews BEFORE I read this book because I would have passed on the book and missed a really great start to a series that captured my attention and made me laugh all the while using my imagination and painting a clear picture of the author's world she was building for us.

\textbf{Label} not-counterfactual

\textbf{Test Modified} Doesn't feel like the quality levi's I am used to. </s> <counterfactual 2.5612902641296387> i was hoping the pants would be thicker but being that it's not too expensive it's understandable. </s> <not-counterfactual 2.572395086288452> But it doesn't have a lining like the last couple models I bought.

\textbf{Label} not-counterfactual

\textbf{Test Modified} If we had wall studs, I believe the enclosed hardware would have been sufficient. </s> <not-counterfactual 2.6638145446777344> It was a little tricky to find the center of the studs using my stud finder but once I felt comfortable with the lines I had drawn, I drilled the pilot holes and bolted this thing to the wall. </s> <counterfactual 2.771395206451416> Wish it had a little more padding, otherwise just as advertised.

\textbf{Label} counterfactual

\textbf{Test Modified} If this ever turns into a film, I hope they do it justice! </s> <not-counterfactual 2.671574354171753> I read this book because of the motion picture that is coming out soon. </s> <counterfactual 3.141676187515259> Wish this story would have been longer and turned into a book, with some gut wrenching action, love/hate lovers quarrels scenes, with a happy ending at the end...

\textbf{Label} counterfactual

\textbf{Test Modified} If you don't want a prominent display this rack is too large for most bed or living rooms, it is wider and taller than my tall Broyhill wardrobe style dresser which was the largest piece in the room until this shoe rack. </s> <counterfactual 2.7353768348693848> I bought this mount because I wanted one that would sit on three studs instead of two because my TV is quite heavy and I would have had a hard time centering it on my wall if I didn't have the wide hanging rail that this one has. </s> <not-counterfactual 2.873617172241211> Good for under the bed shoe storage, IF the wife wants to use it.

\textbf{Label} counterfactual

\textbf{Test Modified} I wish I could have seen all of the places he recommends! </s> <counterfactual 2.7999041080474854> I wish I had had him as an instructor at college. </s> <not-counterfactual 3.2604622840881348> I wanted to order him a few afordable hats I wouldn't mind him loosing.

\textbf{Label} counterfactual

\textbf{Test Modified} I wish I could replace just that small stupid piece, since there's nothing wrong with the rest of the hose assembly. </s> <counterfactual 2.474032402038574> I wish I could just hook up a hose rather than connecting and routing an tube internally </s> <not-counterfactual 2.9200568199157715> The only downside is my laptop does not have the screw holes on it and the screws do not retract far enough back for me to push the connector all the way in, but a simple smash will rid that issue (this thing is durable!)

\textbf{Label} counterfactual

\textbf{Toxic Conversations Step 1}

\textbf{Test Modified} Homos demand that you accept their fudge packing. But none of us ever will- </s> <not toxic 3.5619421005249023> Sounds just awful, working for the state that is. So it begs the question, why work for the state if the pay is so bad versus the private sector? Seems logical to just make the switch? </s> <toxic 4.056504249572754> I would imagine it is a miserable existence. Being a whining sore loser is no way to go through life.

\textbf{Label} not toxic

\textbf{Test Modified} I don't think anyone likes this health care bill, it stinks for everyone. 50 years and older are going to get hammered with higher premiums. People with preexisting conditions will also see their premiums go through the roof. Eventually no one will be able to afford it. They're still not addressing the reason and the solution. Prescription drugs. Why are Americans paying 2 to 3 times higher for prescription drugs than Canadians are? They should sit down with Democrats and fix it not play musical chairs with our health care. Here is the add that AARP put out, and it has a fact check at the end. It's worth watching.
http://www.thedenverchannel.com/news/politics

/aarp-ad-says-house-gop-health-care-bill-would-boost-insurance-rates-for-older-americans </s> <toxic 3.6495330333709717> Calling everybody that disagrees with you a racist and anti-Muslim is pretty uncivil.
Also intellectually bankrupt... </s> <not toxic 3.8587634563446045> Sounds just awful, working for the state that is. So it begs the question, why work for the state if the pay is so bad versus the private sector? Seems logical to just make the switch?

\textbf{Label} not toxic

\textbf{Test Modified} Rather than call you a liar, I'll just post a link to HART's helicopter footage of this "most densely populated" ag land. 

https://www.youtube.com/watch?
v=opJzRWWVDdw

And if you've been following a little closer, you'd know about the luxury towers and boutique hotels that developers are planning on fulfilling their TOD requirement with "cash gifts" to the city instead of actual affordable TOD. </s> <not toxic 3.8009092807769775> Sounds just awful, working for the state that is. So it begs the question, why work for the state if the pay is so bad versus the private sector? Seems logical to just make the switch? </s> <toxic 4.463983535766602> I would imagine it is a miserable existence. Being a whining sore loser is no way to go through life.

\textbf{Label} not toxic

\textbf{Test Modified} Get the "real news" from Breitbart and Drudge Report? Ahahahahaha! Are you for real? </s> <not toxic 3.559427499771118> Sounds just awful, working for the state that is. So it begs the question, why work for the state if the pay is so bad versus the private sector? Seems logical to just make the switch? </s> <toxic 4.239322662353516> I would imagine it is a miserable existence. Being a whining sore loser is no way to go through life.

\textbf{Label} not toxic

\textbf{Test Modified} Wud luv to see one of those American Boats with all the planes on it at the bottom of the sea. </s> <not toxic 3.8268911838531494> Sounds just awful, working for the state that is. So it begs the question, why work for the state if the pay is so bad versus the private sector? Seems logical to just make the switch? </s> <toxic 4.241443157196045> I would imagine it is a miserable existence. Being a whining sore loser is no way to go through life.

\textbf{Label} not toxic

\textbf{Test Modified} Wtf, nowhere in scripture does it say that Elizabeth was a kid when married to Zachariah and nowhere does it say that Mary was a child when married to Joseph. That's a complete and utter fabrication. In fact, it says in Luke 1:42, "And she spake out with a loud voice, and said, Blessed art thou among women, and blessed is the fruit of thy womb." It says blessed art thou among WOMEN. </s> <not toxic 3.6674933433532715> Sounds just awful, working for the state that is. So it begs the question, why work for the state if the pay is so bad versus the private sector? Seems logical to just make the switch? </s> <toxic 4.0054121017456055> I would imagine it is a miserable existence. Being a whining sore loser is no way to go through life.

\textbf{Label} toxic

\textbf{Test Modified} Angela Merkel and all other European political leaders who have aided and abetted the ongoing invasion of Europe by the forces of the crescent moon death cult should be tried as accessories to Mr. Urban's murder. </s> <toxic 3.262410879135132> Calling everybody that disagrees with you a racist and anti-Muslim is pretty uncivil.
Also intellectually bankrupt... </s> <not toxic 3.8916428089141846> It's always important to remember what can happen when you have sociopaths as leaders and also have compliant followers. Some of the younger posters on this site might want to Google "Jim Jones and Jonestow"... There were no "checks and balances" in Jonestown; I fear there are none in North Korea....and I can only hope those in our country are firmly in place and functioning. Gary Crum

\textbf{Label} toxic

\textbf{Test Modified} I hope you don't have kids if you see this woman's actions as acceptable. And I applaud the den for kicking the kid out. She brought unwanted negative attention upon them. However, she will, and is already likely, pay the the price for her stupid stunt. </s> <toxic 3.0406124591827393> Calling everybody that disagrees with you a racist and anti-Muslim is pretty uncivil.
Also intellectually bankrupt... </s> <not toxic 4.094666481018066> Christ never said he would give grace, mercy, and acceptance to those who determinedly violate Scripture. In fact, he often spoke of hell.

\textbf{Label} toxic

\textbf{Test Modified} no one cares what a paid liberal trolling hack like you believes lunatic,, </s> <toxic 2.8411786556243896> Calling everybody that disagrees with you a racist and anti-Muslim is pretty uncivil.
Also intellectually bankrupt... </s> <not toxic 4.034884929656982> Christ never said he would give grace, mercy, and acceptance to those who determinedly violate Scripture. In fact, he often spoke of hell.

\textbf{Label} toxic

\textbf{Test Modified} Ok all you NDP "LEAP" manifesto types, where is your hero Naomi Klein? Her fawning adoration of Chavez and Venezuelan thuggery knows no bounds. I'm sure she's awfully hysterical over the thought that such a pathetic dictstorship could ever be sanctioned. </s> <toxic 3.3616135120391846> Calling everybody that disagrees with you a racist and anti-Muslim is pretty uncivil.
Also intellectually bankrupt... </s> <not toxic 3.903903007507324> I have very high respect for teachers that get the job done. Teaching is an extremely difficult and important job. And it is quite apparent that we are in desperate need of teachers that can actually do the job. Pride of workmanship would have teachers wanting to have their students periodically evaluated and tested to show how well they have done their job. We have some very competent teachers that get the job done and welcome student testing (in spite of sorry admin and unfair union). But the majority of teachers here instead of doing their jobs they band together wear purple shirts and mob the government for a better contract, and no accountability in the form of testing students. Many teachers are middle class kids that took the path of least resistance in what was expected by their parents (college) and because they lacked drive ended up teachers. That lack of drive shows by what the private sector taxpayers get for their money. Your degrees mean nothing if you don't do your job.

\textbf{Label} toxic

\textbf{Toxic Conversations Step 5}

\textbf{Test Modified} Homos demand that you accept their fudge packing. But none of us ever will- </s> <toxic 3.1383495330810547> So you admit you would exterminate inferior humans. </s> <not toxic 3.2954952716827393> Mark MacKinnon and the interests he work for would like us to 'get used to it', because they don't want to do anything practical to stop it.

\textbf{Label} not toxic

\textbf{Test Modified} I don't think anyone likes this health care bill, it stinks for everyone. 50 years and older are going to get hammered with higher premiums. People with preexisting conditions will also see their premiums go through the roof. Eventually no one will be able to afford it. They're still not addressing the reason and the solution. Prescription drugs. Why are Americans paying 2 to 3 times higher for prescription drugs than Canadians are? They should sit down with Democrats and fix it not play musical chairs with our health care. Here is the add that AARP put out, and it has a fact check at the end. It's worth watching.
http://www.thedenverchannel.com/news/politics/

aarp-ad-says-house-gop-health-care-bill-would-boost-insurance-rates-for-older-americans </s> <not toxic 2.5086519718170166> so in the mean time tens of thousands of Oregonians go without health insurance which will now be unaffordable to them. And sorry, the republicans have had 8 years to figure out a better system, they aren't going to do it anytime soon. Neither party has any desire to actually find a solution to all this.
Hopefully Trump will also soon eliminate the tax penalty for not having insurance so all us folks who buy our own insurance don't get penalized for not being able to afford to buy the insurance we are required to have. Of course he is probably clueless that detail even exists </s> <toxic 2.801957607269287> reducing number of brackets. Another is lowering corporate tax rates which would be OK if all loopholes, including tax-free political donations for wealthy people only. Another is doubling the earned income tax credit which would help families with children but for people like me, would not make up for loss of the state tax deduction. Essentially the proposed tax "reform bill as it stands is a huge wealth transfer, from working people and the treasury to the offshore accounts of wealthy people and corporations. Median earners like me will be screwed. Poor people with children will be screwed. The rich will get a huge bundle of cash, and the Republicans will lead us further into deep debt.

That's if Ryan and McConnell can pass this huge sack of excrement.

\textbf{Label} not toxic

\textbf{Test Modified} Rather than call you a liar, I'll just post a link to HART's helicopter footage of this "most densely populated" ag land. 

https://www.youtube.com/watch?v=
opJzRWWVDdw

And if you've been following a little closer, you'd know about the luxury towers and boutique hotels that developers are planning on fulfilling their TOD requirement with "cash gifts" to the city instead of actual affordable TOD. </s> <not toxic 3.0351996421813965> So are you saying we should build the road. Lots of doom and gloom but I'm missing your point. </s> <toxic 3.1464040279388428> Isn't that terrible? Caldwell and Inouye do not know the amount a 10-year extension would raise yet they are proposing it. Caldwell asked for two years and he got it. Two years later he is now asking for ten years. Now, which is only weeks later, HART director is claiming 10 years is insufficient. This scenario shows our mayor and HART do not know the scope of this project. Pathetic!

\textbf{Label} not toxic

\textbf{Test Modified} Get the "real news" from Breitbart and Drudge Report? Ahahahahaha! Are you for real? </s> <toxic 2.634126901626587> "If one read the Dispatch one would think Trump is the most evil person on the planet."
Not evil, just idiotic. And it would be easy to give his behaviour a pass if he wasn't POTUS.
".....five to eight anti-Trump stories per day. Never any good one's or one's that just stuck to the facts." Well, when there's good a news Trump story to print, I'm sure ADN will be all over it, problem is, there's been a dearth of those since his election. Facts? Ok Rich, give us a list of incorrect facts in the above story. </s> <not toxic 3.0248677730560303> With dismay I noticed that ADN had printed yet another column from Fox commentator Charles Krauthammer but after reading it I'm glad the editors chose it as the feature article on the opinion page. Krauthammer is also a psychiatrist so his analyses of Trump as a man who has never emotionally, intellectually developed beyond adolescence holds some weight. But what does it say about Trumps supporters that so many millions of them can't see through the boorish, confrontational attitude of the man? How can so many Americans have devolved into such anger, fear and irrationality that they would/could find redemption in Trump after how he has exposed his true narcissistic self for all to see. When you've lost the Jennifer Rubins and Charles Krauthammer's of the media world you've lost the battle yet the Trumpian cult members will soldier on and then become even angrier and more full of fear after the election. Something to do with their choice of "information" sources no doubt.

\textbf{Label} not toxic

\textbf{Test Modified} Wud luv to see one of those American Boats with all the planes on it at the bottom of the sea. </s> <toxic 3.3901054859161377> I bet Regent Seven Seas will never offer Mr Hammond another trip.
Wow, what a snarky article. He makes , I assume, some valid
points about food and atmosphere. However, after discovering the
treats available on his "massive deck" he "blew off" his remaining
restaurant reservations , donned his comfy bathrobe and ordered-in.
He was certainly not an ideal passenger and, for one floating on a freebie,
he's a total ingrate! </s> <not toxic 3.409156084060669> Now replaced by the sexy EA-18G Growler! Using a preexisting Military Operating Area! Get over it!!!!!!

\textbf{Label} not toxic

\textbf{Test Modified} Wtf, nowhere in scripture does it say that Elizabeth was a kid when married to Zachariah and nowhere does it say that Mary was a child when married to Joseph. That's a complete and utter fabrication. In fact, it says in Luke 1:42, "And she spake out with a loud voice, and said, Blessed art thou among women, and blessed is the fruit of thy womb." It says blessed art thou among WOMEN. </s> <not toxic 2.769857406616211> I was informed that my first grandchild had been conceived the evening of the day when I had inserted a prayer note in the Wailing Wall in Jerusalem that asked God to help my daughter conceive after a year of frustrated attempts. Maybe Elizabeth did the same thing? After all, she was in the same neighborhood. :-) </s> <toxic 3.286393880844116> Christians who support Trump are the most mind-boggling to me. I just don't see how they square the circle between Trump and their moral foundations. 

“Beware then of useless grumbling, and keep your tongue from slander; because no secret word is without result, and a lying mouth destroys the soul.” (Wisdom 1:11)

If that is the case, then Trump's soul was utterly destroyed decades ago.

\textbf{Label} toxic

\textbf{Test Modified} Angela Merkel and all other European political leaders who have aided and abetted the ongoing invasion of Europe by the forces of the crescent moon death cult should be tried as accessories to Mr. Urban's murder. </s> <toxic 3.2037758827209473> that's what happens when you betray the people of your country for foreign bs. let's go Le Pen, Geert Wilders. If the media refuses to mention the muslim crisis the total incompatibility of primitive, uneducated muslim males swarming countries and turning them into misogynistic fundamentalist religious areas then we need these people to save us from YOU! </s> <not toxic 3.2326502799987793> your first mistake is believing what a politician says because generally it has nothing to do with what they do.

The Libs will be happy to let this die because Monsef is now a very poor salesman given her own immigration dishonesty. That said if the election prospects sour significantly for the Libs I have no doubts that PM Butts will ram through Ranked Ballot

\textbf{Label} toxic

\textbf{Test Modified} I hope you don't have kids if you see this woman's actions as acceptable. And I applaud the den for kicking the kid out. She brought unwanted negative attention upon them. However, she will, and is already likely, pay the the price for her stupid stunt. </s> <toxic 2.8730525970458984> Ms. Van Brocklin: You state that there is Payne's conduct is inexcusable, then proceed to use the rest of your space to justify him and his actions. You have denigrated the proud courage of countless people who took up causes via civil disobedience. I marched in the non-violent peaceful Civil Rights protest. So the dogs and the firehoses used by a certain southern sheriff were justified, by your logic. So were the citizens beaten by Chicago police during the Democratic Convention Police riots. Resolved in 20 minutes? Nonsense. If Miss Wubbels hadn't protested as she did, she'd likely have ended up in a back room, somewhere, and who know when she would have been granted the presence of a lawyer and what she would have gone through prior to that. 
 You are completely ignorant of the shortage of nurses in this country - in some cases, critical shortages. And why would anyone want to be a nurse when they are disrespected by a former state and federal prosecutor such as you. </s> <not toxic 2.92986798286438> Acquit her, then commit her. This womens cheese has slid so far off the cracker she's a danger to herself and to others.

Animal rights activism is a just cause, but her and her group have gone off the deep end into radical extremist territory.

\textbf{Label} toxic

\textbf{Test Modified} no one cares what a paid liberal trolling hack like you believes lunatic,, </s> <toxic 2.6076736450195312> aa another hate filled left winger again! save the stupid nonsense sheep, trump is not causing anything, our weak leadership is. </s> <not toxic 2.810284376144409> Ouch... didn’t see that one coming. A liberal stealing my own line... just like they take everything else they like.

\textbf{Label} toxic

\textbf{Test Modified} Ok all you NDP "LEAP" manifesto types, where is your hero Naomi Klein? Her fawning adoration of Chavez and Venezuelan thuggery knows no bounds. I'm sure she's awfully hysterical over the thought that such a pathetic dictstorship could ever be sanctioned. </s> <toxic 2.941824436187744> Ms. Van Brocklin: You state that there is Payne's conduct is inexcusable, then proceed to use the rest of your space to justify him and his actions. You have denigrated the proud courage of countless people who took up causes via civil disobedience. I marched in the non-violent peaceful Civil Rights protest. So the dogs and the firehoses used by a certain southern sheriff were justified, by your logic. So were the citizens beaten by Chicago police during the Democratic Convention Police riots. Resolved in 20 minutes? Nonsense. If Miss Wubbels hadn't protested as she did, she'd likely have ended up in a back room, somewhere, and who know when she would have been granted the presence of a lawyer and what she would have gone through prior to that. 
 You are completely ignorant of the shortage of nurses in this country - in some cases, critical shortages. And why would anyone want to be a nurse when they are disrespected by a former state and federal prosecutor such as you. </s> <not toxic 3.0424022674560547> So anyone who doesn't share your opinion is a "raving loony?"

And as for my post being "speculation? - which part - that the Liberals are the party in power, or that this involves money?

As for me not knowing what is going on, you are correct, I am not a member of the Liberal party insider clique, as you apparently are.

\textbf{Label} toxic

\textbf{Toxic Conversations Step 10}

\textbf{Test Modified} Homos demand that you accept their fudge packing. But none of us ever will- </s> <toxic 3.1383132934570312> So you admit you would exterminate inferior humans. </s> <not toxic 3.295428514480591> Mark MacKinnon and the interests he work for would like us to 'get used to it', because they don't want to do anything practical to stop it.

\textbf{Label} not toxic

\textbf{Test Modified} I don't think anyone likes this health care bill, it stinks for everyone. 50 years and older are going to get hammered with higher premiums. People with preexisting conditions will also see their premiums go through the roof. Eventually no one will be able to afford it. They're still not addressing the reason and the solution. Prescription drugs. Why are Americans paying 2 to 3 times higher for prescription drugs than Canadians are? They should sit down with Democrats and fix it not play musical chairs with our health care. Here is the add that AARP put out, and it has a fact check at the end. It's worth watching.
http://www.thedenverchannel.com/news/politics/

aarp-ad-says-house-gop-health-care-bill-would-boost-insurance-rates-for-older-americans </s> <not toxic 2.5086519718170166> so in the mean time tens of thousands of Oregonians go without health insurance which will now be unaffordable to them. And sorry, the republicans have had 8 years to figure out a better system, they aren't going to do it anytime soon. Neither party has any desire to actually find a solution to all this.
Hopefully Trump will also soon eliminate the tax penalty for not having insurance so all us folks who buy our own insurance don't get penalized for not being able to afford to buy the insurance we are required to have. Of course he is probably clueless that detail even exists </s> <toxic 2.5147175788879395> hate to bust the bubble but over 60 of people including all those trump voters never liked or wanted obamacare and dont want it now. , trump has NOT told the gop to back off you lying sack of bs.. he wants it gone and replaced period and the gop are doing just that together WITH trump. , you will continue to turn reality into stupidity

\textbf{Label} not toxic

\textbf{Test Modified} Rather than call you a liar, I'll just post a link to HART's helicopter footage of this "most densely populated" ag land. 

https://www.youtube.com/watch?v=
opJzRWWVDdw

And if you've been following a little closer, you'd know about the luxury towers and boutique hotels that developers are planning on fulfilling their TOD requirement with "cash gifts" to the city instead of actual affordable TOD. </s> <not toxic 2.9202661514282227> I suppose you just support urban sprawl then with that logic. </s> <toxic 2.9730660915374756> Why don't you go and live in one of their buildings and see what they're like? "Deadbeats" - you're an idiot. They're my neighbours.

\textbf{Label} not toxic

\textbf{Test Modified} Get the "real news" from Breitbart and Drudge Report? Ahahahahaha! Are you for real? </s> <toxic 2.634126901626587> "If one read the Dispatch one would think Trump is the most evil person on the planet."
Not evil, just idiotic. And it would be easy to give his behaviour a pass if he wasn't POTUS.
".....five to eight anti-Trump stories per day. Never any good one's or one's that just stuck to the facts." Well, when there's good a news Trump story to print, I'm sure ADN will be all over it, problem is, there's been a dearth of those since his election. Facts? Ok Rich, give us a list of incorrect facts in the above story. </s> <not toxic 2.9902079105377197> "a gift from the political gods when the struggling effort to pass a health bill dominates the headlines."

It was a gift from media that isn't worried about the actual news, they are more worried about trying to influence soft heads. CNN has been screwing up a lot when it comes to Trump, same with the NYT that is now being sued for libel. 

Every stupid mistake they make gives an even larger advantage to Trump and strengthens his supporters that already believe the MSM is biased against him and makes fence sitters begin to question what's news and what's crap. Like I've been saying, the MSM is slitting it's own throat.

\textbf{Label} not toxic

\textbf{Test Modified} Wud luv to see one of those American Boats with all the planes on it at the bottom of the sea. </s> <toxic 3.390166759490967> I bet Regent Seven Seas will never offer Mr Hammond another trip.
Wow, what a snarky article. He makes , I assume, some valid
points about food and atmosphere. However, after discovering the
treats available on his "massive deck" he "blew off" his remaining
restaurant reservations , donned his comfy bathrobe and ordered-in.
He was certainly not an ideal passenger and, for one floating on a freebie,
he's a total ingrate! </s> <not toxic 3.409156084060669> Now replaced by the sexy EA-18G Growler! Using a preexisting Military Operating Area! Get over it!!!!!!

\textbf{Label} not toxic

\textbf{Test Modified} Wtf, nowhere in scripture does it say that Elizabeth was a kid when married to Zachariah and nowhere does it say that Mary was a child when married to Joseph. That's a complete and utter fabrication. In fact, it says in Luke 1:42, "And she spake out with a loud voice, and said, Blessed art thou among women, and blessed is the fruit of thy womb." It says blessed art thou among WOMEN. </s> <not toxic 2.769857406616211> I was informed that my first grandchild had been conceived the evening of the day when I had inserted a prayer note in the Wailing Wall in Jerusalem that asked God to help my daughter conceive after a year of frustrated attempts. Maybe Elizabeth did the same thing? After all, she was in the same neighborhood. :-) </s> <toxic 2.9830095767974854> I don't know if the bishops ever told priests that if they rape a child and celebrate Mass they are committing sacrilege. But even if they didn't, the priest should know that...it is pretty basic Catholicism that receiving Communion with a mortal sin on your soul is sacrilege. 

By the logic of your second paragraph, there can never be a mortally sinful sexual act, since sex acts by definition occur in a state of passion. Which begs the question, why in this case would the Scriptures go through the trouble of condemning sexual immorality? This sounds like something your example of a rapist priest could say to himself to justify himself taking Communion in a state of mortal sin.

\textbf{Label} toxic

\textbf{Test Modified} Angela Merkel and all other European political leaders who have aided and abetted the ongoing invasion of Europe by the forces of the crescent moon death cult should be tried as accessories to Mr. Urban's murder. </s> <toxic 3.024935007095337> About 415 million Europeans cannot compete with Google, Amazon, Facebook, Oracle, Intel, Apple, etc. and the socialist European welfare states need more revenue because they are running out of other peoples' money.

Thus the Euro-socialist-bureaucrats pick the low-hanging fruit with litigious persecution of American firms which dominate because unlike their pathetic Euro-competitors, the U.S. firms are clever, hard-working, and well-capitalized.

If the the Europeans wish to engage in this transparent financial inquisition, then the US should respond with counter litigation for trillions against corrupt scofflaws like VW ( think diesel fiddle!) as well as UBS/Credit Suisse/HSBC/Credit Lyonnaise (think tax cheats!)and sue/litigate them out of existence.

If the lazy, corrupt, incompetent Euros want to play with fire, then let them be financially incinerated! </s> <not toxic 3.2237966060638428> Rome should never have made such inane pronouncements at Trent in their attempt to define the substance of holy Eucharist. Most reasonable people understand that perfectly well. That Rome also made their pronouncements (faith and morals) "infallible" is equally tragic, for the simple reason that so-called infallible statements cannot be retracted without calling into question other so-called infallible statements.

Sincere question for you: If Jesus and his followers celebrated Eucharist as a communal meal seated around a table, what gives Rome the right to alter this simple act of worship (perhaps "fellowship" is a better word--more suited toward love of God and neighbor), given to us by the Lord himself?

\textbf{Label} toxic

\textbf{Test Modified} I hope you don't have kids if you see this woman's actions as acceptable. And I applaud the den for kicking the kid out. She brought unwanted negative attention upon them. However, she will, and is already likely, pay the the price for her stupid stunt. </s> <toxic 2.873025417327881> Ms. Van Brocklin: You state that there is Payne's conduct is inexcusable, then proceed to use the rest of your space to justify him and his actions. You have denigrated the proud courage of countless people who took up causes via civil disobedience. I marched in the non-violent peaceful Civil Rights protest. So the dogs and the firehoses used by a certain southern sheriff were justified, by your logic. So were the citizens beaten by Chicago police during the Democratic Convention Police riots. Resolved in 20 minutes? Nonsense. If Miss Wubbels hadn't protested as she did, she'd likely have ended up in a back room, somewhere, and who know when she would have been granted the presence of a lawyer and what she would have gone through prior to that. 
 You are completely ignorant of the shortage of nurses in this country - in some cases, critical shortages. And why would anyone want to be a nurse when they are disrespected by a former state and federal prosecutor such as you. </s> <not toxic 2.8961446285247803> Well, I can't very well respect or fear an imaginary sky-being. As for my concept of character, it was good enough for the Alaska Judicial Council and Governor Knowles. But that was long ago. I've gotten older and, crikey, maybe I am going downhill. You're right about the inappropriateness of my comment. First Lady Walker's piece is very laudable and I shouldn't have taken it as an occasion to rant. (But look on the bright side: my misplaced comment gave you yet another occasion to rant about how your Fosterism is saving civilization.)

\textbf{Label} toxic

\textbf{Test Modified} no one cares what a paid liberal trolling hack like you believes lunatic,, </s> <toxic 2.5876824855804443> It always amuses me when a troll gets on, they like their own comments and simply assert everyone else is wrong. Never any evidence to rebut it just blind assertions. </s> <not toxic 2.810284376144409> Ouch... didn’t see that one coming. A liberal stealing my own line... just like they take everything else they like.

\textbf{Label} toxic

\textbf{Test Modified} Ok all you NDP "LEAP" manifesto types, where is your hero Naomi Klein? Her fawning adoration of Chavez and Venezuelan thuggery knows no bounds. I'm sure she's awfully hysterical over the thought that such a pathetic dictstorship could ever be sanctioned. </s> <toxic 2.883418321609497> Um, no. The major left-wing Labour party was decimated; Mr. Rutte lost 8 seats; and Mr. Wilders Freedom party GAINED 4 seats. Now Mr. Rutte will have to garner favour among RIGHT-WING parties to cobble together his coalition. And as to your glib little comment about people "embracing left-leaning parties", you need only look to Canada to see the fiasco that results... The corrupt and divisive Trudeau government was elected on a slate of fuzzy, alt-left promises; only to renege on most of them ! </s> <not toxic 3.0424022674560547> So anyone who doesn't share your opinion is a "raving loony?"

And as for my post being "speculation? - which part - that the Liberals are the party in power, or that this involves money?

As for me not knowing what is going on, you are correct, I am not a member of the Liberal party insider clique, as you apparently are.

\textbf{Label} toxic

\textbf{Hate Speech Offensive Step 1}

\textbf{Test Modified} ' If this ugly ass niggah can be with her , I can too . Tf she with Chuu Baka for anyways this niggah look like a... http://t.co/lVNDRDPpQT </s> <hate speech 3.680654525756836>  AtBlackface  MrMooncricket  UncleUnco  BobbyBotsods  FuckTheMallcops  RacistNegro69 ape? Fuck u broke ass racist </s> <offensive language 3.9933786392211914>  Tamela X But a wannabe ghetto white girl will deal with an unaccomplished black man for more than sex. </s> <neither 4.20903205871582>  VNNForum These Marxist kikes promoted Anti-Semitism by creating this film, what was their incentive? They just promote their enemies. 

\textbf{Label} hate speech

\textbf{Test Modified}   8220; iamkrause: No need to thank me, killing niggers is it's own reward.  8221; 

Dis my pal from our graduation when we was 26 </s> <neither 3.5232555866241455>  LanceTHESPOKEN Statement of the Day: Ppl who RT  ChiefsatWar are even dumber than he is. The ole birds of a feather type of deal. </s> <hate speech 3.5796432495117188>  Brenddan  they said some bullshit. Damn niggers </s> <offensive language 3.7068209648132324> BRUH lmaoo   8220; NoBeeetch: Them hoes was nasty but I kept eating them " roblockk: What the fuck happened to these? http://t.co/G9w10SmQdr"  8221; 

\textbf{Label} hate speech

\textbf{Test Modified} RT  IsaidNick: niggers are so ignorant http://t.co/P9jDdVsRVb </s> <hate speech 3.2627384662628174>  AtBlackface  MrMooncricket  UncleUnco  BobbyBotsods  FuckTheMallcops  RacistNegro69 ape? Fuck u broke ass racist </s> <neither 3.7804527282714844>  VNNForum These Marxist kikes promoted Anti-Semitism by creating this film, what was their incentive? They just promote their enemies. </s> <offensive language 3.9285027980804443>  Tamela X But a wannabe ghetto white girl will deal with an unaccomplished black man for more than sex. 

\textbf{Label} hate speech

\textbf{Test Modified} RT  RosieZaya1: Ur fucking white trash </s> <hate speech 2.951470136642456>  AtBlackface  MrMooncricket  UncleUnco  BobbyBotsods  FuckTheMallcops  RacistNegro69 ape? Fuck u broke ass racist </s> <offensive language 3.6144936084747314>  Tamela X But a wannabe ghetto white girl will deal with an unaccomplished black man for more than sex. </s> <neither 3.7668633460998535>  VNNForum These Marxist kikes promoted Anti-Semitism by creating this film, what was their incentive? They just promote their enemies. 

\textbf{Label} hate speech

\textbf{Test Modified}  mike ray7 congratulations, you are officially fucking retarded. </s> <neither 3.4077796936035156> RT  JakeG BasedGod: "Never go full retard" </s> <hate speech 3.479813575744629>  Brenddan  they said some bullshit. Damn niggers </s> <offensive language 3.601623773574829> BRUH lmaoo   8220; NoBeeetch: Them hoes was nasty but I kept eating them " roblockk: What the fuck happened to these? http://t.co/G9w10SmQdr"  8221; 

\textbf{Label} hate speech

\textbf{Test Modified} gonna have them pussies mix up some concrete today. teach them to pose like me. I am a badass motherfucker. and I will let you be too (: </s> <hate speech 3.227602243423462>  AtBlackface  MrMooncricket  UncleUnco  BobbyBotsods  FuckTheMallcops  RacistNegro69 ape? Fuck u broke ass racist </s> <offensive language 3.5520639419555664> BRUH lmaoo   8220; NoBeeetch: Them hoes was nasty but I kept eating them " roblockk: What the fuck happened to these? http://t.co/G9w10SmQdr"  8221; </s> <neither 3.809976816177368> RT  Venus Lynn:   8220; dylxnl: look ghetto but it work http://t.co/chrvW9dPca  8221;  128557;  128557;  128557;  128557;  128557;  128557;  128557;  128557;  128557;  128557; 

\textbf{Label} offensive language

\textbf{Test Modified}  maddieevaans u call ur bestie a bitch I'm guessing she's a dog that barks too much </s> <offensive language 3.2484242916107178> RT  Im Amy Bitches: I'm not always a bitch, sometimes I sleep </s> <hate speech 3.3403964042663574> Women who are feminist are the ugly bitches who cant find a man for themselves </s> <neither 3.8210768699645996> Gary's girl was a tranny. RT  Em 99car: I wonder what would have become of  rickspringfield if he'd written 'Gary's Girl'. 

\textbf{Label} offensive language

\textbf{Test Modified} No less than 3 bad bitches in my bed at a time... </s> <offensive language 3.2010550498962402> Then again... i shoot bitches in the ass for doing some stupid shit like key my car or bust my windows. </s> <hate speech 3.666656732559204> bitch kill yoself, go on to the bathroom and eat the pills bitch, all of em </s> <neither 4.249817848205566>   9733; BEST ASIAN MASSAGE ON THE park slope --TOP RATED SPA   9733; 718-622-0221 - 24 http://t.co/ZsAAzFL0p5 

\textbf{Label} offensive language

\textbf{Test Modified} RT  TheDrugTribe: mary isn't a backstabbing bitch that lies and deceives me </s> <offensive language 3.4536943435668945> RT  Im Amy Bitches: I'm not always a bitch, sometimes I sleep </s> <hate speech 3.6065785884857178>  vinny2vicious faggot I knew you weren't really my friend. </s> <neither 3.638406753540039> Gary's girl was a tranny. RT  Em 99car: I wonder what would have become of  rickspringfield if he'd written 'Gary's Girl'. 

\textbf{Label} offensive language

\textbf{Test Modified}  porn, android, iphone, ipad, sex, xxx, |  CloseUp | Squirting pussy and fingered asshole http://t.co/bKYeoUwWv2 </s> <offensive language 3.5574071407318115> BRUH lmaoo   8220; NoBeeetch: Them hoes was nasty but I kept eating them " roblockk: What the fuck happened to these? http://t.co/G9w10SmQdr"  8221; </s> <neither 3.6928675174713135>  DegenerateArtist Sniffs whiffy balls involuntary, cuz a FAIRY walks DOWNTOWN HAIRY,  climatechange all our scary sense of hillbilly sneeze gt; </s> <hate speech 3.7489092350006104> Wake up you stoopid dyke  RandyPawlik 

\textbf{Label} offensive language

\textbf{Test Modified} My baby guinea pig is so cute  Adorable </s> <hate speech 3.82438588142395>  whiteangelss84 y dont u end us monkeys then? U hate us because were better than crackers  amp; I love it. Deep down u know weve done nothin 2 u </s> <offensive language 3.8302881717681885> I always wanted a bull dog them hoes clean fuck a pit </s> <neither 3.8650126457214355> Breakfast fried chicken jerk chicken Tater tots white rice nd press yellow rice nd beans Mac nd cheese http://t.co/Usz8gJnZl0 

\textbf{Label} neither

\textbf{Test Modified} RT  Kick Man: Giants- Pitiful .. Jets-Pitiful .. Mets- Pitiful .. Yankees-Pitiful .. Nets- Pitiful .. Knicks-Pitiful ... Ny sports- Pitiful </s> <neither 3.8152754306793213> You know I'm not big on the NFL, but I'm so sick of hearing all of this "Black and yellow" shit. LOL   bandwagon fans and hell, GO PACKERS! </s> <offensive language 4.071953773498535> BRUH lmaoo   8220; NoBeeetch: Them hoes was nasty but I kept eating them " roblockk: What the fuck happened to these? http://t.co/G9w10SmQdr"  8221; </s> <hate speech 4.095180511474609>  whiteangelss84 y dont u end us monkeys then? U hate us because were better than crackers  amp; I love it. Deep down u know weve done nothin 2 u 

\textbf{Label} neither

\textbf{Test Modified}  jesstoth we could get matching burner phones and be ghetto fab for a few months </s> <hate speech 3.4667954444885254>  whiteangelss84 y dont u end us monkeys then? U hate us because were better than crackers  amp; I love it. Deep down u know weve done nothin 2 u </s> <offensive language 3.595543622970581> RT  NickBratton3: I wish my parents bought me a car man..
People bitch about not getting what car they want when they want it, and its free  8230; </s> <neither 3.6075007915496826> RT  Venus Lynn:   8220; dylxnl: look ghetto but it work http://t.co/chrvW9dPca  8221;  128557;  128557;  128557;  128557;  128557;  128557;  128557;  128557;  128557;  128557; 

\textbf{Label} neither

\textbf{Test Modified} Thw White Iron Band plays this weekend in Fargo,ND at the Aquarium(21+) ,Friday(10-29-10) with Charlie Parr. The next night,Saturday... </s> <hate speech 3.8393898010253906>   128514;  128514;  128514;RT  kwagiheath: Them 1st 48 Charlotte goon niccas gon Fkkk around and burn Bank Of America stadium down  128293; </s> <offensive language 3.9061098098754883> I be telling Mcgirt music ain't enough.You gotta have a non music related agenda.Them crackers want to sell something with your face. </s> <neither 3.980140447616577> You know I'm not big on the NFL, but I'm so sick of hearing all of this "Black and yellow" shit. LOL   bandwagon fans and hell, GO PACKERS! 

\textbf{Label} neither

\textbf{Test Modified} RT  dsrtvet:  FoxNews  tjoy7 And I don't have any confidence NONWHATSOEVER in you Barack! You're the sole reason why this country is in thi  8230; </s> <neither 3.78818678855896> RT  lachlan: So who wants to tell the Washington Post that Charlie Crist is a Democrat? http://t.co/LGzzYusEKZ http://t.co/2jCVv7qxqf </s> <hate speech 3.942119598388672>  whiteangelss84 y dont u end us monkeys then? U hate us because were better than crackers  amp; I love it. Deep down u know weve done nothin 2 u </s> <offensive language 3.984135150909424> BRUH lmaoo   8220; NoBeeetch: Them hoes was nasty but I kept eating them " roblockk: What the fuck happened to these? http://t.co/G9w10SmQdr"  8221; 

\textbf{Label} neither

\textbf{Hate Speech Offensive Step 5}

\textbf{Test Modified} ' If this ugly ass niggah can be with her , I can too . Tf she with Chuu Baka for anyways this niggah look like a... http://t.co/lVNDRDPpQT </s> <hate speech 2.7132835388183594> RT  WaavyLee: His balls ashy RT  Yattabing:  Trelaire1st: Real women do this http://t.co/VG5DBqH8aT  8221; and real faggots let em do that.. Smh </s> <offensive language 2.7854363918304443> RT  JayyMfCarter: If you gotta nigga or bitch.... PLEASE STAY OUT MY DM's I'm tired of y'all taken girls </s> <neither 3.0996105670928955>  NeonTreezz  PoCBeauty ......so.....white people selling  African art? NOT COOL? Okay.  blackish  redskins  doublestandard 

\textbf{Label} hate speech

\textbf{Test Modified}   8220; iamkrause: No need to thank me, killing niggers is it's own reward.  8221; 

Dis my pal from our graduation when we was 26 </s> <hate speech 2.6058554649353027>  samzbikowski some negro amigo pulled a gun on Nate and I a few weeks ago. I was STOKED!  128299;  128299;  128299; </s> <neither 2.670192003250122> RT  jennaferjenbub:   8220; BarryClerjuste: "Anything below a A+ and we disown you ling ling" http://t.co/m1QiWK4xZg  8221;  AustinBedsaul </s> <offensive language 2.740609645843506>   8220; Alondra Lu: Ain't that a bitch  8221; 

\textbf{Label} hate speech

\textbf{Test Modified} RT  IsaidNick: niggers are so ignorant http://t.co/P9jDdVsRVb </s> <hate speech 2.057800769805908> RT  WhitesOnly 1:  niggers! http://t.co/Hb3uJaLky2 </s> <neither 2.7749483585357666>  amp; thots are wearing Uggs RT  BigBootyJudy814:  ItsFallBecause negros are pulling out their Timbs" </s> <offensive language 2.926729440689087> RT  Jayy Gee96: Dumb bitches 

\textbf{Label} hate speech

\textbf{Test Modified} RT  RosieZaya1: Ur fucking white trash </s> <hate speech 2.422173500061035>  FrankieJGrande fugly queer white trash </s> <offensive language 2.6756434440612793> RT  Jayy Gee96: Dumb bitches </s> <neither 2.783188819885254> RT  BeardedNixon: Poont gotta be trash 

\textbf{Label} hate speech

\textbf{Test Modified}  mike ray7 congratulations, you are officially fucking retarded. </s> <hate speech 2.4854748249053955>  darthdanaa Yes you do retard. </s> <offensive language 2.851564645767212> Lol!!   8220; ItzSweetz Bitch: Ooop! QT  TiFFANY P0RSCHE: You little twats.  8221; </s> <neither 2.8971688747406006> RT  jennaferjenbub:   8220; BarryClerjuste: "Anything below a A+ and we disown you ling ling" http://t.co/m1QiWK4xZg  8221;  AustinBedsaul 

\textbf{Label} hate speech

\textbf{Test Modified} gonna have them pussies mix up some concrete today. teach them to pose like me. I am a badass motherfucker. and I will let you be too (: </s> <offensive language 2.7589027881622314>  40oz VAN IYCMI. I can't get any work done if you keep showin off your bitches. </s> <hate speech 2.8690829277038574>  SlightlyAdjusted RT  CapoToHeaven Alls niggers wanna do is fuck, tweet, and drink pineapple soda all day </s> <neither 3.0193798542022705>  cakedjake We're laying rock around our lake. You're welcome to join a redneck workout.  muscles   128170;  128513; 

\textbf{Label} offensive language

\textbf{Test Modified}  maddieevaans u call ur bestie a bitch I'm guessing she's a dog that barks too much </s> <offensive language 3.0170230865478516> Lol!!   8220; ItzSweetz Bitch: Ooop! QT  TiFFANY P0RSCHE: You little twats.  8221; </s> <hate speech 3.0850884914398193>  Princesslexii16 Fucking coon </s> <neither 3.21132493019104> lmaoooo RT  ComedyTruth: Girls, don't let a guy treat you like a yellow starburst. You are a pink starburst. 

\textbf{Label} offensive language

\textbf{Test Modified} No less than 3 bad bitches in my bed at a time... </s> <offensive language 3.023017406463623> Don't lose sleep bout these bitches bc they come and go   128076; </s> <hate speech 3.2786214351654053> bitch kill yoself, go on to the bathroom and eat the pills bitch, all of em </s> <neither 3.4171059131622314> Keep those away frm Charlie Day RT  JhonenV: Just once in my life I'd like for someone's favorite part of my body to be my disgusting knees. 

\textbf{Label} offensive language

\textbf{Test Modified} RT  TheDrugTribe: mary isn't a backstabbing bitch that lies and deceives me </s> <offensive language 2.991457939147949> RT  StevStiffler: If her bio says "Only God can judge me" she's a hoe. </s> <hate speech 3.098494529724121> RT  sorryimalex: I got called a faggot for buying girl toms so now I'm gonna fuck that person in the ass </s> <neither 3.3266849517822266> lmaoooo RT  ComedyTruth: Girls, don't let a guy treat you like a yellow starburst. You are a pink starburst. 

\textbf{Label} offensive language

\textbf{Test Modified}  porn, android, iphone, ipad, sex, xxx, |  CloseUp | Squirting pussy and fingered asshole http://t.co/bKYeoUwWv2 </s> <neither 1.5677733421325684>  porn, android, iphone, ipad, sex, xxx, |  Desi | paki http://t.co/XxcdQvzI9t </s> <hate speech 2.8566393852233887> RT  mitchmancuso:  BrantPrintup:What straight guys take a picture of themselves naked in a hot tun.What fags  jakesiwy  Ryan Murphy3  Randy  8230; </s> <offensive language 2.932191848754883> Lol!!   8220; ItzSweetz Bitch: Ooop! QT  TiFFANY P0RSCHE: You little twats.  8221; 

\textbf{Label} offensive language

\textbf{Test Modified} My baby guinea pig is so cute  Adorable </s> <neither 3.1643435955047607> Our female guinea pig is pregnant   127882;  127881;  127873;  128525;  128525;  128525; </s> <offensive language 3.4907007217407227> I impress da young white girl next doe by taking out my giant negro thang and usin it to flip da hamburgers for da  KoolQueefTribute  160; </s> <hate speech 3.5861661434173584> What a wetback looks like when he gets caught crossing the border.  Ilovebamf http://t.co/j3Uf1TYubO 

\textbf{Label} neither

\textbf{Test Modified} RT  Kick Man: Giants- Pitiful .. Jets-Pitiful .. Mets- Pitiful .. Yankees-Pitiful .. Nets- Pitiful .. Knicks-Pitiful ... Ny sports- Pitiful </s> <hate speech 3.3251242637634277> RT    J R: Smh nigga is mildly retarded RT  Thotcho: LMFAO RT  JustDoJ: If Griff wasn  8217;t injuries we  8217;d legit be 6-1 </s> <neither 3.328580856323242> Don't follow the astros they said. They're trash they said. Well now look at them  astros </s> <offensive language 3.3433356285095215> Them shits ugly hoe. RT  SirRocObama: ......... RT  BurgerKing: All these nuggets  amp; u still actin chicken. http://t.co/tRy8Lvyo9O 

\textbf{Label} neither

\textbf{Test Modified}  jesstoth we could get matching burner phones and be ghetto fab for a few months </s> <offensive language 3.122525930404663>  JZolly23  JBilinovich we need to grow mullets together so we can get all the bitches and  HannahKubiak can hate on us </s> <hate speech 3.291858434677124> RT  NoWomanIsRight: You can be a good girl all you want and those hoes still gonna get us niggas attention from time to time </s> <neither 3.3428289890289307> RT  Venus Lynn:   8220; dylxnl: look ghetto but it work http://t.co/chrvW9dPca  8221;  128557;  128557;  128557;  128557;  128557;  128557;  128557;  128557;  128557;  128557; 

\textbf{Label} neither

\textbf{Test Modified} Thw White Iron Band plays this weekend in Fargo,ND at the Aquarium(21+) ,Friday(10-29-10) with Charlie Parr. The next night,Saturday... </s> <neither 3.4018728733062744> Lmaooo naw man RT  DipOnline Yo want in RT  HumbltonBanks: U serious bro?? lol RT  CheezMoeJenkinz 2-3:10am early bird special </s> <hate speech 3.544551372528076>   128514;  128514;  128514;RT  kwagiheath: Them 1st 48 Charlotte goon niccas gon Fkkk around and burn Bank Of America stadium down  128293; </s> <offensive language 3.711003065109253> I be telling Mcgirt music ain't enough.You gotta have a non music related agenda.Them crackers want to sell something with your face. 

\textbf{Label} neither

\textbf{Test Modified} RT  dsrtvet:  FoxNews  tjoy7 And I don't have any confidence NONWHATSOEVER in you Barack! You're the sole reason why this country is in thi  8230; </s> <hate speech 2.696760654449463> RT  veeveeveeveevee: If I was Obama Id call a press conference  amp; slit joe bidens neck on live tv just 2 show these crackers I mean business  8230; </s> <neither 2.762817144393921> RT  jennaferjenbub:   8220; BarryClerjuste: "Anything below a A+ and we disown you ling ling" http://t.co/m1QiWK4xZg  8221;  AustinBedsaul </s> <offensive language 2.9660327434539795> RT  CoffyBrownChi: If he don't believe you, no refunds hoe. 

\textbf{Label} neither

\textbf{Hate Speech Offensive Step 10}

\textbf{Test Modified} ' If this ugly ass niggah can be with her , I can too . Tf she with Chuu Baka for anyways this niggah look like a... http://t.co/lVNDRDPpQT </s> <offensive language 2.6535706520080566> RT  CurrenSy Spitta: And if a bitch can't respect a nigga wit some paper and a fresh pair of bball shorts then she was raised terribly.. </s> <hate speech 2.7132835388183594> RT  WaavyLee: His balls ashy RT  Yattabing:  Trelaire1st: Real women do this http://t.co/VG5DBqH8aT  8221; and real faggots let em do that.. Smh </s> <neither 3.0996105670928955>  NeonTreezz  PoCBeauty ......so.....white people selling  African art? NOT COOL? Okay.  blackish  redskins  doublestandard 

\textbf{Label} hate speech

\textbf{Test Modified}   8220; iamkrause: No need to thank me, killing niggers is it's own reward.  8221; 

Dis my pal from our graduation when we was 26 </s> <hate speech 2.545886278152466> RT  Tae Rhodes:   8220; kim92493:   8220; Tae Rhodes:  kim92493  patpatbush uhhh you've been judged  8221; it happens.  whitepower...I'll hang you nigger  8221; wo  8230; </s> <offensive language 2.6044790744781494>   8220; NoRapist: on my way to fuck ur bitch http://t.co/SgVBBrwOg2  8221;  mckinley719 </s> <neither 2.670259714126587> RT  jennaferjenbub:   8220; BarryClerjuste: "Anything below a A+ and we disown you ling ling" http://t.co/m1QiWK4xZg  8221;  AustinBedsaul 

\textbf{Label} hate speech

\textbf{Test Modified} RT  IsaidNick: niggers are so ignorant http://t.co/P9jDdVsRVb </s> <hate speech 2.057800769805908> RT  WhitesOnly 1:  niggers! http://t.co/Hb3uJaLky2 </s> <neither 2.7749483585357666>  amp; thots are wearing Uggs RT  BigBootyJudy814:  ItsFallBecause negros are pulling out their Timbs" </s> <offensive language 2.8298287391662598> This Uncle Tom mother fucking wants to invoke 3/5 a man in his speech? Dude you ain't white no matter how much... http://t.co/3yrcyC9ezc 

\textbf{Label} hate speech

\textbf{Test Modified} RT  RosieZaya1: Ur fucking white trash </s> <hate speech 2.422173500061035>  FrankieJGrande fugly queer white trash </s> <offensive language 2.6756434440612793> RT  Jayy Gee96: Dumb bitches </s> <neither 2.783188819885254> RT  BeardedNixon: Poont gotta be trash 

\textbf{Label} hate speech

\textbf{Test Modified}  mike ray7 congratulations, you are officially fucking retarded. </s> <hate speech 2.4854748249053955>  darthdanaa Yes you do retard. </s> <offensive language 2.8516175746917725> Lol!!   8220; ItzSweetz Bitch: Ooop! QT  TiFFANY P0RSCHE: You little twats.  8221; </s> <neither 2.8972203731536865> RT  jennaferjenbub:   8220; BarryClerjuste: "Anything below a A+ and we disown you ling ling" http://t.co/m1QiWK4xZg  8221;  AustinBedsaul 

\textbf{Label} hate speech

\textbf{Test Modified} gonna have them pussies mix up some concrete today. teach them to pose like me. I am a badass motherfucker. and I will let you be too (: </s> <offensive language 2.758687734603882>  40oz VAN IYCMI. I can't get any work done if you keep showin off your bitches. </s> <hate speech 2.8036365509033203> Just to get u mad go on your search bar on here and search up "stupid niggers"  amp; hop on somebodys head then mention me lol  stonethegreat23 </s> <neither 3.012741804122925>  charloosss  keepitplur  nicoleariel  I'll chug my tall can . but homegirl won't approve lol 

\textbf{Label} offensive language

\textbf{Test Modified}  maddieevaans u call ur bestie a bitch I'm guessing she's a dog that barks too much </s> <hate speech 2.8469488620758057>  RylannWilliams whooooo? Chelsey? Fuck her lol. She juss a bitch </s> <offensive language 2.8842358589172363> RT   Ezzzylove: She a bad bitch, let's get to it right away . </s> <neither 3.0819990634918213>  charliesheen Charlie, im an old lady. don't EVER SAY UGLY THINGS ABOUT UR CHILDRENS MOM.. I GET IT!!!, JUS DONT! BIG HUG 

\textbf{Label} offensive language

\textbf{Test Modified} No less than 3 bad bitches in my bed at a time... </s> <offensive language 2.8522520065307617> Bad bitches in the pen make my toes curl </s> <hate speech 3.2539432048797607> I didn't forsake all other bitches for my wife to be getting fucked on by another nigga. and you know she married? you gotta die. </s> <neither 3.4170782566070557> Keep those away frm Charlie Day RT  JhonenV: Just once in my life I'd like for someone's favorite part of my body to be my disgusting knees. 

\textbf{Label} offensive language

\textbf{Test Modified} RT  TheDrugTribe: mary isn't a backstabbing bitch that lies and deceives me </s> <offensive language 2.9916186332702637> RT  StevStiffler: If her bio says "Only God can judge me" she's a hoe. </s> <hate speech 3.02489972114563>  triple6em96  Hunglikerobby  bitch you watch your fucking mouth you dirty whore. I swear to god that's a thin line </s> <neither 3.1058743000030518> RT  shakiraevanss: Criticize Amanda for saying the n word, sure, but don't make jokes about her sexual assault, don't be trash. 

\textbf{Label} offensive language

\textbf{Test Modified}  porn, android, iphone, ipad, sex, xxx, |  CloseUp | Squirting pussy and fingered asshole http://t.co/bKYeoUwWv2 </s> <neither 1.5677733421325684>  porn, android, iphone, ipad, sex, xxx, |  Desi | paki http://t.co/XxcdQvzI9t </s> <offensive language 2.8408925533294678> RT  FunnyPicsDepot: bitches be like "I'm a virgin" http://t.co/mFDwXmg8ic </s> <hate speech 2.8566393852233887> RT  mitchmancuso:  BrantPrintup:What straight guys take a picture of themselves naked in a hot tun.What fags  jakesiwy  Ryan Murphy3  Randy  8230; 

\textbf{Label} offensive language

\textbf{Test Modified} My baby guinea pig is so cute  Adorable </s> <neither 3.1643435955047607> Our female guinea pig is pregnant   127882;  127881;  127873;  128525;  128525;  128525; </s> <offensive language 3.4907007217407227> I impress da young white girl next doe by taking out my giant negro thang and usin it to flip da hamburgers for da  KoolQueefTribute  160; </s> <hate speech 3.5861661434173584> What a wetback looks like when he gets caught crossing the border.  Ilovebamf http://t.co/j3Uf1TYubO 

\textbf{Label} neither

\textbf{Test Modified} RT  Kick Man: Giants- Pitiful .. Jets-Pitiful .. Mets- Pitiful .. Yankees-Pitiful .. Nets- Pitiful .. Knicks-Pitiful ... Ny sports- Pitiful </s> <neither 3.0366640090942383>  Buster ESPN Huh.....last 10 games..Tampa 8-2/Balt 7-3/Yanks 6-4...and they lost their best pitcher. Please explain your logic. </s> <hate speech 3.3251242637634277> RT    J R: Smh nigga is mildly retarded RT  Thotcho: LMFAO RT  JustDoJ: If Griff wasn  8217;t injuries we  8217;d legit be 6-1 </s> <offensive language 3.3433356285095215> Them shits ugly hoe. RT  SirRocObama: ......... RT  BurgerKing: All these nuggets  amp; u still actin chicken. http://t.co/tRy8Lvyo9O 

\textbf{Label} neither

\textbf{Test Modified}  jesstoth we could get matching burner phones and be ghetto fab for a few months </s> <hate speech 3.034785270690918>  SAMMI boyden bruh we can finally roll like rednecks (': ((drug dealers)) </s> <offensive language 3.122525930404663>  JZolly23  JBilinovich we need to grow mullets together so we can get all the bitches and  HannahKubiak can hate on us </s> <neither 3.271000623703003> RT  sassytbh: a girl tweeted "you might be ghetto if u bring food from outside into the movies"

no u might be stupid if u pay 4.99 for a b  8230; 

\textbf{Label} neither

\textbf{Test Modified} Thw White Iron Band plays this weekend in Fargo,ND at the Aquarium(21+) ,Friday(10-29-10) with Charlie Parr. The next night,Saturday... </s> <neither 3.2462401390075684> RT  toddknife: Full  weakenednachos set (except the last song) from Southern Darkness Fest last month. Who's the ape on guitar? https://t.c  8230; </s> <hate speech 3.3524651527404785> Eagles fuck around  amp; lose it'll be kill the cracker at the Sophi crib smfh </s> <offensive language 3.511016368865967> My dawg  ceomiamimike told me it's a must I be  901k2lounge this Saturday ROCKIN that bitch wit Tha  8230; http://t.co/0NV9cHtwOs 

\textbf{Label} neither

\textbf{Test Modified} RT  dsrtvet:  FoxNews  tjoy7 And I don't have any confidence NONWHATSOEVER in you Barack! You're the sole reason why this country is in thi  8230; </s> <hate speech 2.696760654449463> RT  veeveeveeveevee: If I was Obama Id call a press conference  amp; slit joe bidens neck on live tv just 2 show these crackers I mean business  8230; </s> <neither 2.762908458709717> RT  jennaferjenbub:   8220; BarryClerjuste: "Anything below a A+ and we disown you ling ling" http://t.co/m1QiWK4xZg  8221;  AustinBedsaul </s> <offensive language 2.881894588470459>   8220; LongMoneyTonyy:  vintage monroe  DONT Say Shit Else ! Just Stfu !  8221;bitch we can do a lot more off this Twitter shit you can come see me 

\textbf{Label} neither

\textbf{LIAR (collapsed) Step 1}

\textbf{Test Modified} Afscme says In labor negotiations with city employees, Milwaukee Mayor Tom Barrett demanded concessions that went beyond those mandated by Gov. Scott Walkers collective bargaining law </s> a letter to members </s> <true statement 3.833270311355591> Donald Trump says Libya Ambassador (Christopher) Stevens sent 600 requests for help in Benghazi. </s> the second 2016 presidential debate </s> <false statement 4.013778209686279> Donald Trump says The federal government is sending refugees to states with governors who are Republicans, not to the Democrats. </s> an interview on Laura Ingraham's radio show 

\textbf{Label} true statement

\textbf{Test Modified} Rick Scott says All Aboard Florida is a 100 percent private venture. There is no state money involved. </s> a TV interview </s> <false statement 3.664231777191162> Donald Trump says The federal government is sending refugees to states with governors who are Republicans, not to the Democrats. </s> an interview on Laura Ingraham's radio show </s> <true statement 3.831820011138916> Patrick Murphy says Marco Rubio opposes immigration reform. Worse, Rubio supports Donald Trump. His plan would deport 800,000 children, destroying families. </s> a TV ad 

\textbf{Label} true statement

\textbf{Test Modified} Julie Pace says The Obama administration is using as its legal justification for these airstrikes (on the Islamic State), an authorization for military force that the president himself has called for repeal of. </s> a question to White House Press Secretary Josh Earnest </s> <false statement 3.5803754329681396> Donald Trump says Hillary Clinton invented ISIS with her stupid policies. She is responsible for ISIS. </s> an interview on 60 Minutes </s> <true statement 3.869307518005371> Donald Trump says Libya Ambassador (Christopher) Stevens sent 600 requests for help in Benghazi. </s> the second 2016 presidential debate 

\textbf{Label} true statement

\textbf{Test Modified} John Kasich says We are now eighth in the nation in job creation . . . we are No. 1 in the Midwest. </s> a news conference </s> <true statement 3.851958990097046> Jorge Elorza says In the last six years of Ciancis administration violent crime was down in the United States. It was down in the region. It was down in Rhode Island. But it was up in Providence. </s> a debate </s> <false statement 4.010262966156006> Donald Trump says The federal government is sending refugees to states with governors who are Republicans, not to the Democrats. </s> an interview on Laura Ingraham's radio show 

\textbf{Label} true statement

\textbf{Test Modified} Mike Pence says It was Hillary Clinton who left Americans in harms way in Benghazi and after four Americans fell said, What difference at this point does it make? </s> the Republican national convention </s> <true statement 3.7440342903137207> Jorge Elorza says In the last six years of Ciancis administration violent crime was down in the United States. It was down in the region. It was down in Rhode Island. But it was up in Providence. </s> a debate </s> <false statement 3.746598958969116> Donald Trump says You will learn more about Donald Trump by going down to the Federal Elections to see the financial disclosure form than by looking at tax returns. </s> a Presidential debate at Hofstra University 

\textbf{Label} true statement

\textbf{Test Modified} Rand Paul says Of the roughly 15 percent of Americans who dont have health insurance, half of them made more than 50,000 a year. </s> an interview on Comedy Central's "The Daily Show" </s> <true statement 3.7997491359710693> Bernie S says We have the highest rate of childhood poverty of any major country on Earth. </s> an interview on CNN </s> <false statement 3.9633538722991943> Donald Trump says The federal government is sending refugees to states with governors who are Republicans, not to the Democrats. </s> an interview on Laura Ingraham's radio show 

\textbf{Label} false statement

\textbf{Test Modified} Barack Obama says Stimulus tax cuts "began showing up in paychecks of 4.8 million Indiana households about three months ago." </s> a speech in Wakarusa, Ind. </s> <true statement 3.8199117183685303> Jorge Elorza says In the last six years of Ciancis administration violent crime was down in the United States. It was down in the region. It was down in Rhode Island. But it was up in Providence. </s> a debate </s> <false statement 3.916092872619629> Donald Trump says The federal government is sending refugees to states with governors who are Republicans, not to the Democrats. </s> an interview on Laura Ingraham's radio show 

\textbf{Label} false statement

\textbf{Test Modified} Allen West says If you look at the application for a security clearance, I have a clearance that even the president of the United States cannot obtain because of my background. </s> a candidate forum </s> <false statement 3.760773181915283> Rush Limbaugh says 11 straight years of no major hurricanes striking land in the United States bores a hole right through the whole climate change argument. </s> a radio show broadcast </s> <true statement 3.77760648727417> Arizona Citizens Defense League says a gun bill before the Senate would make it a federal felony to leave town for more than seven days, and leave someone else at home with your firearms. </s> an email to supporters 

\textbf{Label} false statement

\textbf{Test Modified} Bernie S says We now work the longest hours of any people around the world. </s> a C-SPAN interview </s> <true statement 3.7155606746673584> Bernie S says We have the highest rate of childhood poverty of any major country on Earth. </s> an interview on CNN </s> <false statement 4.0561442375183105> Rush Limbaugh says 11 straight years of no major hurricanes striking land in the United States bores a hole right through the whole climate change argument. </s> a radio show broadcast 

\textbf{Label} false statement

\textbf{Test Modified} Sarah Palin says Donald Trumps conversion to pro-life beliefs are akin to Justin Biebers, who said in the past that abortion was no big deal to him. </s> an interview on CNN </s> <false statement 3.7367687225341797> Donald Trump says The federal government is sending refugees to states with governors who are Republicans, not to the Democrats. </s> an interview on Laura Ingraham's radio show </s> <true statement 3.7425951957702637> Patrick Murphy says Marco Rubio opposes immigration reform. Worse, Rubio supports Donald Trump. His plan would deport 800,000 children, destroying families. </s> a TV ad 

\textbf{Label} false statement

\textbf{LIAR (collapsed) Step 5}

\textbf{Test Modified} Afscme says In labor negotiations with city employees, Milwaukee Mayor Tom Barrett demanded concessions that went beyond those mandated by Gov. Scott Walkers collective bargaining law </s> a letter to members </s> <false statement 3.131746292114258> Tom Barrett says Gov. Scott Walker said no to equal pay for equal work for women. </s> a TV ad </s> <true statement 3.1800825595855713> Scott Walker says If public employees dont pay more for benefits starting April 1, 2011, the equivalent is 1,500 state employee layoffs by June 30, 2011 and 10,000 to 12,000 state and local government employee layoffs in the next two years. </s> a news conference 

\textbf{Label} true statement

\textbf{Test Modified} Rick Scott says All Aboard Florida is a 100 percent private venture. There is no state money involved. </s> a TV interview </s> <true statement 3.0582425594329834> Charlie Crist says All Aboard Florida is receiving millions in Florida taxpayer dollars. </s> a fundraising email </s> <false statement 3.1522974967956543> Corey Lewandowski says Mr. Trump is self-financing his campaign, so we dont have any donors. </s> a radio interview. 

\textbf{Label} true statement

\textbf{Test Modified} Julie Pace says The Obama administration is using as its legal justification for these airstrikes (on the Islamic State), an authorization for military force that the president himself has called for repeal of. </s> a question to White House Press Secretary Josh Earnest </s> <true statement 2.9627556800842285> Martha Raddatz says The Obama administration originally wanted 10,000 troops to remain in Iraq -- not combat troops, but military advisers, special operations forces, to watch the counterterrorism effort. </s> comments on ABC's "This Week" </s> <false statement 3.246009588241577> Rick Perry says Obama has chosen to deny the vicious anti-Semitic motivation of the attack on a kosher Jewish grocery in Paris. </s> a statement 

\textbf{Label} true statement

\textbf{Test Modified} John Kasich says We are now eighth in the nation in job creation . . . we are No. 1 in the Midwest. </s> a news conference </s> <false statement 2.610369920730591> Ted Strickland says Gov. John Kasich incorrectly claimed Ohios economy was 38th in the nation when he took office. We were sixth in the nation in terms of economic job growth. </s> an interview on CNN </s> <true statement 3.028876543045044> Terry Mcauliffe says If you take the population growth here in Virginia, we are net zero on job creation since (Bob McDonnell) became governor. </s> a speech. 

\textbf{Label} true statement

\textbf{Test Modified} Mike Pence says It was Hillary Clinton who left Americans in harms way in Benghazi and after four Americans fell said, What difference at this point does it make? </s> the Republican national convention </s> <true statement 2.5875017642974854> Hillary Clinton says When terrorists killed more than 250 Americans in Lebanon under Ronald Reagan, the Democrats didnt make that a partisan issue. </s> a CNN town hall </s> <false statement 2.9331557750701904> Facebook Posts says Hillary Clinton refuses to testify before Congress about the 2012 attack in Benghazi. </s> a meme on social media 

\textbf{Label} true statement

\textbf{Test Modified} Rand Paul says Of the roughly 15 percent of Americans who dont have health insurance, half of them made more than 50,000 a year. </s> an interview on Comedy Central's "The Daily Show" </s> <true statement 2.932455062866211> Joe Biden says Among the money spent on health care in the United States, "46 cents on every dollar spent is through Medicare and Medicaid." </s> an interview on NBC's 'Meet the Press' </s> <false statement 3.02447247505188> Trent Franks says The top 1 percent pay over half of the entire revenue for this country. </s> an interview on MSNBC's 'The Dylan Ratigan Show' 

\textbf{Label} false statement

\textbf{Test Modified} Barack Obama says Stimulus tax cuts "began showing up in paychecks of 4.8 million Indiana households about three months ago." </s> a speech in Wakarusa, Ind. </s> <false statement 2.8908281326293945> Paul Broun says Stimulus money funded a government board that made recommendations that would cost 378,000 jobs and 28.3 billion in sales. </s> a tweet </s> <true statement 2.9225375652313232> Sarah Palin says "One state even spent a million bucks to put up signs that advertise that they were spending on the federal stimulus projects." </s> an address at the Tea Party convention 

\textbf{Label} false statement

\textbf{Test Modified} Allen West says If you look at the application for a security clearance, I have a clearance that even the president of the United States cannot obtain because of my background. </s> a candidate forum </s> <false statement 3.050549268722534> Ted Cruz says One of the most troubling aspects of the Rubio-Schumer Gang of Eight bill was that it gave President Obama blanket authority to admit refugees, including Syrian refugees, without mandating any background checks whatsoever. </s> a Republican presidential debate in Las Vegas </s> <true statement 3.196129560470581> David Shuster says Said former U.S. Ambassador to Kenya Scott Gration was forced to resign two years ago because of his personal use of emails. </s> a Hillary Clinton press conference 

\textbf{Label} false statement

\textbf{Test Modified} Bernie S says We now work the longest hours of any people around the world. </s> a C-SPAN interview </s> <true statement 3.08957576751709> Jim Sensenbrenner says We have the highest corporate tax rate in the world. Its 35 percent. </s> an interview </s> <false statement 3.3488667011260986> Mitt Romney says Today there are more men and women out of work in America than there are people working in Canada. </s> a speech to the Conservative Political Action Conference 

\textbf{Label} false statement

\textbf{Test Modified} Sarah Palin says Donald Trumps conversion to pro-life beliefs are akin to Justin Biebers, who said in the past that abortion was no big deal to him. </s> an interview on CNN </s> <false statement 3.1018259525299072> Herman Cain says Said Planned Parenthoods early objective was to help kill black babies before they came into the world. </s> a talk at a conservative think tank </s> <true statement 3.1297004222869873> Greg Abbott says After Texas defunded Planned Parenthood, both the unintended pregnancy and abortion rates dropped. </s> a tweet 

\textbf{Label} false statement

\textbf{LIAR (collapsed) Step 10}

\textbf{Test Modified} Afscme says In labor negotiations with city employees, Milwaukee Mayor Tom Barrett demanded concessions that went beyond those mandated by Gov. Scott Walkers collective bargaining law </s> a letter to members </s> <false statement 3.131746292114258> Tom Barrett says Gov. Scott Walker said no to equal pay for equal work for women. </s> a TV ad </s> <true statement 3.1403446197509766> Portland Association Teachers says Did you know that if you accepted the Districts proposal today you would have NO pay increase for 4 years? Seven years of frozen wages = Disrespect. </s> a newsletter 

\textbf{Label} true statement

\textbf{Test Modified} Rick Scott says All Aboard Florida is a 100 percent private venture. There is no state money involved. </s> a TV interview </s> <true statement 3.0582022666931152> Charlie Crist says All Aboard Florida is receiving millions in Florida taxpayer dollars. </s> a fundraising email </s> <false statement 3.152191162109375> Corey Lewandowski says Mr. Trump is self-financing his campaign, so we dont have any donors. </s> a radio interview. 

\textbf{Label} true statement

\textbf{Test Modified} Julie Pace says The Obama administration is using as its legal justification for these airstrikes (on the Islamic State), an authorization for military force that the president himself has called for repeal of. </s> a question to White House Press Secretary Josh Earnest </s> <true statement 2.962770462036133> Martha Raddatz says The Obama administration originally wanted 10,000 troops to remain in Iraq -- not combat troops, but military advisers, special operations forces, to watch the counterterrorism effort. </s> comments on ABC's "This Week" </s> <false statement 2.9962246417999268> Rand Paul says The president is advocating a drone strike program in America. </s> a tweet 

\textbf{Label} true statement

\textbf{Test Modified} John Kasich says We are now eighth in the nation in job creation . . . we are No. 1 in the Midwest. </s> a news conference </s> <false statement 2.610369920730591> Ted Strickland says Gov. John Kasich incorrectly claimed Ohios economy was 38th in the nation when he took office. We were sixth in the nation in terms of economic job growth. </s> an interview on CNN </s> <true statement 2.896986246109009> John Kasich says We are in the bottom 10 in dollars in the classroom and the top 10 in dollars in the bureaucracy and red tape. </s> an interview on Fox News 

\textbf{Label} true statement

\textbf{Test Modified} Mike Pence says It was Hillary Clinton who left Americans in harms way in Benghazi and after four Americans fell said, What difference at this point does it make? </s> the Republican national convention </s> <true statement 2.5874826908111572> Hillary Clinton says When terrorists killed more than 250 Americans in Lebanon under Ronald Reagan, the Democrats didnt make that a partisan issue. </s> a CNN town hall </s> <false statement 2.849807024002075> Donald Trump says Sidney Blumenthal wrote that the Benghazi attack was almost certainly preventable. Clinton was in charge of the State Department, and it failed to protect U.S. personnel and an American consulate in Libya. </s> a rally in Wilkes-Barre, Pa. 

\textbf{Label} true statement

\textbf{Test Modified} Rand Paul says Of the roughly 15 percent of Americans who dont have health insurance, half of them made more than 50,000 a year. </s> an interview on Comedy Central's "The Daily Show" </s> <false statement 2.9004263877868652> Rand Paul says Over half of the young people in medical, dental and law schools are women. </s> an interview with CNN </s> <true statement 2.932455062866211> Joe Biden says Among the money spent on health care in the United States, "46 cents on every dollar spent is through Medicare and Medicaid." </s> an interview on NBC's 'Meet the Press' 

\textbf{Label} false statement

\textbf{Test Modified} Barack Obama says Stimulus tax cuts "began showing up in paychecks of 4.8 million Indiana households about three months ago." </s> a speech in Wakarusa, Ind. </s> <false statement 2.8908281326293945> Paul Broun says Stimulus money funded a government board that made recommendations that would cost 378,000 jobs and 28.3 billion in sales. </s> a tweet </s> <true statement 2.898074150085449> Chain Email says Having an entirely Democrat congressional delegation in 2009, when the [federal stimulus] bill passed, increases the per capita stimulus dollars that the state receives per person by 460. </s> a message via the Internet 

\textbf{Label} false statement

\textbf{Test Modified} Allen West says If you look at the application for a security clearance, I have a clearance that even the president of the United States cannot obtain because of my background. </s> a candidate forum </s> <false statement 3.02140736579895> Steve Southerland says 92 percent of President Barack Obamas administration has never worked outside government. </s> comments at the Liberty County Chamber of Commerce annual dinner. </s> <true statement 3.1747167110443115> John Mccain says "The fact is it's not amnesty." </s> a debate in Manchester, N.H. 

\textbf{Label} false statement

\textbf{Test Modified} Bernie S says We now work the longest hours of any people around the world. </s> a C-SPAN interview </s> <false statement 3.0254147052764893> Bernie S says We spend twice as much per capita on health care as any other nation on Earth. </s> an appearance on the Rachel Maddow Show </s> <true statement 3.08957576751709> Jim Sensenbrenner says We have the highest corporate tax rate in the world. Its 35 percent. </s> an interview 

\textbf{Label} false statement

\textbf{Test Modified} Sarah Palin says Donald Trumps conversion to pro-life beliefs are akin to Justin Biebers, who said in the past that abortion was no big deal to him. </s> an interview on CNN </s> <false statement 2.7887768745422363> Donald Trump says Public support for abortion is actually going down a little bit, polls show. </s> comments on CNN's "State of the Union" </s> <true statement 3.1297004222869873> Greg Abbott says After Texas defunded Planned Parenthood, both the unintended pregnancy and abortion rates dropped. </s> a tweet 

\textbf{Label} false statement

\end{document}